\documentclass{article}
\usepackage{preprint,times}

\usepackage[utf8]{inputenc} 
\usepackage[T1]{fontenc}    
\usepackage{hyperref}       
\usepackage{url}            
\usepackage{booktabs}       
\usepackage{amsfonts}       
\usepackage{nicefrac}       
\usepackage{microtype}      
\usepackage{xcolor}         
\usepackage{hyperref}       
\usepackage{url}            
\usepackage{booktabs}       
\usepackage{amsfonts}       
\usepackage{nicefrac}       
\usepackage{microtype}      
\usepackage{lipsum}		
\usepackage{graphicx}
\usepackage{natbib}
\usepackage{doi}
\usepackage{float} 
\usepackage{amsmath}
\usepackage{wrapfig}
\usepackage{amssymb}
\usepackage{mathtools}
\usepackage{amsthm}
\usepackage{xcolor}
\usepackage{listings}
\usepackage{xcolor}
\usepackage{titletoc}
\usepackage{amsmath,amssymb}
\usepackage{caption}
\usepackage{aliascnt}
\usepackage{wrapfig}
\usepackage{caption}        
\usepackage{algorithmicx}
\usepackage{algpseudocode}
\usepackage{mdframed}       
\usepackage{booktabs}
\usepackage{multirow}
\usepackage{graphicx}
\usepackage{booktabs}
\usepackage{subcaption}
\usepackage{caption}
\newcounter{algorithm}
\renewcommand{\thealgorithm}{\arabic{algorithm}}
\theoremstyle{plain}
\newtheorem{theorem}{Theorem}[section]
\newaliascnt{proposition}{theorem}
\newtheorem{proposition}[proposition]{Proposition}
\aliascntresetthe{proposition}
\newaliascnt{challenge}{theorem}

\aliascntresetthe{challenge}
\newaliascnt{question}{theorem}

\aliascntresetthe{question}
\newtheorem{lemma}{Lemma}

\theoremstyle{definition}

\theoremstyle{remark}

\usepackage{threeparttable}
\title{Physics-Informed One-Step Generation on Long-short Flow Maps}

\author{%
  Zhiqi Li, Duowen Chen, Jinjin He, Bo Zhu \\
  Georgia Institute of Technology \\
  \texttt{zli3167@gatech.edu}
}

\usepackage{needspace}

\begin{document}

\maketitle

\begin{abstract}
We propose Physics-Informed Long-Short Flow Maps with Direct Guidance (PI-LSFM), a one-step generative method for producing data-aligned and physically valid samples. 
Existing one-step physics-informed generative models must jointly control data fidelity and physical constraints without the iterative correction of multi-step samplers, making it difficult to coordinate indirect surrogate supervision with strong physics enforcement. 
PI-LSFM addresses this challenge by decomposing generation into a long flow map for global transport and a near-terminal short flow map, where closed-form data and physics potentials provide directly computable guidance targets instead of indirect regression surrogates. 
The short map guides the long map during training, yielding a single one-step generator that requires neither staged distillation nor inference-time correction. 
Across Darcy flow, Kolmogorov flow, dynamic stall, and topology optimization, PI-LSFM achieves a favorable trade-off among distributional fidelity, physical residuals, and inference efficiency while using only one function evaluation at inference.

\end{abstract}

\vspace{-2mm}
\section{Introduction}

Many problems in scientific machine learning are inherently distributional: under uncertain boundary conditions, perturbed parameters, incomplete observations, or intrinsic stochasticity, one seeks not a single deterministic solution, but a distribution of physically feasible ones. This makes generative modeling a natural framework for physical systems, as it can capture high-dimensional solution distributions while incorporating PDEs, conservation laws, and other physical constraints. Recent works have therefore introduced physics-informed diffusion and related generative frameworks to produce samples that are both statistically plausible and physically valid \cite{shu2023physics, huang2024diffusionpde,jacobsen2025cocogen}. However, these methods typically rely on iterative multi-step generation and are often slow for practical use, which naturally motivates fast and ideally one-step physics-informed generation.

The main difficulty, however,  is not merely that one-step generation is harder to train, but that existing physics-informed diffusion methods rely on indirect supervision for both data and physics.  On the data side, flow matching learns the data distribution through the surrogate regression objective $\mathcal{L}_{\mathrm{data}}=\mathbb{E}[|u_t^\theta(x_t)-u_t(x_t| x_1)|^2]$,  where the model is supervised by the conditional velocity $u_t(x_t| x_1)$ because the true marginal velocity $u_t(x_t)$ is unavailable.  On the physics side, directly applying a nonlinear residual to intermediate states causes a Jensen-type mismatch~\cite{bastek2024physics}; existing methods therefore impose the residual on a reconstructed endpoint, $\mathcal{L}_{\mathrm{phys}}=\mathbb{E}[|\mathcal{R}(\hat{x}_1(x_t))|^2]$,  and use it to supervise intermediate states indirectly. 
This double indirectness makes distributional and physical objectives difficult to coordinate \cite{baldan2026physics}.  The issue is amplified in one-step methods such as MeanFlow~\cite{geng2025mean} and ShortCut~\cite{frans2025shortcut}.  A long-range flow map $\psi_{t\to r}$ is difficult to learn and, unlike an instantaneous velocity, lacks a consistent data-conditioned counterpart $\psi_{t\to r}(\cdot| x_1)$ for surrogate supervision~\cite{li2025functional,li2026trajectory}.  Moreover, one-step sampling removes the iterative correction available in multi-step methods~\cite{bastek2024physics}, requiring stronger physics enforcement during training.  Thus, one-step physics-informed generation must couple a harder indirect distribution objective with a stronger physics objective, making training fragile. Even in endpoint-prediction methods such as sCM \cite{lu2024simplifying}, where the residual can be applied directly to $\psi_{t\to 1}(x_t)$, the data objective remains indirect. \cite{chang2026stabilizing} show that coupling strong physics constraints with indirect distribution objectives can lead to trivial solutions, severe mode collapse, and the need for two-stage distillation.

\begin{wrapfigure}{r}{0.45\linewidth}
    \vspace{-5mm}
    \centering
    \includegraphics[width=0.9\linewidth]{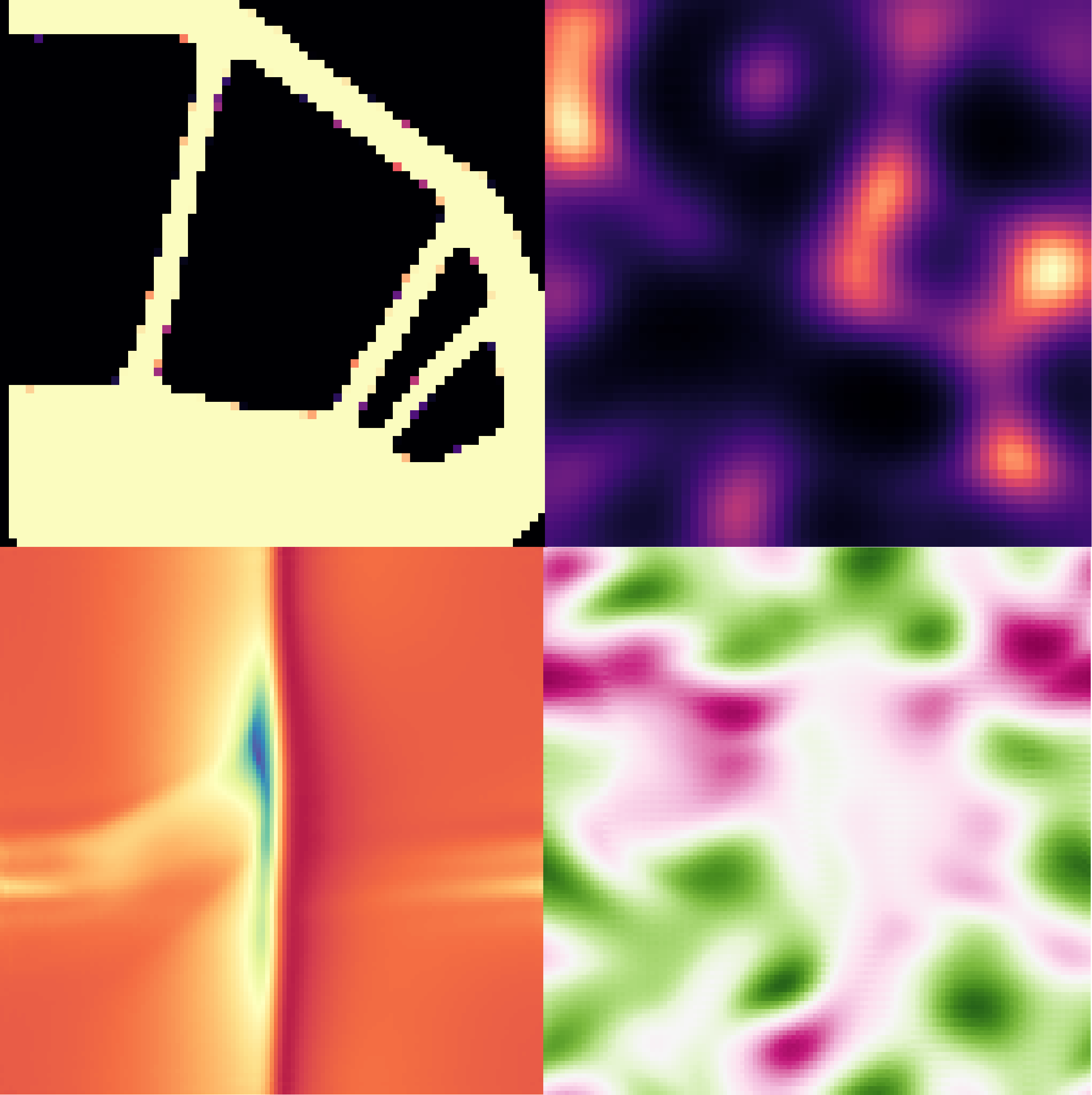}
    \vspace{-2mm}
    \caption{PI-LSFM enables physics-informed one-step generation from scratch via long-short flow-map direct guidance, achieving multi-step-level performance on unconditional generation, conditional generation, and inverse design without increasing batch size or training steps, and without extra inference-time parameters or iterative correction.}
    \label{fig:intro_result}
    \vspace{-5mm}
\end{wrapfigure}
In this work, we study physics-informed one-step generation in flow-map-based frameworks and aim to enable stable training from scratch through direct data-distribution and physics-constraint objectives. This is difficult in existing methods: closed-form flow-matching solutions are typically too inefficient for practical training \cite{bertrand2025closed}, while directly imposing physical constraints at intermediate states leads to Jensen's gap. Our key observation is that both objectives can be imposed directly in the near-terminal data neighborhood, where they provide effective and correct guidance for learning the full flow map. This motivates a long-short flow-map approach, which decomposes one-step generation into a long flow map for global evolution and a near-terminal short flow map driven by computable closed-form solutions and physics potentials. As $\Delta t\to 0$, the short flow map shrinks away while providing direct data-distribution and physics-constraint guidance for learning the long flow map. This yields Long-Short Flow Maps with Direct Guidance (PI-LSFM), where data learning and physical validity are guided by near-terminal, directly computable targets rather than conditional-regression surrogates or reconstructed terminal states. A detailed discussion of related work is provided in \autoref{sec:related_work}.

\textbf{Our contributions:} 
(1) We propose a long-short flow-map framework that addresses indirect-objective instability by modeling the short flow map as dynamics jointly driven by data and physics potentials, enabling direct guidance from distributional and physical constraints for one-step generation. (2) We study the numerical design and computation of the short flow map, leading to a stable and efficient training procedure with improved distributional fidelity and physical consistency.  (3) We validate PI-LSFM on unconditional generation, conditional generation, and topology optimization, where it consistently achieves strong performance on practical physics-informed generation tasks.

\vspace{-1mm}
\section{Preliminaries and Motivations}
\vspace{-1.mm}
\subsection{Physics-Informed Generation and Indirect Objectives}\label{sec:physics_informed_generation}

Given a dataset $\mathcal{D}=\{x^i \in \mathcal{X}\}_{i=1}^n$ drawn from an unknown distribution $p_{\mathrm{data}}$ on $\mathcal{X}\subset\mathbb{R}^d$, a physics-informed generative model aims to learn a transport from a base distribution $x_0\sim p_0$ to the target distribution $x_1\sim p_1=p_{\mathrm{data}}$, while enforcing the generated samples to satisfy a physical constraint $\mathcal{R}(x_1)=0$. A common continuous-time formulation, exemplified by Flow Matching, parameterizes a velocity field $u^\theta(x,t)$ and defines generation through the ODE
\vspace{-1mm}
{
\setlength{\belowdisplayskip}{0pt}
\begin{equation}\label{eq:ODE_sampling}
\frac{d x_t}{d t} = u^\theta(x_t, t),
\qquad x_0 \sim p_0 .
\end{equation}
}
For training the model, modern generative methods, including diffusion and flow-based models, generally do not admit direct supervision for the final generative transport from the dataset, since no efficiently computable closed-form solution is available (see \autoref{appendix:closed_form}), and therefore rely on surrogate training objectives. For example, Flow Matching constructs a conditional path $p_t(x_t|x_1)$ with $x_1 \sim p_1$ and uses the corresponding conditional velocity $u_t(x_t | x_1) = \frac{x_1-x_t}{1-t}$ as a surrogate target, with $u_t(x_t) = \mathbb{E}_{x_1\sim p_{1|t}(x_t)}[u_t(x_t | x_1)]$, and optimizes 
\vspace{-1mm}
{
\setlength{\belowdisplayskip}{-2pt}
\begin{equation}\label{eq:flow_matching}
    \begin{aligned}
        \mathcal{L}_{\mathrm{data}}(\theta) = \mathbb{E}_{t, x_1\sim p_1, x_t\sim p_{t|1}(x_t|x_1)}[\|u^\theta(x_t,t)-u_t(x_t| x_1)\|^2]
    \end{aligned}
\end{equation} 
}

To enforce $\mathcal{R}(x_1)=0$, prior work shows that directly imposing the constraint on the intermediate state $x_t$, or on its instantaneous prediction $x_1(x_t)=u_t^\theta(x_t)(1-t)+x_t$, leads to Jensen's gap $\mathcal{R}\!(\mathbb{E}[x_1 | x_t]) \neq \mathbb{E}[\mathcal{R}(x_1(x_t))| x_t]$ \cite{bastek2024physics}. Existing methods therefore enforce the residual on a surrogate endpoint $\hat{x}_1^\theta(x_t)$ reconstructed from $x_t$, typically through a multi-step auxiliary DDIM- or Flow-Matching-style inference process, and optimize the indirect physics loss $\mathcal{L}_{phys}(\theta) = \mathbb{E}_{t, x_1\sim p_1, x_t\sim p_{t|1}(x_t|x_1)}[\|\mathcal{R}(\hat{x}_1^\theta(x_t))\|^2]$ with respect to $\theta$, encouraging the final generated sample associated with $x_t$ to satisfy the physical constraint.

\begin{wrapfigure}{r}{0.4\columnwidth}
  \centering
  \vspace{-2.0em}
  \includegraphics[width=1.0\linewidth, trim=6pt 6pt 6pt 6pt, clip]{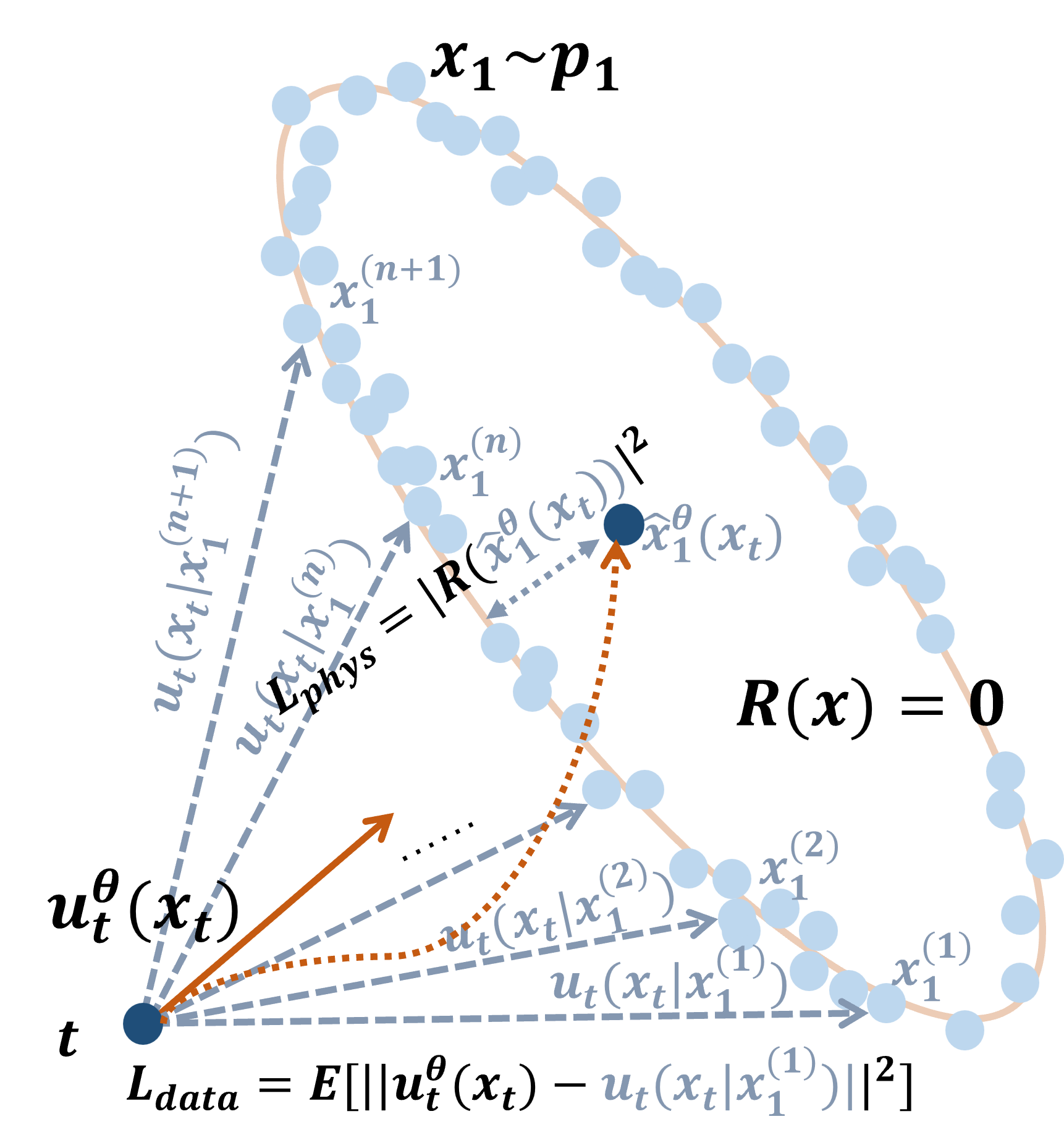}
\caption{\textbf{Indirect Losses.} $L_{\mathrm{data}}$ does not directly supervise $u_t^\theta$ with its true target $u_t(\cdot)$, but instead regresses to the conditional velocity $u_t(x_t | x_1)$. Likewise, $L_{\mathrm{physics}}$ is not applied directly to $u_t^\theta$, but only to the terminal prediction $\hat{x}_1^\theta(x_t)$ obtained after several integration steps. The indirectly supervised quantities are highlighted in muted blue.}
  \label{fig:Indirect_loss}
  \vspace{-2.2em}
\end{wrapfigure}

In practice, these two indirect objectives can pull optimization toward either better data fidelity or better constraint satisfaction, often at the expense of the other \cite{bastek2024physics, jacobsen2025cocogen, utkarsh2025physics}. Recent work~\cite{baldan2026physics} mitigates this issue in multi-step generative models through conflict-free optimization. In addition, multi-step sampling with $u^\theta(x,t)$ requires numerically discretizing \autoref{eq:ODE_sampling} from $x_0 \sim p_0$ over many steps, which remains expensive in practice and motivates one-step generation.


\vspace{-1.5mm}
\subsection{One-Step Generation on Flow Maps}\label{sec:one_step_generation}
\vspace{-1.5mm}

Specifically, flow-map methods, including ShortCut~\cite{frans2025shortcut}, MeanFlow~\cite{geng2025mean}, and consistency-based models~\cite{geng2024consistency}, learn a family of mappings $\psi_{t\to r}:\mathcal{X}\to\mathcal{X}$ along trajectories of \autoref{eq:ODE_sampling}, such that $\psi_{0\to t}(x_0)=x_t$ and $\psi_{t\to r}=\psi_{0\to r}\circ \psi_{0\to t}^{-1}$. Flow maps satisfy
\begin{equation}\label{eq:evolv_flow_map}
\frac{\partial}{\partial t}\psi_{0\to t} = u_t\circ \psi_{0\to t},\qquad \psi_{0\to 0}=\mathrm{Id}_{\mathcal{X}}.
\end{equation}
Accordingly, the marginal probability path $x_t\sim p_t$ can be written as the pushforward $p_t = (\psi_{0\to t})_\sharp p_0$. Once $\psi_{t\to r}$ is learned, flow-map methods enable one-step generation via $x_1=\psi_{0\to 1}(x_0)$.  A defining property of flow maps is their \emph{trajectory consistency}:
\begin{proposition}[Trajectory Consistency]\label{prop:trajectory_consistency}
For any $0\le t \le s \le r \le 1$, the flow map satisfies
\begin{equation}
\psi_{t\to r} = \psi_{s\to r}\circ \psi_{t\to s}.
\end{equation}
\end{proposition}
This property can be encouraged by the trajectory-consistency loss $\mathcal{L}^{\mathrm{TC}}(\theta) = \mathbb{E}_{t,s,r,x_t\sim p_t}[ \| \psi^\theta_{t\to r}(x_t) - \psi^\theta_{s\to r}\!(\psi^\theta_{t\to s}(x_t)) \|_2^2 ]$. However, $\mathcal{L}^{\mathrm{TC}}$ itself contains no dataset-level supervision. In particular, introducing data supervision for \emph{long-range} maps $\psi_{t\to r}$ is difficult: unlike the instantaneous velocity $u(x,t)$, such maps generally do not admit an easily computable conditional counterpart $\psi_{t\to r}(\cdot | x_1)$ from data samples over large time gaps; see \autoref{thm:non_exist}. Thus, indirectly learning $\psi_{t\to r}$ from data is substantially harder than supervising local dynamics.

Existing flow-map methods inject dataset supervision through progressive extension or derivative-based objectives; see \autoref{sec:flow_map_learning}. 
For instance, MeanFlow use the averaged velocity 
$u_{t\to r}(x)=\frac{\psi_{t\to r}(x)-x}{r-t}$ and trains it using a derivative-based surrogate regression target $\|u_{t\to r}^\theta(x)\!-\! sg(u_t(x|x_1)\!+\!(r\!-\!t)(\partial_t u_{t\to r}^\theta(x) \!+\! u_t(x|x_1)\partial_x u_{t\to r}^\theta(x)))\|^2$ with the conditional velocity $u_t(x\mid x_1)$.  On the physical-supervision side, variable-endpoint methods such as ShortCut and MeanFlow remain problematic: $\psi_{t\to r}(x_t)$ is still an intermediate state, so directly imposing a nonlinear residual suffers from the Jensen-type mismatch.  Endpoint-fixed consistency models can apply the residual to $\psi_{t\to 1}(x_t)$, but without multi-step correction they require stronger physics enforcement during training.  This strong physics loss interacts poorly with the indirect, high-order flow-map objective, making optimization fragile, as illustrated in \autoref{fig:toy_3x3}, motivating the staged distillation of~\cite{chang2026stabilizing}. {\color{black}Table~\ref{tab:toy_constraint_satisfaction} complements this visualization with a comparison of physical residuals across batch sizes in the same toy experiment.}

\vspace{-1.5mm}
\section{Physics-Informed Long-Short Flow-Map Method}
\vspace{-1.5mm}
To address the above challenges, we propose a \textbf{Physics-Informed Long-Short Flow Map with Direct Guidance (PI-LSFM)} framework, which treats dataset-level supervision and physical constraints as direct near-terminal guidance derived from closed-form solutions. Applying such guidance to the full generative transport is difficult, since closed-form data guidance is expensive away from the terminal time and direct physics guidance can suffer from Jensen-type mismatch; see \autoref{sec:physics_informed_generation}. Near $t=1$, however, both issues are alleviated: the closed-form solution becomes efficient to compute (\autoref{appendix:closed_form}), and physics constraints can be applied at the endpoint, avoiding intermediate-state Jensen-type mismatch since $\mathcal{R}(\mathbb{E}[x_1 \mid x_t]) = \mathbb{E}[\mathcal{R}(x_1(x_t)) \mid x_t] = \mathcal{R}(x_1)$ at $t=1$. This motivates our long-short decomposition with Proposition~\autoref{prop:trajectory_consistency}.
\begin{equation}\label{eq:decomposition}
\psi_{0\to 1}
=
\psi_{1-\Delta t\to 1}
\circ
\psi_{0\to 1-\Delta t},
\end{equation}
where a long flow map $\psi_{0\to 1-\Delta t}$ transports noise close to the data manifold, and a short terminal flow map $\psi_{1-\Delta t\to 1}$ provides direct near-terminal guidance.  

\begin{wrapfigure}{r}{0.4\columnwidth}
  \centering
  \vspace{-2.0em}
  \includegraphics[width=1.0\linewidth, trim=6pt 6pt 8pt 0pt, clip]{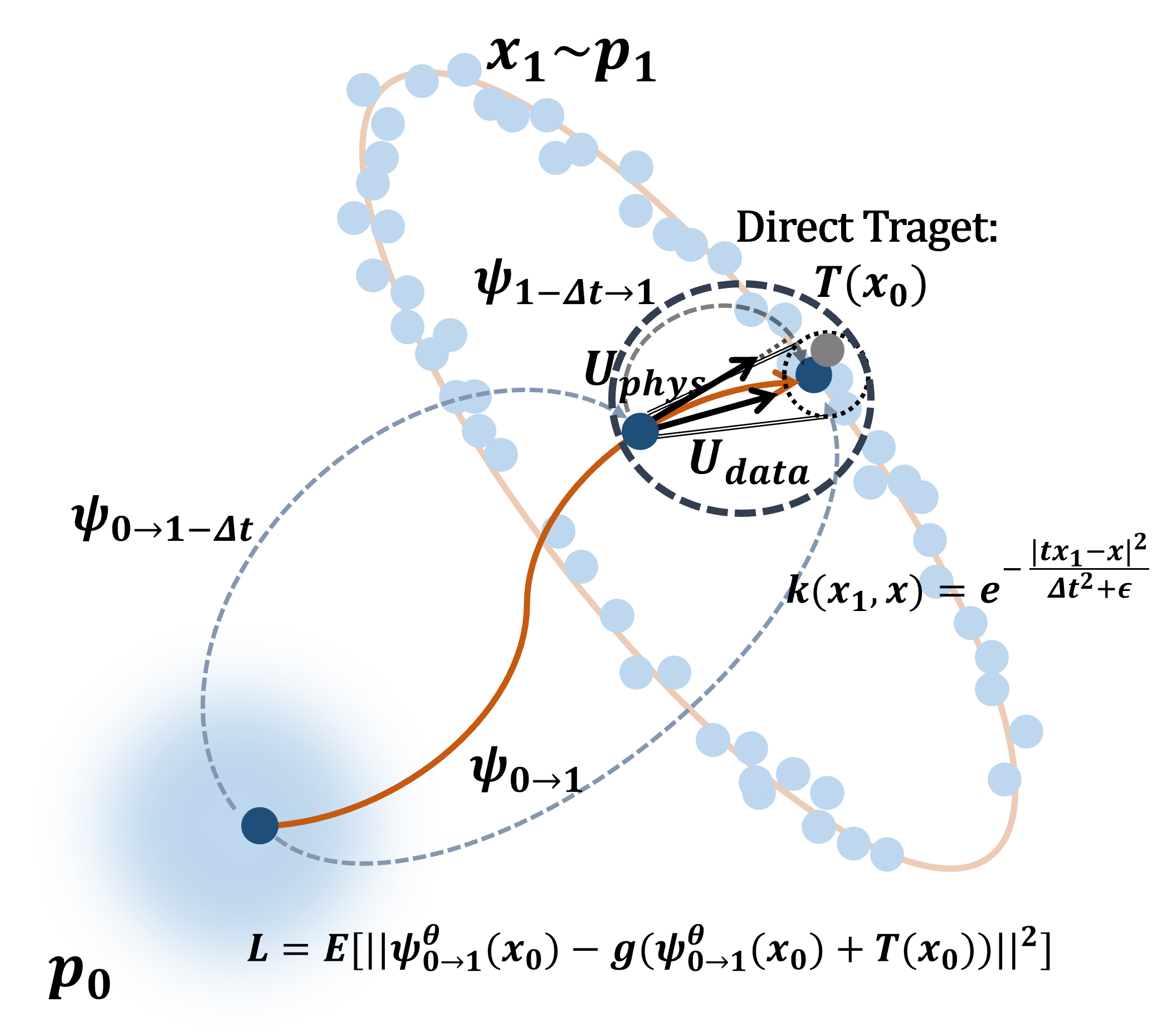}
  \caption{\textbf{Long-Short Flow Map.} The flow map $\psi_{0\to 1}$ is decomposed into a long flow map $\psi_{0\to 1-\Delta t}$ and a short flow map $\psi_{1-\Delta t\to 1}$. The short flow map is modeled through dynamics driven by the potentials $U_{\mathrm{phys}}$ and $U_{\mathrm{data}}$, providing guidance for the long flow map. As $\Delta t \to 0$, this becomes direct guidance for learning $\psi_{0\to1}$, integrating both data and physical information.}
  \label{fig:long_short_flow_map}
  \vspace{-8.0em}
\end{wrapfigure}
To let the short-range terminal flow map $\psi_{1-\Delta t\to 1}$ provide near-terminal direct guidance, we model its dynamics as jointly driven by a data-distribution potential and a physics-constraint potential. We compute this dynamics numerically as $\mathcal{I}[\frac{d}{dt}\psi_{1-\Delta t\to t}(x);\,1-\Delta t, 1, \psi^\theta_{0\to 1-\Delta t}(x_0)]$, where $\mathcal{I}[f(x,t);s,r,x_0]$ denotes numerical integration of the integrand $f(x,t)$ over $[s,r]$ with initial state $x_0$. As $\Delta t \to 0$, \autoref{eq:decomposition} leads to the following loss:
\begin{equation}\label{eq:overview_loss}
    \begin{aligned}
        \mathcal{L}(\theta)
        &\!=\!
        \mathbb{E}_{x_0\sim p_0}
        \|
        \psi_{0\to 1}^\theta(x_0)
        \!-\!
        \lim_{\Delta t\to 0}sg\!(\psi^\theta_{0\to 1-\Delta t}(x_0)\!\\       &+\!
        \mathcal{I}[\frac{d}{dt}\psi_{1\!-\!\Delta t\to t}(x);1\!-\!\Delta t, 1, \psi^\theta_{0\to 1\!-\!\Delta t}(x_0)]
        )
        \|^2\\
        &\!=\!
        \mathbb{E}_{x_0\sim p_0}
        \|
        \psi_{0\to 1}^\theta(x_0)
        \!-\!
        sg\!(\psi^\theta_{0\to 1}(x_0)\!\\
        &+\!\lim_{\Delta t\to 0}
        \mathcal{I}[\frac{d}{dt}\psi_{1\!-\!\Delta t\to t}(x);1\!-\!\Delta t, 1, \psi^\theta_{0\to 1\!-\!\Delta t}(x_0)]
        )
        \|^2
    \end{aligned}
\end{equation}
Here $sg$ denotes the stop-gradient operator. The short-range dynamics defines \textbf{a direct guidance target} for $\psi_{0\to 1}^\theta(x_0)$ as $\psi_{0\to 1}^\theta(x_0) + \mathcal{T}(x_0)$, where $\mathcal{T}(x_0)=\lim_{\Delta t\to 0} \mathcal{I}[\frac{d}{dt}\psi_{1-\Delta t\to t}(x);\,1-\Delta t, 1, \psi^\theta_{0\to 1-\Delta t}(x_0)]$. This avoids indirect objectives based on regression surrogates or intermediate-state reconstruction. When the limit exists, $\psi^\theta_{0\to 1-\Delta t}(x_0)$ inside $\mathcal{T}(x_0)$ converges to $\psi^\theta_{0\to 1}(x_0)$, so $\mathcal{L}(\theta)$ depends only on the single flow map $\psi^\theta_{0\to 1}(x_0)$ and is therefore self-consistent. Taking $\Delta t \to 0$ is also necessary to avoid Jensen's gap~\cite{bastek2024physics} and ensure well-defined kernel computation, as summarized in \autoref{sec:algorithm}.

We now specify the short-range dynamics $\psi_{1-\Delta t\to t}(x)$. We model it as a dynamics jointly driven by a data-distribution potential $U_{\mathrm{data}}$ and a physics-constraint potential $U_{\mathrm{phys}}$:
\begin{equation}\label{eq:short_dynamics}
\frac{d x_{\tau}}{d\tau}
=
-\nabla U_{\mathrm{data}}(x_{\tau},\tau)
-
\lambda\nabla U_{\mathrm{phys}}(x_{\tau}, \tau),
\qquad \tau\in[t,1].
\end{equation}
Here $\lambda$ is constant coefficient and $x_\tau = \psi_{1-\Delta t\to \tau}(x)$ . The data-distribution potential drives the short flow map toward the target data distribution $x_1 \sim p_1$, corresponding to the closed-form optimal Flow Matching solution in \autoref{eq:flow_matching}, whose explicit form is given in \autoref{sec:numerical_data_dynamics}. The physics-constraint potential drives the dynamics toward physical feasibility, i.e., $\mathcal{R}(x)=0$, and we instantiate it using a remaining-time-normalized constraint energy,
\vspace{-2mm}
{
\setlength{\belowdisplayskip}{-1pt}
\begin{equation}\label{eq:physics_potention_solution}
U_{\mathrm{phys}}(x,t)=\frac{1}{1-t}\|\mathcal{R}(x)\|^2.
\end{equation}
}
The factor $\frac{1}{1-t}$ increasingly emphasizes physical feasibility near the terminal time, guiding the short flow toward the constraint manifold $\mathcal{R}(x)=0$. Then, we numerically solve \autoref{eq:short_dynamics} and separately examine the data and physics terms in \autoref{sec:numerical_data_dynamics} and \autoref{sec:numerical_physics_dynamics}.

\vspace{-0.mm}
\subsection{Numerical Calculation For Data Term}\label{sec:numerical_data_dynamics}
\vspace{-0.mm}
We first present the expression of the data potential $U_{\mathrm{data}}$, which is consistent with the training objective in \autoref{eq:flow_matching} and admits the following potential-driven formulation.
\begin{theorem}[Potential-Driven System for Flow Matching]\label{thm:potential_close_form}
{\color{black}Assume $p_0=\mathcal N(0,I)$, independent of the data endpoint $x_1$, and that $p_1$ is a smooth distribution with $p_1(x)>0$.}  The Flow Matching dynamics with velocity $u_t$, learned in \autoref{eq:flow_matching}, can be interpreted as a dynamical system that starts from an initial state $x_0 \sim p_0$ and evolves under a closed-form potential.
\begin{equation}\label{eq:data_dynamics}
\frac{d}{dt}x_t = -\nabla {U_{\mathrm{data}}}(x,t),
\end{equation}
where ${U_{\mathrm{data}}}(x,t)$ is defined as
\begin{equation}\label{eq:closed_form_solution_U}
    \begin{aligned}
        {U_{\mathrm{data}}}(x,t) =-\frac{1-t}{t} \log \mathbb{E}_{x_1\sim p_1}[k_t(x_1,x)]-\frac{1}{2t}\|x\|^2 + C_t,\qquad k_{t}(y,x) &= e^{-\frac{\|ty-x\|^2}{2(1-t)^2}}
    \end{aligned}
\end{equation}
where $C_t$ is a function that depends only on $t$.  See \autoref{proof:closed_form} for proof.
\end{theorem}
\vspace{-2mm}
Using the closed-form solution of $U_{\mathrm{data}}(x,t)$ in \autoref{thm:potential_close_form}, we numerically solve \autoref{eq:data_dynamics} to compute the data-distribution component $\mathcal{T}_{\mathrm{data}}(x_0)$ of the target $\mathcal{T}(x_0)$ in \autoref{eq:overview_loss}.  In particular, we consider three standard time-discretization schemes, namely Forward Euler (FE), Backward Euler (BE), and the Trapezoidal Rule (TR), and then take the limit $\Delta t \to 0$, which yields the update formulas and targets shown in \autoref{tab:data_potential_discretization}.  To avoid singular behavior during computation of the limit, we mollify the kernel $k_t$ in \autoref{eq:closed_form_solution_U} as $k^\epsilon_{t}(y,x) = e^{-\frac{\|ty-x\|^2}{2(1-t)^2+\epsilon}}$.  Notably, during numerical integration, the time step $\Delta t$ cancels the factor $\frac{1}{1-t}$ in $\nabla U_{\mathrm{data}}$, making the limit well defined and computable.  Detailed derivations are given in \autoref{proof:data_tab}.
\begin{table}[H]
\vspace{-2mm}
\caption{Three standard discretizations of the data potential-driven dynamics in \autoref{eq:data_dynamics}.}
\vspace{-3mm}
\label{tab:data_potential_discretization}
\centering
\small
\setlength{\tabcolsep}{8pt}
\renewcommand{\arraystretch}{1.4}
\resizebox{\linewidth}{!}{%
\begin{tabular}{p{0.05\linewidth} p{0.29\linewidth} p{0.53\linewidth}}
\toprule
\textbf{Scheme} & \textbf{Update formula} & \textbf{Target} \\
\midrule

\begin{tabular}[t]{@{}l@{}}
FE
\end{tabular}
&
$
\begin{aligned}
&x_{n+1}
=
x_n-\Delta t\, \nabla U_{\mathrm{data}}(x_n,t_n)
\end{aligned}
$
&
$
\begin{aligned}
\mathcal{T}_{data}(x_0)&=\frac{\mathbb{E}_{x_1\sim p_1}[(x_1-\psi^{\theta}_{0\to 1}(x_0))k^\epsilon_{1}(x_1,\psi^{\theta}_{0\to 1}(x_0))]}{\mathbb{E}_{x_1\sim p_1}[k^\epsilon_{1}(x_1,\psi^{\theta}_{0\to 1}(x_0))]}
\end{aligned}
$
\\

\begin{tabular}[t]{@{}l@{}}
BE
\end{tabular}
&
$
\begin{aligned}
&x_{n+1}
=
x_n\\
&-\Delta t \nabla U_{\mathrm{data}}(x_{n+1},t_{n+1})
\end{aligned}
$
&
$
\begin{aligned}
\mathcal{T}_{data}(x_0)&=\frac{\mathbb{E}_{\tilde x_{1}\sim [\psi_{0\to 1}^\theta]_\sharp p_0}[(\psi_{0 \to 1}^\theta(x_0)-\tilde x_{1})k^\epsilon_{1}(\psi_{0 \to 1}^\theta(x_0),\tilde x_{1})]}{\mathbb{E}_{\tilde x_{1}\sim [\psi_{0\to 1}^\theta]_\sharp p_0}[k^\epsilon_{1}(\psi_{0 \to 1}^\theta(x_0),\tilde x_{1})]}
\end{aligned}
$
\\

\begin{tabular}[t]{@{}l@{}}
TR
\end{tabular}
&
$
\begin{aligned}
&x_{n+1}
=
x_n
-\frac{\Delta t}{2}
(
\nabla U_{\mathrm{data}}(x_n,t_n)
\\&+
\nabla U_{\mathrm{data}}(x_{n+1},t_{n+1})
)
\end{aligned}
$
&
$
\begin{aligned}
\mathcal{T}_{data}(x_0)&= \frac{1}{2}\frac{\mathbb{E}_{x_1\sim p_1}[(x_1-\psi^{\theta}_{0\to 1}(x_0))k^\epsilon_{1}(x_1,\psi^{\theta}_{0\to 1}(x_0))]}{\mathbb{E}_{x_1\sim p_1}[k^\epsilon_{1}(x_1,\psi^{\theta}_{0\to 1}(x_0))]}\\
&+\frac{1}{2}\frac{\mathbb{E}_{\tilde x_{1}\sim [\psi_{0\to 1}^\theta]_\sharp p_0}[(\psi_{0 \to 1}^\theta(x_0)-\tilde x_{1})k^\epsilon_{1}(\psi_{0 \to 1}^\theta(x_0),\tilde x_{1})]}{\mathbb{E}_{\tilde x_{1}\sim [\psi_{0\to 1}^\theta]_\sharp p_0}[k^\epsilon_{1}(\psi_{0 \to 1}^\theta(x_0),\tilde x_{1})]}
\end{aligned}
$
\\
\bottomrule
\end{tabular}%
}
\vspace{-5mm}
\end{table}
Here, $k_1(y,x) = \lim_{t\to 1} k_t^\epsilon(y,x) = e^{-\frac{\|y-x\|^2}{\epsilon}}$.  In \autoref{sec:darcy}, we compare different numerical schemes and their impact on training performance. The results indicate that the computation of the data term is highly sensitive to the order of discretization, with the second-order Trapezoidal Rule showing a significant advantage over the lower-order alternatives.
\vspace{-0.0mm}
\subsection{Numerical Calculation For Physics Term}\label{sec:numerical_physics_dynamics}
\vspace{-0.0mm}


Using $U_{\mathrm{phys}}$ from \autoref{eq:physics_potention_solution}, we derive the physics-driven dynamics $\frac{d}{dt}x_t = -\lambda \nabla U_{\mathrm{phys}}(x,t)$ under three schemes: Forward Euler (FE), Backward Euler (BE), and Backward Euler with $\lambda \to \infty$ (BE-Lim). As in \autoref{sec:numerical_data_dynamics}, we integrate over the terminal interval and take $\Delta t \to 0$ to obtain the physics component $\mathcal{T}_{\mathrm{phys}}(x_0)$ of $\mathcal{T}(x_0)$ in \autoref{eq:overview_loss}. The resulting updates and targets are summarized in \autoref{tab:physics_potential_discretization}, with derivations in \autoref{proof:physics_tab}.


\begin{table}[H]
\vspace{-2mm}
\caption{Three standard discretizations of the Physics potential-driven dynamics.}
\vspace{-3mm}
\label{tab:physics_potential_discretization}
\centering
\small
\begin{threeparttable}
\setlength{\tabcolsep}{8pt}
\renewcommand{\arraystretch}{1.4}
\begin{tabular}{p{0.08\linewidth} p{0.35\linewidth} p{0.45\linewidth}}
\toprule
\textbf{Scheme} & \textbf{Update formula} & \textbf{Target} \\
\midrule

FE
&
$
\begin{aligned}
x_{n+1}
=
x_n-\Delta t \lambda \nabla U_{\mathrm{phys}}(x_n,t_n)
\end{aligned}
$
&
$
\begin{aligned}
\mathcal{T}_{phys}(x_0)&= -\lambda J_{R}^\top(\psi_{0\to 1}^\theta(x_0))R(\psi_{0\to 1}^\theta(x_0))
\end{aligned}
$
\\

BE
&
$
\begin{aligned}
x_{n+1}
=
x_n
-\Delta t \lambda  \nabla U_{\mathrm{phys}}(x_{n+1},t_{n+1})
\end{aligned}
$
&
$
\begin{aligned}
\mathcal{T}_{phys}(x_0)&=-(J_{R}^\top(\psi_{0\to 1}^\theta(x_0))J_{R}(\psi_{0\to 1}^\theta(x_0)) + \gamma E)^{-1}\\
&J_{R}^\top(\psi_{0\to 1}^\theta(x_0))R(\psi_{0\to 1}^\theta(x_0))
\end{aligned}
$
\\

BE-Lim
&
$
\begin{aligned}
&x_{n+1}
=
x_n
-\Delta t \lambda  \nabla U_{\mathrm{phys}}(x_{n+1},t_{n+1}),\\
&\lambda\to \infty
\end{aligned}
$
&
$
\begin{aligned}
\mathcal{T}_{phys}(x_0)&= -J_{R}^\top(\psi_{0\to 1}^\theta(x_0))(J_{R}(\psi_{0\to 1}^\theta(x_0))\\
&J^\top_{R}(\psi_{0\to 1}^\theta(x_0)))^{-1}R(\psi_{0\to 1}^\theta(x_0))
\end{aligned}
$
\\
\bottomrule
\end{tabular}%
\begin{tablenotes}
\footnotesize
\item $J_R$ denotes the Jacobian of $R$, $E$ denotes the identity matrix, and $\gamma=\frac{1}{2\lambda}$.
\end{tablenotes}
\end{threeparttable}
\vspace{-5mm}
\end{table}
Unlike the data term, we consider only first-order schemes, which already perform well in \autoref{sec:darcy}. We additionally include the $\lambda\to\infty$ Backward Euler limit due to its connection to classical position-based dynamics (\autoref{appendix:PBD}). Formally, Forward Euler corresponds to direct gradient evaluation, whereas Backward Euler gives a Gauss-Newton-type gradient computation.

\vspace{-1.5mm}
\subsection{Algorithm and Analysis}\label{sec:algorithm}
\vspace{-1.5mm}
\begin{wraptable}{r}{0.52\linewidth}
\vspace{-3.3em}
\centering
\small
\begin{minipage}{1.0\linewidth}
\hrule
\vspace{0.3em}
\refstepcounter{algorithm}
\textbf{Algorithm \thealgorithm} PI-LSFM Training
\label{alg:training}
\vspace{0.3em}
\hrule
\vspace{0.4em}
\begin{algorithmic}[1]
\Require dataset $\mathcal{D}$, batch size $B$, initial model parameters $\theta$, learning rate $\eta$, hyperparameter $\bar{\lambda}$
\Repeat
    \State Sample $\mathcal{X}_1 = \{x_1^j\}_{j=1}^B$ from the dataset $\mathcal{D}$
    \State Sample noise $\mathcal{X}_0 = \{x_0^i\}_{i=1}^B$ from ${\color{black}\mathcal{N}(0,I)}$
    \State $x^{i,\theta}_1 \gets \psi_{0\to 1}^\theta(x_0^i)$ and $\mathcal{X}^\theta_1 = \{x_1^{i,\theta}\}_{i=1}^B$
    \State Compute $\{\mathcal{T}_{\mathrm{data}}(x_0^i)\}_{i=1}^B$ from $\mathcal{X}^\theta_1$ and $\mathcal{X}_1$ using tab. \ref{tab:data_potential_discretization} and the expectation computation in eq.\ref{eq:expectation_calculation}
    \State Compute $\{\mathcal{T}_{\mathrm{phys}}(x_0^i)\}_{i=1}^B$ from $\mathcal{X}^\theta_1$ using tab. \ref{tab:physics_potential_discretization}
    \State $\mathcal{T}(x_0^i) \gets \mathcal{T}_{data}(x_0^i) + \bar\lambda \mathcal{T}_{phys}(x_0^i)$ 
    \State $\mathcal{L}(\theta) \gets \frac{1}{B}\sum_{i=1}^B\|x^{i,\theta}_1-sg(x^{i,\theta}_1+\mathcal{T}(x_0^i))\|^2$  
    \State $\theta \gets \theta - \eta \nabla_\theta \mathcal{L}(\theta)$
\Until{convergence}
\end{algorithmic}
\vspace{0.5em}
\hrule
\end{minipage}
\end{wraptable}

Combining $\mathcal{T}_{\mathrm{data}}(x_0)$ from \autoref{sec:numerical_data_dynamics} and $\mathcal{T}_{\mathrm{phys}}(x_0)$ from \autoref{sec:numerical_physics_dynamics}, we obtain the final target for the short-flow dynamics,
\vspace{-1.5mm}
{
\setlength{\belowdisplayskip}{0pt}
\begin{equation}
    \begin{aligned}
    \mathcal{T}(x_0) = \mathcal{T}_{\mathrm{data}}(x_0) + \bar{\lambda}\,\mathcal{T}_{\mathrm{phys}}(x_0),
    \end{aligned}
\end{equation}
}
\noindent where $\bar{\lambda}$ is a tunable hyperparameter introduced to unify different numerical schemes. This helps give the concrete loss form of \autoref{eq:overview_loss}. Here, $\mathcal{T}_{\mathrm{data}}$ can be chosen from the three forms in \autoref{tab:data_potential_discretization}, while $\mathcal{T}_{\mathrm{phys}}$ can be chosen from the three forms in \autoref{tab:physics_potential_discretization}, yielding different combinations.  Our training algorithm based on this direct-target loss is shown in \autoref{alg:training}. After training, $\psi_{0\to 1}^\theta$ is encouraged to generate $\psi_{0\to 1}^\theta(x_0)$ that match the target distribution and satisfy the physical constraints.


\noindent\textbf{Expectation computation\qquad} The formula of $\mathcal{T}_{\mathrm{data}}(x_0)$ in \autoref{tab:data_potential_discretization} involves expectations. In practice, we approximate them using samples within the same batch in \autoref{alg:training}. Specifically, for each $\mathcal{T}(x_0^i)$, the expectations $\mathbb{E}_{x_1\sim p_1}[F(x_1,x_0^i)]$ and $\mathbb{E}_{\bar{x}_1\sim[\psi^\theta_{0\to1}]_\sharp p_0}[F(\bar{x}_1,x_0^i)]$ are estimated using $\mathcal{X}_1$ and $\mathcal{X}_1^\theta$, respectively, where $F$ is an arbitrary function:
\vspace{-3mm}
{
\setlength{\belowdisplayskip}{3pt}
\begin{equation}\label{eq:expectation_calculation}
    \begin{aligned}
        \mathbb{E}_{x_1\sim p_1}[F(x_1,x_0^i)] = \frac{1}{B}\sum_{j=1}^B F(x^j_1,x_0^i),\quad \mathbb{E}_{\bar{x}_1\sim[\psi^\theta_{0\to1}]_\sharp p_0}[F(\bar{x}_1,x_0^i)] = \frac{1}{B} \sum_{j=1}^B F( x^{j,\theta}_1,x_0^i)
    \end{aligned}
\end{equation}
}
This computation is fully parallelizable; the PyTorch implementation is given in \autoref{appedix:code}.

\noindent\textbf{Conditional generation\qquad} For conditional generation, the learned map becomes $\psi^\theta_{0\to1}(\cdot| c)$, and the corresponding expectations in \autoref{tab:data_potential_discretization} are taken under $p_1(\cdot| c)$ and $[\psi^\theta_{0\to1}(\cdot| c)]_\sharp p_0$ (see \autoref{appedix:conditional_generation}). Accordingly, \autoref{alg:training} only requires a different grouping strategy: we sample $B_c$ conditions, sample $B_s$ data samples for each condition, and sample $B=B_cB_s$ noise samples. The update in \autoref{alg:training} is then computed within each condition group and averaged over all $B_c$ conditions. The detailed algorithm is given in \autoref{alg:conditional_training}. This supports both label-conditioned generation (\autoref{sec:kolmogorov}) and generation conditioned on complex fields (\autoref{sec:topology}).

\noindent\textbf{Necessity of the $\Delta t\to 0$ limit\qquad} Taking the limit $\Delta t\to0$ in \autoref{eq:overview_loss} is necessary for three reasons. First, it turns $\psi^\theta_{0\to1-\Delta t}$ into $\psi^\theta_{0\to1}$, so that the loss depends on a single unknown flow map and becomes self-consistent. Second, it makes the kernel $k_t^\epsilon(x,y)$ sufficiently sharp near $t=1$, allowing accurate estimation of the $k_t^\epsilon(x,y)$-weighted expectation in \autoref{tab:data_potential_discretization} with a moderate batch size.  Third, it applies $\mathcal{R}(\cdot)$ directly to $\psi^\theta_{0\to1}(x_0)$ rather than to the intermediate state $\psi^\theta_{0\to1-\Delta t}(x_0)$ for \autoref{tab:physics_potential_discretization}, avoiding the Jensen-type mismatch from intermediate residual enforcement.

\begin{figure}[t]
\centering
\begin{subfigure}[t]{0.50\linewidth}
\vspace{25pt}
    \centering
    \includegraphics[width=\linewidth]{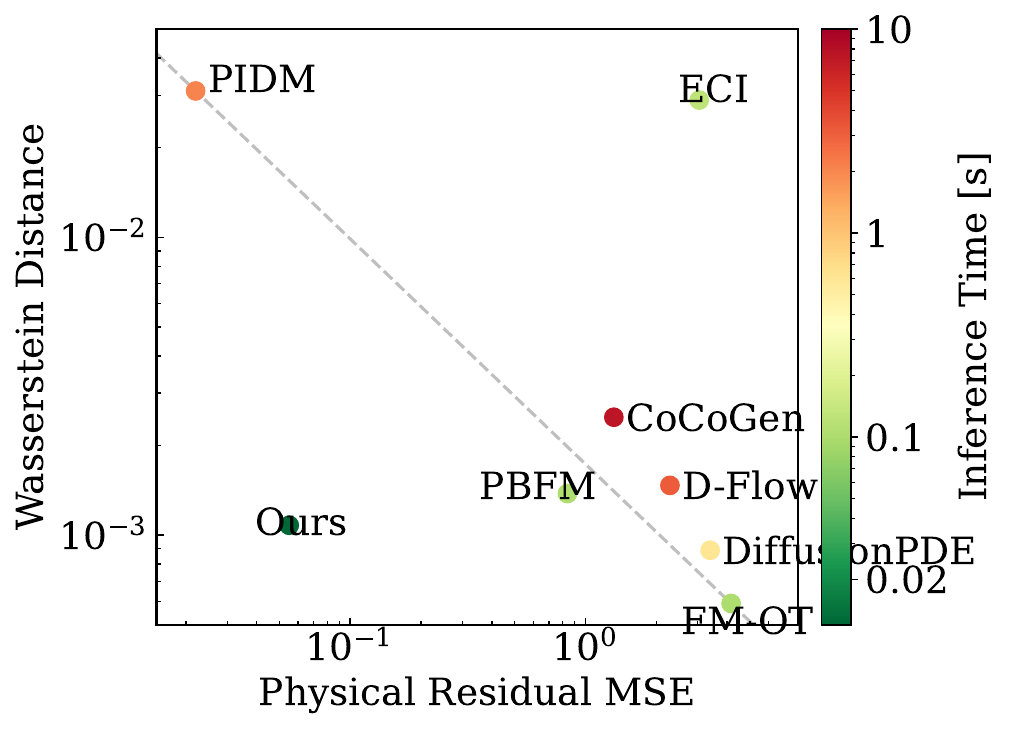}
    \caption{Pareto front of physical residual MSE vs.\ Wasserstein distance for Darcy flow for the proposed methods.}
    \label{fig:pareto_darcy}
\end{subfigure}
\hfill
\begin{minipage}[t]{0.48\linewidth}
\centering
\small
\vspace{-0.5em}
\begin{subtable}[t]{\linewidth}
\vspace{-0.5mm}
\caption{Feature extractors (100K training steps)}
\label{tab:feature_extractor}
\vspace{-1mm}
\centering
\resizebox{\linewidth}{!}{%
\begin{tabular}{lccccc}
\toprule
Extractor & Best $\bar\lambda$ & RE & WD $\cdot 10^2$ & JS $\cdot 10^1$ & Sec/Step \\
\midrule
MAE  & 3.0 & 0.117 & 0.213 & 0.216 & 0.271 \\
Poseidon  & 1.0 & 0.410 & 0.238 & 0.271 & 0.834 \\
DPOT  & 10.0 & 0.596 & 0.071 & 0.126 & 0.65 \\
MPP  & 10.0 & 0.109 & 1.027 & 0.694 & 0.652 \\
\bottomrule
\end{tabular}%
}
\end{subtable}

\begin{subtable}[t]{\linewidth}
\vspace{-1mm}
\caption{Physics term methods ({\color{black}400K} training steps)}
\vspace{-0.5mm}
\label{tab:physics_term_methods_comparison}
\centering
\begin{threeparttable}
\resizebox{1.1\linewidth}{!}{%
\begin{tabular}{lccccc}
\toprule
Method & Best $\bar\lambda$ & RE & WD $\cdot 10^2$ & JS $\cdot 10^1$ & Sec/Step \\
\midrule
FE & 3.0 & 0.055 & 0.108 & 0.171 & 0.271 \\
BE & piecewise & 0.436 & 0.122 & 0.159 & 0.388 \\
BE-Lim. & piecewise & 2.540 & 0.184 & 0.192 & 0.376 \\
\bottomrule
\end{tabular}%
}
\begin{tablenotes}
\footnotesize 
\item[$\dagger$]  ``piecewise'' denotes the $\bar\lambda$ scheduler described in \autoref{appendix:physics_design_study}.
\end{tablenotes}
\end{threeparttable}
\end{subtable}

\begin{subtable}[t]{\linewidth}
\vspace{-1mm}
\caption{Data term methods ({\color{black}100K} training steps)}
\vspace{-0.5mm}
\label{tab:data_term_methods_comparison}
\centering
\resizebox{\linewidth}{!}{%
\begin{tabular}{lccccc}
\toprule
Method & Best $\bar\lambda$ & RE & WD $\cdot 10^2$ & JS $\cdot 10^1$ & Sec/Step \\
\midrule
FE & 3.0 & 8.125 & 4.244 & 5.846 & 0.268 \\
BE & 10.0 & 1.853 & 4.857 & 7.462 & 0.267 \\
TR & 3.0 & 0.117  & 0.213 & 0.216&0.271  \\
\bottomrule
\end{tabular}%
}
\end{subtable}
\end{minipage}

\begin{subtable}[t]{\linewidth}
\centering
\vspace{-1mm}
\caption{Comparison of different methods in terms of RE, WD, JS, and NFE. \textbf{Best} and \underline{second-best} results are highlighted. Our model uses the same architecture and number of parameters as PIDM, training with 400K steps}
\label{tab:comprison_darcy}
\begin{threeparttable}
\resizebox{\linewidth}{!}{%
\begin{tabular}{lccccccccc}
\hline
Metric & Ours &  sCM-PINN & PBFM & FM-OT & CoCoGen$\dagger$ & PIDM$\dagger$ & DiffusionPDE & D-Flow$\ddagger$ & ECI$\S$ \\
\hline
RE & \underline{0.055} & 0.107 & 0.838 & 4.159 & 1.320 & \textbf{0.022} & 3.388 & 2.286 & 3.045 \\
WD $\cdot 10^2$ & 0.108 & 6.453 & 0.138 & \textbf{0.059} & 0.249 & 3.103 & \underline{0.089} & 0.147 & 2.892 \\
JS $\cdot 10^1$ & 0.171 & 7.401 & 0.256 & \textbf{0.131} & 0.360 & 3.179 & \underline{0.139} & 0.237 & 2.818 \\
NFE & \textbf{1} & \textbf{1} & 20 & 20 & 100 & 100 & 20 & 20 & 20 \\
IT [s] & \textbf{0.012} & 0.025 & 0.101 & 0.100 & 7.395 & 2.050 & 0.590 & 3.126 & 0.122 \\
\hline
\end{tabular}%
}
\begin{tablenotes}
\footnotesize
\item[$\dagger$]  Our method and the reproduced sCM-PINN share the same U-Net architecture. By removing time-conditioning and \\modulation (since $t$ is not required for our formulation), our method achieves faster inference.
\end{tablenotes}
\end{threeparttable}
\end{subtable}

\vspace{-2mm}
\caption{Comparison of design choices and overall performance in our framework. (a) Pareto front of physical residual MSE vs.\ Wasserstein distance for Darcy flow. (b)--(d) compare different feature extractors, numerical discretizations for the physics term and the data term respectively. Training time per step (s) is also reported in these comparisons. More details are provided in \autoref{appendix:design_study}. (e) Overall comparison with competing methods on RE, WD, JS, NFE, and inference time.}
\label{fig:darcy_ablation_all}
\vspace{-5mm}
\end{figure}
\vspace{-3mm}
\paragraph{Stabilization}
Computing the kernel $k_t^\epsilon(x,y)$ in high-dimensional $\mathcal{X}$ may suffer from the curse of dimensionality and destabilize optimization. Following \cite{deng2026generative}, we therefore perform the computation in feature spaces $\mathcal{Z}=\phi(X)$, where $\phi$ denotes an encoder. The resulting loss is
\vspace{-1mm}
{
\setlength{\belowdisplayskip}{-2pt}
\begin{equation}\label{eq:feature_loss}
    \begin{aligned}
    \mathcal{L}(\theta) = \mathbb{E}_{x_0\sim p_0}[\|\phi(\psi_{0\to 1}^\theta(x_0)) - sg(\phi(\psi_{0\to 1}^\theta(x_0) + \bar\lambda \mathcal{T}_{phys}(x_0))+ \mathcal{T}^{\mathcal{Z}}_{data}(x_0))\|^2]
    \end{aligned}
\end{equation}
}
where $\mathcal{T}_{\mathrm{data}}^{\phi}(\cdot)$ denotes the corresponding data term computed in the feature space $\mathcal{Z}$.  In \autoref{sec:darcy}, we compare different encoder choices, including several physics foundation models. The encoder is used only to stabilize the kernel-weighted estimation of the data guidance target in high-dimensional field spaces. Without the encoder, Darcy flow still achieves reasonable physical residuals and only slightly poorer distribution matching (RE=0.044, WD$\cdot 10^2$ = 0.446, and JS$\cdot 10^1$ = 0.889), but becomes more prone to training collapse, as illustrated in \autoref{fig:darcy_no_feature_encoder}. {\color{black}Figure~\ref{tab:feature_extractor}} shows that different encoders yield competitive feature-space training performance, with further comparisons in \autoref{appendix:encoder_study}. Thus, the feature encoder is a stabilization device rather than an inherent component of PI-LSFM. We use it only for the weakly conditioned Darcy and Kolmogorov flow benchmarks (Sections~\ref{sec:darcy} and~\ref{sec:kolmogorov}), while dynamic stall (Section~\ref{sec:kolmogorov}) and topology optimization (Section~\ref{sec:topology}) are trained directly without any encoder.  Although MAE roughly doubles the per-step training time (0.271s vs. 0.132s without an encoder), PI-LSFM uses about one third of the training steps of~\cite{baldan2026physics}, largely offsetting this overhead. Unlike latent-based generation, PI-LSFM still generates $\psi_{0\to1}^\theta(x_0)$ in the original space and computes only the loss in feature space; see \autoref{appedix:feature_space} and \autoref{alg:feature_space_training}.
\vspace{-2mm}
\section{Experiments}
\vspace{-2mm}

{\color{black}We choose the four benchmarks to progressively evaluate unconditional generation, conditional generalization, coupled multi-field physical constraints, and physics-constrained inverse design.} We use the U-Net architecture of~\cite{bastek2024physics} without architecture-specific optimization, and match parameter counts whenever possible. For fair comparison, we report NFE and wall-clock inference time when available, using standard sampling settings or reported numbers for multi-step baselines. Since multi-step samplers benefit from iterative refinement, we target a favorable accuracy-efficiency trade-off rather than uniformly dominating every metric. All experiments use an NVIDIA A100 GPU. {\color{black}Our main benchmark results are obtained from single training runs with the fixed seed 42 and no seed selection. Appendix~\ref{app:add_seeds}, Table~\ref{tab:add_seeds} reports a three-seed, 100K-step Darcy stability check.}

\vspace{-1.5mm}
\subsection{Darcy Flow and Design Studies}\label{sec:darcy}
\vspace{-1.5mm}
Darcy flow is a two-dimensional steady-state porous-media benchmark on $64\times64$ pressure $p$ and permeability $K$ fields, with 10k training and 1k validation samples. Following \cite{baldan2026physics}, we enforce the Darcy residual
$\mathcal{R}_{\mathrm{pde}}=\nabla\cdot(K\nabla p)+f=0$
together with the boundary-condition loss, and evaluate physical validity and distributional fidelity using residual error (RE), Wasserstein distance (WD), and Jensen--Shannon divergence (JS) of pressure field. This is an unconditional generation task, and lower values indicate better performance.


On Darcy flow, we compare with multi-step baselines PBFM \cite{baldan2026physics}, FM-OT \cite{lipman2023flow}, CoCoGen \cite{jacobsen2025cocogen}, PIDM \cite{bastek2024physics}, DiffusionPDE \cite{huang2024diffusionpde}, D-Flow \cite{ben2024d}, and ECI \cite{cheng2024gradient}, using reported numbers from \cite{baldan2026physics}, as well as our one-step flow-map baselines built on sCM-PINN \cite{chang2026stabilizing}. As shown in {\color{black}Figure~\ref{tab:comprison_darcy}} and \autoref{fig:pareto_darcy}, our method significantly outperforms other one-step methods, matches strong multi-step baselines, and lies on the best Pareto front, while using fewer training steps with the same batch size and roughly matched parameter count to the PBFM baseline. At inference, its one-step sampling yields substantial acceleration, including about a $5\times$ speedup over architecture-optimized PBFM under a standard backbone.  Additional experimental details, including the training procedure, evaluation metrics, and qualitative visualizations, are given in \autoref{appedix:darcy_flow}.


To stabilize training, we adopt the feature-space stabilization strategy described in \autoref{sec:algorithm}. We study different encoder choices, including a multi-resolution MAE trained on this dataset \cite{he2022masked}, the pretrained multi-resolution U-Net-based physics foundation model Poseidon \cite{herde2024poseidon}, the transformer-based physics foundation model DPOT \cite{hao2024dpot}, and MPP \cite{mccabe2024multiple}. The results show that MAE, Poseidon, and DPOT achieve comparable performance, with MAE providing the strongest overall results and the fastest training speed. We therefore use MAE as the encoder for Darcy and Kolmogorov flow (Sections~\ref{sec:darcy} and~\ref{sec:kolmogorov}).  In {\color{black}Figure~\ref{tab:data_term_methods_comparison}} and {\color{black}Figure~\ref{tab:physics_term_methods_comparison}}, we further compare different numerical schemes for the data term and the physics term, respectively, while fixing the other term to its best-performing choice. The results show that the second-order trapezoidal rule works best for the data term, while Forward Euler performs best for the physics term. A more complete design study and parameter study, including the $\bar\lambda$ schedule, is provided in \autoref{appendix:design_study}.


\begin{table*}[t]
\centering
\vspace{-1mm}
\caption{Generative performance metrics for Kolmogorov flow and dynamic stall problems. RE: physical Residual MSE, WD: Wasserstein Distance, JS: Jensen-Shannon divergence, MMSE: MSE of the mean fields, SMSE: MSE of the standard deviation fields, NFE: Number of function evaluations, IT: inference wall-clock time. \textbf{Best} and \underline{second-best} results are highlighted.}
\label{tab:gen_metrics_kolmogorov_dynamicstall}
\vspace{-2mm}
\begin{threeparttable}
\resizebox{0.8\textwidth}{!}{%
\begin{tabular}{llcccccc}
\toprule
Dataset & Metric & Ours & PBFM & OT-FM & DiffusionPDE & D-Flow$^{\dagger}$ & PCFM$^{\dagger}$ \\
\midrule
\multirow{7}{*}{\shortstack{Kolmogorov\\Flow}}
& RE $\cdot 10^1$    & \textbf{0.404} & \underline{1.362} & 2.314 & {1.930} & -    & - \\
& WD $\cdot 10^1$    & \textbf{0.140} & \underline{1.222} & {2.124} & 3.698 & -    & - \\
& JS $\cdot 10^2$    & \textbf{1.621} & \underline{7.440} & {12.49} & 19.39 & -    & - \\
& MMSE $\cdot 10^4$  & 25.83 & \underline{4.455} & \textbf{4.188} & 5.669 & -    & - \\
& SMSE $\cdot 10^4$  & \textbf{1.765} & \underline{2.484} & {2.574} & 21.82 & -    & - \\
& NFE                & \textbf{1} & 20 & 20 & 20 & 20 & 20 \\
& IT $[ms]$          & \textbf{27.25} & {98.97} & \underline{98.75} & 267.8 & 6431 & - \\
\midrule
\multirow{7}{*}{\shortstack{Dynamic\\Stall}}
& RE $\cdot 10^6$    & \textbf{0.00015} & {0.339} & 11.02 & 12.20 & 11.32 & \underline{0.143} \\
& WD $\cdot 10^4$    & \textbf{1.633} & \underline{1.814} & 2.707 & {2.509} & 3.484 & 4.013 \\
& JS $\cdot 10^2$    & \textbf{0.403} & \underline{0.680} & {0.983} & 1.029 & 1.014 & 1.206 \\
& MMSE $\cdot 10^5$  & 5.218 & \textbf{1.490} & 2.791 & 2.626 & \underline{2.507} & 5.669 \\
& SMSE $\cdot 10^5$  & 4.894 & \textbf{0.874} & 1.458 & \underline{1.236} & 1.372 & 7.674 \\
& NFE                & \textbf{1} & 20 & 20 & 20 & 20 & 20 \\
& IT $[ms]$          & \textbf{21.53} & 60.47 & \underline{59.75} & 171.7 & 138.9 & 3906 \\
\bottomrule
\end{tabular}%
}
\begin{tablenotes}
\scriptsize
\item[$\dagger$] D-Flow and PCFM are excluded from Kolmogorov-flow evaluation due to unstable generation and prohibitive cost.
\end{tablenotes}
\end{threeparttable}
\vspace{-7.5mm}
\end{table*}

\begingroup
\begin{wraptable}[13]{r}{0.40\linewidth}
\vspace{-5mm}
\begingroup
\color{black}\centering\footnotesize
\captionsetup{font={footnotesize,color=black},labelfont={footnotesize,color=black}}
\setlength{\tabcolsep}{2pt}
\caption{Additional one-step Darcy comparisons. All methods use NFE $=1$; IT is inference time.}
\vspace{-4mm}
\label{tab:add_onestep}
\resizebox{\linewidth}{!}{%
\begin{tabular}{lrrrr}
\toprule
Method & RE $\downarrow$ & WD$\cdot10^2$ $\downarrow$ & JS$\cdot10^1$ $\downarrow$ & IT [s] $\downarrow$\\
\midrule
PI-Consistency & 0.240 & 3.889 & 5.023 & 0.025\\
PI-MeanFlow & 4603.573 & 4.243 & 5.457 & 0.024\\
PI-Shortcut & 0.435 & 2.336 & 2.518 & 0.024\\
PI-LSFM & 0.055 & 0.108 & 0.171 & 0.012\\
\bottomrule
\end{tabular}%
}
\endgroup
\end{wraptable}

{\color{black}On Darcy, we compare PI-Consistency (sCM), PI-MeanFlow, and PI-Shortcut with the same terminal physics loss under matched settings. PI-LSFM improves RE, WD, and JS at NFE $=1$ (Table~\ref{tab:add_onestep}). Together with the toy results (Figure~\ref{fig:toy_3x3}, Table~\ref{tab:toy_constraint_satisfaction}), these comparisons support the value of direct data and physics guidance for one-step training. Further evidence is provided by the conflict-free controls in Appendix~\ref{app:add_baselines}, which show that conflict-free updates alone do not close the performance gap.\par}

\par\WFclear
\endgroup

\begingroup
\begin{wraptable}{r}{0.25\linewidth}
\vspace{-12mm}
\begingroup\color{black}\centering
\captionsetup{font={footnotesize,color=black},labelfont={footnotesize,color=black}}
\centering\footnotesize\setlength{\tabcolsep}{3pt}
\caption{Training time per generator-training step on Darcy.}
\label{tab:add_traincost}
\vspace{-3mm}
\resizebox{\linewidth}{!}{%
\begin{tabular}{lr}
\toprule
Method & Seconds/step\\
\midrule
PBFM & 0.566\\
PI-Consistency & 0.223\\
PI-MeanFlow & 0.264\\
PI-Shortcut & 0.328\\
PI-LSFM & 0.308\\
\bottomrule
\end{tabular}%
}
\endgroup
\end{wraptable}

{\color{black}Table~\ref{tab:add_traincost} compares training time on an A100. PI-LSFM costs 0.308 s per generator-training step, including the full MAE pretraining cost amortized over training (0.037 s for MAE plus 0.271 s for the generator; 400K steps each), versus 0.566 s for PBFM. PI-LSFM achieves the reported Darcy results after 400K generator updates, about one third of the 1,248K training iterations reported for PBFM~\cite{baldan2026physics}. The encoder adds no inference-time cost.\par}
\par
\ifnum\value{WF@wrappedlines}>1
  \vspace{\dimexpr\value{WF@wrappedlines}\baselineskip-\baselineskip\relax}
\fi
\WFclear
\endgroup

\vspace{-7.5mm}
\Needspace{8\baselineskip}
\subsection{Kolmogorov Flow and Dynamic Stall}\label{sec:kolmogorov}
\vspace{-1.5mm}
{\color{black}We evaluate conditional generation on two-dimensional incompressible Kolmogorov flow following~\cite{baldan2026physics}. The model is conditioned on the Reynolds number and evaluated on 16 unseen Reynolds numbers, testing generalization across turbulent flow regimes. Physical consistency is enforced through the incompressibility constraint $\nabla\cdot U=0$.  As shown in Table~\ref{tab:gen_metrics_kolmogorov_dynamicstall}, PI-LSFM achieves strong physical and distributional performance while requiring only a single function evaluation, compared with 20-step baselines. We report physical residual error (RE), distributional metrics (WD and JS), conditional-statistics errors (MMSE and SMSE), and inference efficiency (NFE and IT). Full dataset, metric, and implementation details are provided in Appendix~\ref{appedix:kolmogorov_flow}.}

{\color{black}We further evaluate PI-LSFM on conditional generation for dynamic stall, a compressible-flow benchmark around a sinusoidally pitching NACA0012 airfoil following~\cite{baldan2026physics}. The model predicts six coupled aerodynamic and thermodynamic fields conditioned on four operating and motion parameters, and is evaluated on unseen configurations. Physical consistency is enforced through the ideal-gas relation and a temperature-dependent skin-friction constraint based on Sutherland's law.  As shown in Table~\ref{tab:gen_metrics_kolmogorov_dynamicstall}, PI-LSFM achieves competitive performance relative to multi-step baselines while requiring only one function evaluation. The evaluation follows the same distributional, statistical, and efficiency metrics as Kolmogorov flow. Full problem formulation, residual definitions, training details, and qualitative results are provided in Appendix~\ref{appedix:dynamic_stalling}.}

\vspace{-1.5mm}
\subsection{Topology Optimization}\label{sec:topology}
\vspace{-1.5mm}
\begingroup
\begin{wraptable}[15]{r}{0.24\linewidth}
\vspace{-5mm}
\begingroup\color{black}\centering\footnotesize
\captionsetup{font={footnotesize,color=black},labelfont={footnotesize,color=black}}
\caption{Inference timing on the same hardware.}
\vspace{-3mm}
\label{tab:add_topotime}
\setlength{\tabcolsep}{3pt}
\resizebox{\linewidth}{!}{%
\begin{tabular}{lrr}
\toprule
Method & NFE & IT [s]\\
\midrule
Diffusion & 100 & 1.720\\
PG-Diffusion & 100 & 10.488\\
CoCoGen & 100 & 7.911\\
TopoDiff-G & 100 & 11.182\\
DOM & 100 & 3.314\\
PIDM & 100 & 1.701\\
PI-LSFM & 1 & 0.013\\
\bottomrule
\end{tabular}%
}
\endgroup
\vspace{-15mm}
\end{wraptable}

{\color{black}Topology optimization extends our evaluation from physics-constrained field generation to conditional inverse design. Following~\cite{bastek2024physics}, the model conditions on loading and boundary conditions, auxiliary physical fields, and a target volume constraint, and jointly generates the material density and displacement fields. Physical consistency is evaluated through the FEM equilibrium residual, while structural quality and constraint satisfaction are measured by compliance error and volume-fraction error. We evaluate both in-distribution and out-of-distribution boundary conditions.\par}
\par\WFclear
\endgroup

As shown in Table~\ref{tab:topopt_diffusion}, PI-LSFM achieves competitive inverse-design performance compared with multi-step diffusion-based baselines while requiring only a single function evaluation. Table~\ref{tab:add_topotime} further shows that PI-LSFM generates a design in 0.013 s, compared with 1.701--11.182 s for the listed 100-step baselines. Full problem formulation, metric definitions, training details, and qualitative results are provided in Appendix~\ref{appedix:topology}.

\begin{table}[H]
\vspace{-2mm}
\centering
\caption{Topology optimization results under both in-distribution and out-of-distribution boundary-condition settings. PI-LSFM attains competitive performance against 100-step diffusion-based baselines while using only a single function evaluation.}
\vspace{-2mm}
\label{tab:topopt_diffusion}
\begin{threeparttable}
\resizebox{\linewidth}{!}{%
\begin{tabular}{l c c c c c c c}
\toprule
\multirow{2}{*}{Model} & \multirow{2}{*}{Size}
& \multicolumn{3}{c}{\textit{in-distribution}} 
& \multicolumn{3}{c}{\textit{out-of-distribution}} \\
\cmidrule(lr){3-5} \cmidrule(lr){6-8}
& 
& $\mathcal{R}_{\mathrm{MAE}}\downarrow$ & MDN \% CE $\downarrow$ & \% VFE $\downarrow$
& $\mathcal{R}_{\mathrm{MAE}}\downarrow$ & MDN \% CE $\downarrow$ & \% VFE $\downarrow$ \\
\midrule
Diffusion & 136M & $1.86\mathrm{e}{-3}$ & -0.20 & 2.93 & $1.97\mathrm{e}{-3}$ & 0.30 & 2.80 \\
PG-Diffusion (\cite{shu2023physics}) & 136M & $1.82\mathrm{e}{-3}$ & 0.09 & 3.59 & $1.92\mathrm{e}{-3}$ & 0.81 & 3.23 \\
CoCoGen (\cite{jacobsen2025cocogen}) & 136M & $1.51\mathrm{e}{-3}$ & 0.14 & 4.00 & $1.56\mathrm{e}{-3}$ & 0.58 & 3.64 \\
TopoDiff-G (\cite{maze2023diffusion}) & 239M & -- & 0.83 & 1.49 & -- & 1.82 & 1.80 \\
DOM \cite{giannone2023aligning} & 121M & -- & 0.74 & 1.52 & -- & 3.47 & 1.59 \\
PIDM \cite{bastek2024physics} & 136M & $1.24\mathrm{e}{-3}$ & 0.06 & 2.25 & $1.29\mathrm{e}{-3}$ & 0.56 & 1.91 \\
PI-LSFM (ours) & 136M & $1.66\mathrm{e}{-3}$ & -0.00 & 1.53 & $1.95\mathrm{e}{-3}$ & 1.86 & 1.66 \\
\bottomrule
\end{tabular}%
}
\begin{tablenotes}
\scriptsize
\item[] ``--'' indicates that $\mathcal{R}_{\mathrm{MAE}}$ is not applicable because the method does not provide predicted solution fields.
\end{tablenotes}
\end{threeparttable}
\vspace{-7mm}
\end{table}

\vspace{-2.75mm}
\section{Conclusion}
\vspace{-1.75mm}
We introduced Long-Short Flow Maps with Direct Guidance (PI-LSFM), a one-step physics-informed generative framework that decomposes the full flow map into a long-range map and a near-terminal short map. This decomposition converts indirect flow-map and physics-informed supervision into a directly computable near-terminal guidance target, enabling joint data and physics guidance. We studied numerical instantiations of the short-flow dynamics and their effects on stability and generation quality. Experiments on unconditional generation, conditional generation, and physics-constrained inverse design show that PI-LSFM achieves a favorable trade-off among stability, distributional fidelity, physical consistency, and inference efficiency, matching strong multi-step baselines while preserving one-step efficiency without distillation. Our framework has several limitations. Similar to prior physics-informed generative methods~\cite{baldan2026physics,bastek2024physics}, PI-LSFM enforces physical constraints only during training, leaving inference-time constraint enforcement to future work. It would be interesting to combine PI-LSFM with inference-time constraint methods~\cite{cheng2024gradient,chang2026stabilizing}. We study only a limited set of numerical schemes for the data and physics terms; broader numerical families may further improve stability and performance. Scaling to larger models, datasets, and more complex physics tasks is another important direction.

\par\WFclear
\typeout{REFERENCE-WIDTH: linewidth=\the\linewidth; textwidth=\the\textwidth}
\subsection*{\textcolor{black}{AI use statement}}
{\color{black}We used generative AI tools primarily to collect, organize, and transcribe experimental information for the appendix, and to polish the writing of the main text. We also used these tools for literature retrieval and discovery to identify relevant related work. For mathematical work, AI assisted with developing the counterexample construction for Theorem~\ref{thm:potential_close_form} and writing the proof of Theorem~\ref{thm:potential_close_form}. The authors take responsibility for the final content of this paper, including the AI-assisted text, mathematical claims, and proofs.}

\bibliographystyle{preprint}
\bibliography{example_paper,3D_generation,physics_informed}


\clearpage
\appendix

\startcontents[appendix]
\section*{Appendix}
\addcontentsline{toc}{section}{Appendix}

\printcontents[appendix]{}{1}{\setcounter{tocdepth}{2}}

\section{Related Work}\label{sec:related_work}

\paragraph{Physics-informed generation.} Physics-informed generative methods can be broadly divided into training-time and inference-time approaches. Training-time methods incorporate physical constraints directly into the learning objective. For example, physics-informed diffusion models \cite{bastek2024physics,shu2023physics} add PDE residuals or boundary losses during training, while CoCoGen \cite{jacobsen2025cocogen} and DiffusionPDE \cite{huang2024diffusionpde} study physically consistent generation and PDE-conditioned modeling. FunDiff \cite{wang2025fundiff} instead formulates physics-informed generative modeling over function spaces.  Inference-time methods guide or correct samples during generation. D-Flow \cite{ben2024d} differentiates through the generative flow for controlled generation, while hard-constrained and gradient-free methods \cite{utkarsh2025physics,cheng2024gradient} enforce constraints during sampling without fully retraining the generator. {\color{black}Physics-informed distillation (PIDDM)~\cite{zhang2025physics} follows a teacher-based route, distilling a pretrained diffusion model into a one-step generator for PDE-constrained generation. PI-LSFM instead trains a one-step generator from scratch using joint data and physics guidance, without a pretrained generative teacher.}  Most closely related to our work, Physics vs Distributions \cite{baldan2026physics} highlights the trade-off between distribution matching and physical validity, and proposes a Pareto-optimal flow-matching framework with physics constraints. Physics-informed consistency models \cite{chang2026stabilizing}, building on continuous-time consistency models \cite{lu2024simplifying}, aim to stabilize one-step physics-informed generation through structure-preserving training. In contrast, our long-short flow-map direct guidance provides a direct one-step training objective that better balances distributional fidelity and physical validity, without requiring iterative inference-time correction.

\paragraph{One-step generation.}Early one-step generative models were mainly developed through distillation or adversarial training. For example, distribution matching distillation \cite{yin2024one} compresses multi-step diffusion sampling into a single-step generator, while GAN-based methods \cite{heusel2017gans} directly learn one-step mappings with adversarial objectives. More recently, increasing attention has been given to training one-step generators from scratch without relying on a pretrained teacher. A prominent direction formulates one-step generation as learning a flow map from noise to data \cite{boffi2025flow}. Representative methods include consistency models \cite{song2023consistency,lu2024simplifying,geng2024consistency}, shortcut models \cite{frans2025shortcut}, mean-flow methods \cite{geng2025mean}, and extensions such as functional mean flow \cite{li2025functional}. More recently, drifting models~\cite{deng2026generative} revisit one-step generation from a distribution-matching perspective, improving the stability and scalability of training-from-scratch one-step generators and concurrent work~\cite{li2026long} provides a long-short flow-map interpretation of drifting models. Our work instead studies physics-informed one-step generation, where the central challenge is to balance data-distribution fidelity and physical constraint satisfaction within a single flow map. We analyze this interaction through a potential-based formulation, which avoids the terminal-time singularity of velocity-based formulations at $t=1$, and propose a long-short flow-map framework with direct guidance for joint control of distributional and physical objectives.

{\color{black}PI-LSFM starts from a long-short flow-map decomposition of physics-informed one-step generation, providing a unified framework for analyzing and constructing direct guidance from both the data distribution and physical constraints. The near-terminal short flow is jointly driven by data and physics potentials, while different numerical discretizations induce explicit, implicit, and projection-based update rules within the same construction. This framework also clarifies its connection to drifting models~\cite{deng2026generative}: under a particular raw-space specialization, the data-side correction recovers a drifting-style update. PI-LSFM extends beyond this special case by jointly deriving physics guidance, systematically studying alternative data and physics discretizations, and supporting feature-space guidance for high-dimensional physical fields. The matched one-step comparisons and numerical studies evaluate these components and their interaction in physics-informed generation.}

\paragraph{Physics foundation models.}
Self-supervised learning has recently enabled foundation models for scientific systems. Inspired by vision methods such as masked autoencoders (MAE) \cite{he2022masked}, PDE foundation models aim to learn transferable spatiotemporal representations across equations, parameters, and conditions. Representative examples include Poseidon \cite{herde2024poseidon}, which uses multiscale operator learning for fluid-dynamics pretraining; DPOT \cite{hao2024dpot}, which learns autoregressive denoising operators with a transformer backbone; and MPP \cite{mccabe2024multiple}, which pretrains surrogate models across multiple physical systems. PROSE-FD \cite{liu2024prose} incorporates multimodal PDE descriptions and numerical fields, while MORPH \cite{rautela2025morph} and SPUS \cite{siddik2025spus} further address heterogeneous modalities, resolutions, and parameter efficiency. Unlike these works, we use physics foundation models only as feature encoders to stabilize our kernel-based training objective, while generation remains in the original physical field space.

\section{Missing Proofs}
\subsection{Proof of \autoref{thm:potential_close_form}}\label{proof:closed_form}
\paragraph{\autoref{thm:potential_close_form}}(Potential-Driven System for Flow Matching) {\color{black}Assume $p_0=\mathcal N(0,I)$, independent of the data endpoint $x_1$, and that $p_1$ is a smooth distribution with $p_1(x)>0$.}  The Flow Matching dynamics with velocity $u_t$, learned in \autoref{eq:flow_matching}, can be interpreted as a dynamical system that starts from an initial state $x_0 \sim p_0$ and evolves under a closed-form potential.
\begin{equation}
\frac{d}{dt}x_t = -\nabla {U_{\mathrm{data}}}(x,t),
\end{equation}
where ${U_{\mathrm{data}}}(x,t)$ is defined as
\begin{equation}\label{eq:closed_form_solution_U_proof}
    \begin{aligned}
        {U_{\mathrm{data}}}(x,t) =-\frac{1-t}{t} \log \mathbb{E}_{x_1\sim p_1}[k_t(x_1,x)]-\frac{1}{2t}\|x\|^2 + C_t,\qquad k_{t}(y,x) &= e^{-\frac{\|ty-x\|^2}{2(1-t)^2}}
    \end{aligned}
\end{equation}
where $C_t$ is a function that depends only on $t$.
\begin{proof}
Since Flow Matching uses the conditional path $X_t = tX_1 + (1-t)X_0$, Tweedie's formula gives $\mathbb{E}_{x_1\sim p_{1|t}}[x_1] = \mathbb{E}[x_1 | x_t = x] = {\color{black}\frac{1}{t}\left(x+(1-t)^2 \nabla \log(c p_t(x))\right)}$,where $c$ is an arbitrary constant. Moreover, by the property of the regression loss, the optimal velocity corresponding to \autoref{eq:flow_matching} can be written as $u_t^*(x) = \mathbb{E}_{p_{1|t}(x)}\![\frac{x_1-x}{1-t}] = \frac{\mathbb{E}_{p_{1|t}(x)}[x_1]-x}{1-t}$.  Therefore, we have
\begin{equation}
    \begin{aligned}
        u_t^*(x)
        &= \frac{\mathbb{E}_{p_{1|t}(x)}[x_1]-x}{1-t} \\
        &= \frac{{\color{black}\frac{1}{t}\left(x+(1-t)^2\nabla \log(c p_t(x))\right)}-x}{1-t} \\
        &= \frac{x}{t} + \frac{1-t}{t}\nabla \log c p_t(x) \\
        &= -\nabla (-\frac{1}{2t}\|x\|^2-\frac{1-t}{t}\log c p_t(x)).
    \end{aligned}
\end{equation}
Hence, the Flow Matching dynamics can be driven by the potential $U_{\mathrm{data}}(x,t) = -\frac{1}{2t}\|x\|^2-\frac{1-t}{t}\log cp_t(x)$.  Moreover, $p_t(x)$ admits a closed-form expression that can be computed directly from the data distribution:
\begin{equation}
    \begin{aligned}
    p_t(x) &= \int \int p_{t|0,1}(x| x_1,x_0)p_1(x_1)p_0(x_0)dx_0dx_1\\
    &= \int \int \delta(x_t-(tx_1 + (1-t)x_0)p_1(x_1)p_0(x_0)dx_0dx_1\\
    &= \int p_1(x_1)p_0(\frac{x_t-tx_1}{1-t})dx_1\\
    &= \int p_1(x_1)\frac{1}{(2\pi)^\frac{d}{2}(1-t)^d}e^{-\frac{\|x-tx_1\|^2}{2(1-t)^2}}dx_1\\
    &= \frac{1}{c}\mathbb{E}_{x_1\sim p_1}[k_t(x_1,x)]
    \end{aligned}
\end{equation}
Here, we choose $c$ as$(2\pi)^\frac{d}{2}(1-t)^d$.  Then $U_{data}$ have the formula:
\begin{equation}
    \begin{aligned}
        U_{\mathrm{data}}(x,t) &= -\frac{1}{2t}\|x\|^2-\frac{1-t}{t}\log cp_t(x)\\
        &=  -\frac{1}{2t}\|x\|^2-\frac{1-t}{t}\log \mathbb{E}_{x_1\sim p_1}[k_t(x_1,x)]
    \end{aligned}
\end{equation}
\end{proof}

\subsection{Derivation of \autoref{tab:data_potential_discretization}}\label{proof:data_tab}
\subsubsection{Forward Euler}
We first derive the formula for Forward Euler. Applying Forward Euler to numerically integrate \autoref{eq:closed_form_solution_U} over $[1-\Delta t, 1]$ gives:
\begin{equation}
    \begin{aligned}
         \mathcal{I}[\frac{d}{dt}\psi_{1-\Delta t\to t}(x);\,1-\Delta t, 1, \psi^\theta_{0\to 1-\Delta t}(x_0)] =  - \Delta t \nabla U_{data}(\psi^\theta_{0\to 1-\Delta t}(x_0),{\color{black}1-\Delta t})
    \end{aligned}
\end{equation}
Substituting the expression of $U_{data}$ from \autoref{thm:potential_close_form} yields:
\begin{equation}
    \begin{aligned}
         &\mathcal{I}[\frac{d}{dt}\psi_{1-\Delta t\to t}(x);\,1-\Delta t, 1, \psi^\theta_{0\to 1-\Delta t}(x_0)] \\
         &= - \Delta t \nabla [-\frac{1}{2(1-\Delta t)}\|x\|^2-\frac{\Delta t}{1-\Delta t}\log \mathbb{E}_{x_1\sim p_1}[k_{1-\Delta t}(x_1,x)]]|_{x = \psi^\theta_{0\to 1-\Delta t}(x_0)}\\
         &\overset{\textcircled{1}}{=}  \Delta t [\frac{x}{1-\Delta t}+\frac{\Delta t}{1-\Delta t}\frac{\mathbb{E}_{x_1\sim p_1}[\frac{(1-\Delta t)x_1-x}{\Delta t^2}k_{1-\Delta t}(x_1,x)]}{\mathbb{E}_{x_1\sim p_1}[k_{1-\Delta t}(x_1,x)]} ]|_{x = \psi^\theta_{0\to 1-\Delta t}(x_0)}\\
         &= \Delta t [\frac{\mathbb{E}_{x_1\sim p_1}[\frac{x_1-x}{\Delta t}k_{1-\Delta t}(x_1,x)]}{\mathbb{E}_{x_1\sim p_1}[k_{1-\Delta t}(x_1,x)]} ]|_{x = \psi^\theta_{0\to 1-\Delta t}(x_0)}\\
         &=  \frac{\mathbb{E}_{x_1\sim p_1}[(x_1-\psi^\theta_{0\to 1-\Delta t}(x_0))k_{1-\Delta t}(x_1,\psi^\theta_{0\to 1-\Delta t}(x_0))]}{\mathbb{E}_{x_1\sim p_1}[k_{1-\Delta t}(x_1,\psi^\theta_{0\to 1-\Delta t}(x_0))]} \\
    \end{aligned}
\end{equation}
where $\textcircled{1}$ holds because of $\nabla_xk_t(y,x) = \frac{ty-x}{(1-t)^2}k_t(y,x)$. After mollifying the above $k_t$ into $k_t^\epsilon$ and taking the limit, we obtain:
\begin{equation}
    \begin{aligned}
    \mathcal{T}_{data}(x_0) &=\lim_{\Delta t\to 0} \mathcal{I}[\frac{d}{dt}\psi_{1-\Delta t\to t}(x);\,1-\Delta t, 1, \psi^\theta_{0\to 1-\Delta t}(x_0)]\\
&=  \frac{\mathbb{E}_{x_1\sim p_1}[(x_1-\psi^\theta_{0\to 1}(x_0))k_1^\epsilon(x_1,\psi^\theta_{0\to 1}(x_0))]}{\mathbb{E}_{x_1\sim p_1}[k_{1}^\epsilon(x_1,\psi^\theta_{0\to 1}(x_0))]}
    \end{aligned}
\end{equation}

\subsubsection{Backward Euler}
We next consider the computation for Backward Euler. For Backward Euler, we need to use the potential at the endpoint $(t=1)$. By \autoref{thm:potential_close_form}, we have $U_{\mathrm{data}}(x,1)=-\frac12|x|^2$, which gives the direct endpoint velocity $u_1(x)=x$.  Although this velocity is directly available, we seek a form that is symmetric to the Forward Euler case and preserves the same kernel-based computation. This symmetry allows both the forward and backward endpoint velocities to be expressed using the same type of kernel-weighted estimator. It therefore avoids the inconsistency in the trapezoidal rule (\autoref{proof:trapezoidal_rule}) that would arise from using an analytic velocity at one endpoint and a kernel-estimated velocity at the other endpoint, while ensuring that numerical errors are reflected in the same manner at both endpoints. To this end, we use the following lemma.
\begin{lemma}
Let $\tilde p_t=(\psi_{0\to t}^{\theta})_{\#}p_0$. For $t\in(0,1)$, the following equation holds:
\begin{equation}
\frac{ \mathbb E_{\tilde x_t\sim\tilde p_t}[\frac{x-\tilde x_t}{1-t}k_t(x,\tilde x_t)]}{\mathbb E_{\tilde x_t\sim\tilde p_t}[k_t(x,\tilde x_t)]} +\frac{1-t}{t}\nabla_x
\log \mathbb E_{\tilde x_t\sim\tilde p_t} [k_t(x,\tilde x_t)]= x.
\end{equation}
Moreover, assume that $\psi_{0\to t}^{\theta}$ is a well-conditioned $C^1$-diffeomorphism. Then, on any compact region where $p_0((\psi_{0\to t}^{\theta})^{-1}(z))\ge m_0>0$, 
$\nabla_x\log\mathbb E_{\tilde x_t\sim\tilde p_t}[k_t(x,\tilde x_t)]$ is bounded and satisfies
$\|\nabla_x \log \mathbb E_{\tilde x_t\sim\tilde p_t}[ k_t(x,\tilde x_t)]\|\le\frac{e^{-1/2}LC+J}{(2\pi)^{d/2}m_0c}$, and consequently
\begin{equation}
    \begin{aligned}
        -\nabla U_{data}(x,1) = \frac{\mathbb E_{\tilde x_t\sim\tilde p_t}
[\frac{x-\tilde x_t}{1-t}k_t(x,\tilde x_t)]}{\mathbb E_{\tilde x_t\sim\tilde p_t}[k_t(x,\tilde x_t)]}+O(1-t)
    \end{aligned}
\end{equation}

Here, the constants $L,c,C,J>0$ are defined by the bounds
$\|\nabla_z(\psi_{0\to t}^{\theta})^{-1}(z)\| \le L$,
$0<c\le|\det\nabla_z(\psi_{0\to t}^{\theta})^{-1}(z)|\le C$, and $\|\nabla_z|\det\nabla_z(\psi_{0\to t}^{\theta})^{-1}(z)|\|\le J$.
\end{lemma}

\begin{proof}
Since $\nabla_x k_t(x,\tilde x_t)=-\frac{t(tx-\tilde x_t)}{(1-t)^2}k_t(x,\tilde x_t)$, we have
\begin{equation}
\nabla_x \log \mathbb E_{\tilde x_t\sim\tilde p_t}[k_t(x,\tilde x_t)]
=
-\frac{t}{(1-t)^2}
\frac{\mathbb E_{\tilde x_t\sim\tilde p_t}[(tx-\tilde x_t)k_t(x,\tilde x_t)]}{\mathbb E_{\tilde x_t\sim\tilde p_t}[k_t(x,\tilde x_t)]}.
\end{equation}
Therefore,
\begin{equation}
    \begin{aligned}
        &\frac{ \mathbb E_{\tilde x_t\sim\tilde p_t}[\frac{x-\tilde x_t}{1-t}k_t(x,\tilde x_t)]}{\mathbb E_{\tilde x_t\sim\tilde p_t}[k_t(x,\tilde x_t)]} +\frac{1-t}{t}\nabla_x
\log \mathbb E_{\tilde x_t\sim\tilde p_t} [k_t(x,\tilde x_t)] \\
&=\frac{ \mathbb E_{\tilde x_t\sim\tilde p_t}[\frac{x-\tilde x_t}{1-t}k_t(x,\tilde x_t)]}{\mathbb E_{\tilde x_t\sim\tilde p_t}[k_t(x,\tilde x_t)]} -\frac{1}{1-t}
\frac{\mathbb E_{\tilde x_t\sim\tilde p_t}[(tx-\tilde x_t)k_t(x,\tilde x_t)]}{\mathbb E_{\tilde x_t\sim\tilde p_t}[k_t(x,\tilde x_t)]}\\
&= x
\end{aligned}
\end{equation}
This proves the identity.

Next, since $\tilde p_t=(\psi_{0\to t}^{\theta})_{\#}p_0$, the change-of-variables formula gives $\tilde p_t(z)=p_0((\psi_{0\to t}^{\theta})^{-1}(z))|\det\nabla_z(\psi_{0\to t}^{\theta})^{-1}(z)|$.
Hence, 
\begin{equation}
    \begin{aligned}
        \tilde p_t(z)\ge m_0c
    \end{aligned}
\end{equation}
Moreover, since $\nabla_z\tilde p_t(z)
=
\nabla p_0((\psi_{0\to t}^{\theta})^{-1}(z))\nabla_z(\psi_{0\to t}^{\theta})^{-1}(z)
|\det\nabla_z(\psi_{0\to t}^{\theta})^{-1}(z)|
+
p_0((\psi_{0\to t}^{\theta})^{-1}(z))
\nabla_z|\det\nabla_z(\psi_{0\to t}^{\theta})^{-1}(z)|$, taking norms and using $\|\nabla p_0(z)\| = \|z\|\,p_0(z) = \frac{\|z\|}{(2\pi)^{d/2}} e^{-\frac{\|z\|^2}{2}} \le \frac{e^{-1/2}}{(2\pi)^{d/2}}$, $\|\nabla_z(\psi_{0\to t}^{\theta})^{-1}\|\le L$, $|\det\nabla_z(\psi_{0\to t}^{\theta})^{-1}|\le C$, $p_0(z) = \frac{1}{(2\pi)^{d/2}}e^{-\frac{\|z\|^2}{2}}\le \frac{1}{(2\pi)^{d/2}}$, and $\|\nabla_z|\det\nabla_z(\psi_{0\to t}^{\theta})^{-1}|\|\le J$, we obtain
\begin{equation}
\|\nabla_z \tilde p_t(z)\|\le \frac{e^{-1/2}LC+J}{(2\pi)^{d/2}}.
\end{equation}
Since $\mathbb E_{\tilde x_t\sim\tilde p_t}[k_t(x,\tilde x_t)]
=
\int \exp(-\frac{\|tx-z\|^2}{2(1-t)^2})\tilde p_t(z)\,dz$, this quantity is a Gaussian smoothing of $\tilde p_t$ evaluated at $tx$. Gaussian smoothing preserves the lower bound and the gradient bound. Therefore,
\begin{equation}
\|\nabla_x \log \mathbb E_{\tilde x_t\sim\tilde p_t}[ k_t(x,\tilde x_t)]\|\le\frac{e^{-1/2}LC+J}{(2\pi)^{d/2}m_0c}
\end{equation}
It follows that
\begin{equation}
\|-\frac{1-t}{t}\nabla_x \log \mathbb E_{\tilde x_t\sim\tilde p_t}[k_t(x,\tilde x_t)]\| \le \frac{1-t}{t}\frac{e^{-1/2}LC+J}{(2\pi)^{d/2}m_0c} = O(1-t).
\end{equation}
Therefore,
\begin{equation}
-\nabla U_{data}(x,1) = x =\frac{\mathbb E_{\tilde x_t\sim\tilde p_t}[\frac{x-\tilde x_t}{1-t}k_t(x,\tilde x_t)]}{\mathbb E_{\tilde x_t\sim\tilde p_t}[k_t(x,\tilde x_t)]}+O(1-t).
\end{equation}
This completes the proof.
\end{proof}
Using the above lemma, we have
\begin{equation}
    \begin{aligned}
         &\mathcal{I}[\frac{d}{dt}\psi_{1-\Delta t\to t}(x);\,1-\Delta t, 1, \psi^\theta_{0\to 1-\Delta t}(x_0)] \\
         &= \Delta t \frac{\mathbb E_{\tilde x_{1-\Delta t}\sim\tilde p_{1-\Delta t}}[\frac{x-\tilde x_{1-\Delta t}}{\Delta t}k_{1-\Delta t}(x,\tilde x_{1-\Delta t})]}{\mathbb E_{\tilde x_{1-\Delta t}\sim\tilde p_{1-\Delta t}}[k_{1-\Delta t}(x,\tilde x_{1-\Delta t})]}|_{\psi_{0\to 1}^\theta(x_0)} +O(\Delta t^2) \\
         &= \frac{\mathbb E_{\tilde x_{1-\Delta t}\sim\tilde p_{1-\Delta t}}[(\psi_{0\to 1}^\theta(x_0)-\tilde x_{1-\Delta t})k_{1-\Delta t}(\psi_{0\to 1}^\theta(x_0),\tilde x_{1-\Delta t})]}{\mathbb E_{\tilde x_{1-\Delta t}\sim\tilde p_{1-\Delta t}}[k_{1-\Delta t}(\psi_{0\to 1}^\theta(x_0),\tilde x_{1-\Delta t})]} +O(\Delta t^2) 
    \end{aligned}
\end{equation}
After mollifying the above $k_t$ into $k_t^\epsilon$ and taking the limit, we obtain:
\begin{equation}
    \begin{aligned}
    \mathcal{T}_{data}(x_0) &=\lim_{\Delta t\to 0} \mathcal{I}[\frac{d}{dt}\psi_{1-\Delta t\to t}(x);\,1-\Delta t, 1, \psi^\theta_{0\to 1-\Delta t}(x_0)]\\
&=  \frac{\mathbb E_{\tilde x_{1}\sim\tilde p_{1}}[(\psi_{0\to 1}^\theta(x_0)-\tilde x_{1})k^\epsilon_{1}(\psi_{0\to 1}^\theta(x_0),\tilde x_{1})]}{\mathbb E_{\tilde x_{1}\sim\tilde p_{1}}[k^\epsilon_{1}(\psi_{0\to 1}^\theta(x_0),\tilde x_{1})]}
    \end{aligned}
\end{equation}
\subsubsection{Trapezoidal
Rule}\label{proof:trapezoidal_rule}
Based on the calculation of the Forward Euler and Backward Euler cases, for the Trapezoidal Rule we have:
\begin{equation}
    \begin{aligned}
         &\mathcal{I}[\frac{d}{dt}\psi_{1-\Delta t\to t}(x);\,1-\Delta t, 1, \psi^\theta_{0\to 1-\Delta t}(x_0)]\\
         &=  - \Delta t \frac{1}{2}[\nabla U_{data}(\psi^\theta_{0\to 1-\Delta t}(x_0),1-\Delta t) + \nabla U_{data}(\psi^\theta_{0\to 1}(x_0),1)]\\
         &= \frac{1}{2}\frac{\mathbb{E}_{x_1\sim p_1}[(x_1-\psi^\theta_{0\to 1-\Delta t}(x_0))k_{1-\Delta t}(x_1,\psi^\theta_{0\to 1-\Delta t}(x_0))]}{\mathbb{E}_{x_1\sim p_1}[k_{1-\Delta t}(x_1,\psi^\theta_{0\to 1-\Delta t}(x_0))]} + \\
         &\frac{1}{2}\frac{\mathbb E_{\tilde x_{1-\Delta t}\sim\tilde p_{1-\Delta t}}[(\psi_{0\to 1}^\theta(x_0)-\tilde x_{1-\Delta t})k_{1-\Delta t}(\psi_{0\to 1}^\theta(x_0),\tilde x_{1-\Delta t})]}{\mathbb E_{\tilde x_{1-\Delta t}\sim\tilde p_{1-\Delta t}}[k_{1-\Delta t}(\psi_{0\to 1}^\theta(x_0),\tilde x_{1-\Delta t})]} +O(\Delta t^2) 
    \end{aligned}
\end{equation}
After mollifying the above $k_t$ into $k_t^\epsilon$ and taking the limit, we obtain:
\begin{equation}
    \begin{aligned}
    \mathcal{T}_{data}(x_0) &=\lim_{\Delta t\to 0} \mathcal{I}[\frac{d}{dt}\psi_{1-\Delta t\to t}(x);\,1-\Delta t, 1, \psi^\theta_{0\to 1-\Delta t}(x_0)]\\
&=  \frac{1}{2}\frac{\mathbb{E}_{x_1\sim p_1}[(x_1-\psi^\theta_{0\to 1}(x_0))k_1^\epsilon(x_1,\psi^\theta_{0\to 1}(x_0))]}{\mathbb{E}_{x_1\sim p_1}[k_{1}^\epsilon(x_1,\psi^\theta_{0\to 1}(x_0))]}+\frac{1}{2}\frac{\mathbb E_{\tilde x_{1}\sim\tilde p_{1}}[(\psi_{0\to 1}^\theta(x_0)-\tilde x_{1})k^\epsilon_{1}(\psi_{0\to 1}^\theta(x_0),\tilde x_{1})]}{\mathbb E_{\tilde x_{1}\sim\tilde p_{1}}[k^\epsilon_{1}(\psi_{0\to 1}^\theta(x_0),\tilde x_{1})]}
    \end{aligned}
\end{equation}

\subsection{Derivation of \autoref{tab:physics_potential_discretization}}\label{proof:physics_tab}
\subsubsection{Forward Euler Method}
Consider the physics-potential-driven dynamical system
$\frac{d x_\tau}{d\tau} = -\lambda \nabla U_{\mathrm{phys}}(x_\tau,\tau)$, where
$U_{\mathrm{phys}}(x,\tau)
= \frac{1}{1-\tau}\|R(x)\|^2$ and $\tau\in[t,1]$.
Applying Forward Euler to numerically integrate this system over $[1-\Delta t, 1]$ gives:
\begin{equation}
    \begin{aligned}
         \mathcal{I}[\frac{d}{dt}\psi_{1-\Delta t\to t}(x);\,1-\Delta t, 1, \psi^\theta_{0\to 1-\Delta t}(x_0)] &=  - \Delta t \lambda \nabla U_{phys}(\psi^\theta_{0\to 1-\Delta t}(x_0),1-\Delta t)\\
         &= - \Delta t \lambda \frac{1}{\Delta t} \nabla \|\mathcal{R}(x)\|^2|_{x = \psi^\theta_{0\to 1-\Delta t}(x_0)}\\
         &= -\lambda J_\mathcal{R}^\top(\psi^\theta_{0\to 1-\Delta t}(x_0)) \mathcal{R}(\psi^\theta_{0\to 1-\Delta t}(x_0))
    \end{aligned}
\end{equation}
Taking the limit $\Delta t\to 0$, we obtain:
\begin{equation}
    \begin{aligned}
    \mathcal{T}_{phys}(x_0) &=\lim_{\Delta t\to 0} \mathcal{I}[\frac{d}{dt}\psi_{1-\Delta t\to t}(x);\,1-\Delta t, 1, \psi^\theta_{0\to 1-\Delta t}(x_0)]\\
&=  -\lambda J_\mathcal{R}^\top(\psi^\theta_{0\to 1}(x_0)) \mathcal{R}(\psi^\theta_{0\to 1}(x_0))
    \end{aligned}
\end{equation}

\subsubsection{Backward Euler Method}
Applying Backward Euler to numerically integrate $\frac{d x_\tau}{d\tau} = -\lambda \nabla U_{\mathrm{phys}}(x_\tau,\tau)$ over $[1-\Delta t, 1]$ requires evaluating the gradient at the next state $x_1
= x_{1-\Delta t} - \Delta t \lambda \nabla U_{\mathrm{phys}}(x_1,1-\Delta t)$.  This is equivalent to the first-order optimality condition of the proximal problem
\begin{equation}\label{eq:optmial_problem_BE}
    \begin{aligned}
        x_1 &= \arg\min_x \frac{1}{2}\|x-x_{1-\Delta t}\|^2 + \Delta t\lambda U_{\mathrm{phys}}(x,1-\Delta t)\\
&=\arg\min_x
\frac{1}{2}\|x-x_{1-\Delta t}\|^2
+\lambda\|R(x)\|^2.
    \end{aligned}    
\end{equation}
Indeed, this follows by setting the derivative of the above objective to zero, which recovers
$x_1
= x_{1-\Delta t} - \Delta t \lambda \nabla U_{\mathrm{phys}}(x_1,1-\Delta t)$.  Letting $\Delta x = x-x_{1-\Delta t}$, and linearizing the residual around $x_t$ as $R(x_{1-\Delta t}+\Delta x)\approx R(x_{1-\Delta t})+J_R(x_{1-\Delta t})\Delta x$, the above problem becomes 
\begin{equation}
    \begin{aligned}
        \min_{\Delta x}
\frac{1}{2}\|\Delta x\|^2
+
\lambda \|R(x_{1-\Delta t})+J_R(x_{1-\Delta t})\Delta x\|^2
    \end{aligned}
\end{equation}
Define $f(\Delta x) = \frac{1}{2}\|\Delta x\|^2 + \lambda \|R(x_{1-\Delta t})+J_R(x_{1-\Delta t})\Delta x\|^2$. Its gradient is
$\nabla_{\Delta x} f(\Delta x)
=
\Delta x
+
2\lambda J_R^\top(x_{1-\Delta t})(R(x_{1-\Delta t})+J_R(x_{1-\Delta t})\Delta x)$.  To solve the optimization problem, setting $\nabla_{\Delta x} f(\Delta x)=0$ yields $\Delta x = - (J_R^\top(x_{1-\Delta t})J_R(x_{1-\Delta t})+ \tfrac{1}{2\lambda}E)^{-1} J_R^\top(x_{1-\Delta t})R(x_{1-\Delta t})$.
Therefore,
\begin{equation}
    \begin{aligned}
&\mathcal{I}[\frac{d}{dt}\psi_{1-\Delta t\to t}(x);\,1-\Delta t, 1, \psi^\theta_{0\to 1-\Delta t}(x_0)]\\
&=
-
(J_R^\top(x)J_R(x)+\frac{1}{2\lambda}E)^{-1}
J_R^\top(x)R(x)|_{x = \psi^\theta_{0\to 1-\Delta t}(x_0)}\\
&=-
(J_R^\top(\psi^\theta_{0\to 1-\Delta t}(x_0))J_R(\psi^\theta_{0\to 1-\Delta t}(x_0))+\frac{1}{2\lambda}E)^{-1}
J_R^\top(\psi^\theta_{0\to 1-\Delta t}(x_0))R(\psi^\theta_{0\to 1-\Delta t}(x_0))
    \end{aligned}
\end{equation}
Taking the limit $\Delta t\to 0$, we obtain:
\begin{equation}
    \begin{aligned}
    \mathcal{T}_{phys}(x_0) &=\lim_{\Delta t\to 0} \mathcal{I}[\frac{d}{dt}\psi_{1-\Delta t\to t}(x);\,1-\Delta t, 1, \psi^\theta_{0\to 1-\Delta t}(x_0)]\\
&=  -
(J_R^\top(\psi^\theta_{0\to1}(x_0))J_R(\psi^\theta_{0\to1}(x_0))+\frac{1}{2\lambda}E)^{-1}
J_R^\top(\psi^\theta_{0\to1}(x_0))R(\psi^\theta_{0\to1}(x_0))
    \end{aligned}
\end{equation}

\subsubsection{Backward Euler Method with $\lambda\to \infty$}\label{appendix:PBD}
In the limit $\lambda\to\infty$, the soft-penalty formulation \autoref{eq:optmial_problem_BE} degenerates into the hard-constraint projection
\begin{equation}
    \begin{aligned}
        x_1 = \arg\min_x \frac{1}{2}\|x-x_{1-\Delta t}\|^2 \quad \text{s.t.}\quad R(x)=0.
    \end{aligned}
\end{equation}
This is a classical projection problem encountered in Position-Based Dynamics (PBD) \cite{muller2007position}. Following a similar treatment, we linearize the constraint around $x_{1-\Delta t}$, yielding
\begin{equation}
    \begin{aligned}
\min_{\Delta x}
\frac{1}{2}\|\Delta x\|^2
\qquad
\text{s.t.}\qquad
R(x_{1-\Delta t})+J_R(x_{1-\Delta t})\Delta x=0.        
    \end{aligned}
\end{equation}
This is a constrained least-squares problem with a linear constraint. Its solution is given by
\begin{equation}
\Delta x
=
-
J_R^\top(x_{1-\Delta t})
\big(J_R(x_{1-\Delta t})J_R^\top(x_{1-\Delta t})\big)^{-1}
R(x_{1-\Delta t}).
\end{equation}
Therefore,
\begin{equation}
    \begin{aligned}
&\mathcal{I}[\frac{d}{dt}\psi_{1-\Delta t\to t}(x);\,1-\Delta t, 1, \psi^\theta_{0\to 1-\Delta t}(x_0)]\\
&=-J_R^\top(\psi^\theta_{0\to 1-\Delta t}(x_0))
\big(J_R(\psi^\theta_{0\to 1-\Delta t}(x_0))J_R^\top(\psi^\theta_{0\to 1-\Delta t}(x_0))\big)^{-1}
R(\psi^\theta_{0\to 1-\Delta t}(x_0))
    \end{aligned}
\end{equation}
Taking the limit $\Delta t\to 0$, we obtain:
\begin{equation}
    \begin{aligned}
    \mathcal{T}_{phys}(x_0) &=\lim_{\Delta t\to 0} \mathcal{I}[\frac{d}{dt}\psi_{1-\Delta t\to t}(x);\,1-\Delta t, 1, \psi^\theta_{0\to 1-\Delta t}(x_0)]\\
&=-J_R^\top(\psi^\theta_{0\to 1}(x_0))
\big(J_R(\psi^\theta_{0\to 1}(x_0))J_R^\top(\psi^\theta_{0\to 1}(x_0))\big)^{-1}
R(\psi^\theta_{0\to 1}(x_0))
    \end{aligned}
\end{equation}

\section{Additional Theoretical Results and Algorithms}

\subsection{Conditional Generation}\label{appedix:conditional_generation}
In the conditional generation setting, we train a conditional flow map $\psi_{t\to r}(\cdot| c)$, where $c$ denotes the condition. Therefore, under our long-short flow-map formulation, the loss in \autoref{eq:overview_loss} becomes
\begin{equation}\label{eq:conditional_overview_loss}
    \begin{aligned}
        \mathcal{L}(\theta)
        &\!=\!
        \mathbb{E}_{x_0\sim p_0}
        \|
        \psi_{0\to 1}^\theta(x_0|c)
        \!-\!
        sg\!(\psi^\theta_{0\to 1}(x_0|c)\!+\!\lim_{\Delta t\to 0}
        \mathcal{I}[\frac{d}{dt}\psi_{1\!-\!\Delta t\to t}(x|c);1\!-\!\Delta t, 1, \psi^\theta_{0\to 1\!-\!\Delta t}(x_0|c)]
        )
        \|^2
    \end{aligned}
\end{equation}
Here, the direct target becomes $\psi_{0\to 1}^\theta(x_0|c) + \mathcal{T}(x_0|c)$, where $\mathcal{T}(x_0|c)=\lim_{\Delta t\to 0} \mathcal{I}[\frac{d}{dt}\psi_{1-\Delta t\to t}(x|c);\,1-\Delta t, 1, \psi^\theta_{0\to 1-\Delta t}(x_0|c)]$. We can see that the formulas for conditional generation are obtained by replacing $\psi_{t\to r}$ with $\psi_{t\to r}(\cdot| c)$. Similarly, \autoref{tab:data_potential_discretization} and {\color{black}\autoref{tab:physics_potential_discretization}} can be directly modified as follows:
\begin{table}[H]
\vspace{-5mm}
\caption{Three standard discretizations of the data potential-driven dynamics for conditional generation.}
\label{tab:conditional_data_potential_discretization}
\centering
\small
\setlength{\tabcolsep}{8pt}
\renewcommand{\arraystretch}{1.4}
\resizebox{\linewidth}{!}{%
\begin{tabular}{p{0.08\linewidth} p{0.29\linewidth} p{0.53\linewidth}}
\toprule
\textbf{Scheme} & \textbf{Update formula} & \textbf{Target} \\

\begin{tabular}[t]{@{}l@{}}
FE
\end{tabular}
&
$
\begin{aligned}
&x_{n+1}
=
x_n
\\
&-\Delta t\, \nabla U_{\mathrm{data}}(x_n,t_n)
\end{aligned}
$
&
$
\begin{aligned}
\mathcal{T}_{data}(x_0)&=\frac{\mathbb{E}_{x_1\sim p_1(\cdot|c)}[(x_1-\psi^{\theta}_{0\to 1}(x_0|c))k^\epsilon_{1}(x_1,\psi^{\theta}_{0\to 1}(x_0|c))]}{\mathbb{E}_{x_1\sim p_1(\cdot|c)}[k^\epsilon_{1}(x_1,\psi^{\theta}_{0\to 1}(x_0|c))]}
\end{aligned}
$
\\

\begin{tabular}[t]{@{}l@{}}
BE
\end{tabular}
&
$
\begin{aligned}
&x_{n+1}
=
x_n\\
&-\Delta t \nabla U_{\mathrm{data}}(x_{n+1},t_{n+1})
\end{aligned}
$
&
$
\begin{aligned}
\mathcal{T}_{data}(x_0|c)&=\frac{\mathbb{E}_{\tilde x_{1}\sim [\psi_{0\to 1}^\theta(\cdot|c)]_\sharp p_0}[(\psi_{0 \to 1}^\theta(x_0|c)-\tilde x_{1})k^\epsilon_{1}(\psi_{0 \to 1}^\theta(x_0|c),\tilde x_{1})]}{\mathbb{E}_{\tilde x_{1}\sim [\psi_{0\to 1}^\theta(\cdot|c)]_\sharp p_0}[k^\epsilon_{1}(\psi_{0 \to 1}^\theta(x_0|c),\tilde x_{1})]}
\end{aligned}
$
\\

\begin{tabular}[t]{@{}l@{}}
TR
\end{tabular}
&
$
\begin{aligned}
&x_{n+1}
=
x_n
-\frac{\Delta t}{2}
(
\nabla U_{\mathrm{data}}(x_n,t_n)
\\&+
\nabla U_{\mathrm{data}}(x_{n+1},t_{n+1})
)
\end{aligned}
$
&
$
\begin{aligned}
\mathcal{T}_{data}(x_0|c)&= \frac{1}{2}\frac{\mathbb{E}_{x_1\sim p_1(\cdot|c)}[(x_1-\psi^{\theta}_{0\to 1}(x_0|c))k^\epsilon_{1}(x_1,\psi^{\theta}_{0\to 1}(x_0|c))]}{\mathbb{E}_{x_1\sim p_1(\cdot|c)}[k^\epsilon_{1}(x_1,\psi^{\theta}_{0\to 1}(x_0|c))]}\\
&+\frac{1}{2}\frac{\mathbb{E}_{\tilde x_{1}\sim [\psi_{0\to 1}^\theta(\cdot|c)]_\sharp p_0}[(\psi_{0 \to 1}^\theta(x_0|c)-\tilde x_{1})k^\epsilon_{1}(\psi_{0 \to 1}^\theta(x_0|c),\tilde x_{1})]}{\mathbb{E}_{\tilde x_{1}\sim [\psi_{0\to 1}^\theta(\cdot|c)]_\sharp p_0}[k^\epsilon_{1}(\psi^{\theta}_{0\to 1}(x_0|c),\tilde x_{1})]}
\end{aligned}
$
\\
\bottomrule
\end{tabular}%
}
\vspace{-5mm}
\end{table}

\begin{table}[H]
\vspace{-5mm}
\caption{Three standard discretizations of the Physics potential-driven dynamics for conditional generation.}
\label{tab:conditioanl_physics_potential_discretization}
\centering
\small
\setlength{\tabcolsep}{8pt}
\renewcommand{\arraystretch}{1.4}
\resizebox{\linewidth}{!}{%
\begin{tabular}{p{0.16\linewidth} p{0.29\linewidth} p{0.45\linewidth}}
\toprule
\textbf{Scheme} & \textbf{Update formula} & \textbf{Target} \\

\begin{tabular}[t]{@{}l@{}}
FE
\end{tabular}
&
$
\begin{aligned}
&x_{n+1}
=
x_n-\Delta t \lambda \nabla U_{\mathrm{phys}}(x_n,t_n)
\end{aligned}
$
&
$
\begin{aligned}
\mathcal{T}_{phys}(x_0|c)&= -\lambda J_{R}^\top(\psi_{0\to 1}^\theta(x_0|c))R(\psi_{0\to 1}^\theta(x_0|c))
\end{aligned}
$
\\

\begin{tabular}[t]{@{}l@{}}
BE
\end{tabular}
&
$
\begin{aligned}
&x_{n+1}
=
x_n\\
&-\Delta t \lambda  \nabla U_{\mathrm{phys}}(x_{n+1},t_{n+1})
\end{aligned}
$
&
$
\begin{aligned}
\mathcal{T}_{phys}(x_0|c)&=-(J_{R}^\top(\psi_{0\to 1}^\theta(x_0|c))J_{R}(\psi_{0\to 1}^\theta(x_0|c)) + \gamma E)^{-1}\\
&J_{R}^\top(\psi_{0\to 1}^\theta(x_0|c))R(\psi_{0\to 1}^\theta(x_0|c))
\end{aligned}
$
\\

\begin{tabular}[t]{@{}l@{}}
BE-Lim
\end{tabular}
&
$
\begin{aligned}
&x_{n+1}
=
x_n\\
&-\Delta t \lambda  \nabla U_{\mathrm{phys}}(x_{n+1},t_{n+1})\\
&\lambda\to \infty
\end{aligned}
$
&
$
\begin{aligned}
\mathcal{T}_{phys}(x_0|c)&= -J_{R}^\top(\psi_{0\to 1}^\theta(x_0|c))(J_{R}(\psi_{0\to 1}^\theta(x_0|c))\\
&J^\top_{R}(\psi_{0\to 1}^\theta(x_0|c)))^{-1}R(\psi_{0\to 1}^\theta(x_0|c))
\end{aligned}
$
\\
\bottomrule
\end{tabular}%
}
\vspace{-5mm}
\end{table}

Based on the above modification for conditional generation, \autoref{alg:training} can be adapted into a grouping-based implementation, as shown in \autoref{alg:conditional_training}. Specifically, we sample $B_c$ conditions, sample $B_s$ data samples for each condition, and sample $B=B_cB_s$ noise samples. Each step in \autoref{alg:conditional_training} is then computed within each condition group and averaged over all $B_c$ conditions.  After training, the learned flow map $\psi_{0\to 1}^\theta(\cdot|c)$ can be used to generate samples that satisfy both the physical constraints and the data distribution with condition $c$ input, namely,
\begin{equation}
    \begin{aligned}
        x^\theta_1 = \psi_{0\to 1}^\theta(x_0|c) \Rightarrow x^\theta_1\sim p_1(\cdot|c), \mathcal{R}(x_1^\theta) = 0
    \end{aligned}
\end{equation}

\begin{figure}[H]
\centering

\begin{minipage}{0.52\linewidth}
\hrule
\vspace{0.3em}
\refstepcounter{algorithm}
\textbf{Algorithm \thealgorithm} PI-LSFM Conditional Training
\label{alg:conditional_training}
\vspace{0.3em}
\hrule
\vspace{0.4em}
\begin{algorithmic}[1]
\Require dataset $\mathcal{D}$, condition batch size $B_c$, samples per condition $B_s$, initial model parameters $\theta$, learning rate $\eta$, Gaussian noise sampler ${\color{black}\mathcal{N}(0,I)}$, hyperparameter $\bar{\lambda}$
\Repeat
    \State Sample conditions $\{c_k\}_{k=1}^{B_c}$
    \For{$k=1,\ldots,B_c$}
        \State Sample $\mathcal{X}_1^{(k)}\!=\!\{x_{1}^{k,j}\}_{j=1}^{B_s}$ from $\mathcal{D}(\cdot| c_k)$
        \State Sample noise $\mathcal{X}_0^{(k)}=\{x_0^{k,j}\}_{j=1}^{B_s}$ from \\\qquad\qquad ${\color{black}\mathcal{N}(0,I)}$
        \State Compute $x_1^{k,j,\theta}\gets \psi_{0\to 1}^{\theta}(x_0^{k,j}| c_k)$ for \\\qquad\qquad$j=1,\ldots,B_s$
        \State Compute $\mathcal{T}_{\mathrm{data}}(x_0^{k,j}| c_k)$  using tab. \ref{tab:conditional_data_potential_discretization}
        \State Compute $\mathcal{T}_{\mathrm{phys}}(x_0^{k,j}| c_k)$  using tab. \ref{tab:conditioanl_physics_potential_discretization}
        \State $\mathcal{T}(x_0^{k,j}| c_k)\gets \mathcal{T}_{\mathrm{data}}(x_0^{k,j}| c_k)+$\\\qquad\qquad $\bar{\lambda}\mathcal{T}_{\mathrm{phys}}(x_0^{k,j}| c_k)$
    \EndFor
    \State $\mathcal{L}(\theta)\gets \frac{1}{B_cB_s}\sum_{k=1}^{B_c}\sum_{j=1}^{B_s}\|x_1^{k,j,\theta}-$\\\qquad\qquad $sg(x_1^{k,j,\theta}+\mathcal{T}(x_0^{k,j}| c_k))\|^2$
    \State $\theta\gets \theta-\eta\nabla_\theta\mathcal{L}(\theta)$
\Until{convergence}
\end{algorithmic}
\vspace{0.4em}
\hrule
\end{minipage}
\hfill
\begin{minipage}{0.47\linewidth}
\hrule
\vspace{0.3em}
\refstepcounter{algorithm}
\textbf{Algorithm \thealgorithm} PI-LSFM Feature-Space Training
\label{alg:feature_space_training}
\vspace{0.3em}
\hrule
\vspace{0.4em}
\begin{algorithmic}[1]
\Require dataset $\mathcal{D}$, batch size $B$, feature encoder $\phi$, initial model parameters $\theta$, learning rate $\eta$, Gaussian noise sampler ${\color{black}\mathcal{N}(0,I)}$, hyperparameter $\bar{\lambda}$
\Repeat
    \State Sample $\mathcal{X}_1=\{x_1^j\}_{j=1}^B$ from $\mathcal{D}$
    \State Sample noise $\mathcal{X}_0=\{x_0^i\}_{i=1}^B$ from ${\color{black}\mathcal{N}(0,I)}$
    \State $x_1^{i,\theta} \!\gets\! \psi_{0\to 1}^{\theta}(x_0^i)$ and $\mathcal{X}_1^\theta\!=\!\{x_1^{i,\theta}\}_{i=1}^B$
    \State Compute feature sets $\phi(\mathcal{X}_1)$ and $\phi(\mathcal{X}_1^\theta)$
    \State Compute $\{\mathcal{T}_{\mathrm{data}}^{\mathcal{Z}}(x_0^i)\}_{i=1}^B$ in feature \\
    \qquad space using tab. \ref{tab:feature_data_potential_discretization}
    \State Compute $\{\mathcal{T}_{\mathrm{phys}}(x_0^i)\}_{i=1}^B$ in the \\
    \qquad original space using tab. \ref{tab:physics_potential_discretization}
    \State $\mathcal{L}(\theta)\gets \frac{1}{B}\sum_{i=1}^B\|\phi(x_1^{i,\theta})-sg($\\
    \qquad$\phi(x_1^{i,\theta}+\mathcal{T}_{phys}(x_0^i))+\mathcal{T}^\mathcal{Z}_{data}(x_0^i))\|^2$
    \State $\theta\gets \theta-\eta\nabla_\theta\mathcal{L}(\theta)$
\Until{convergence}
\end{algorithmic}
\vspace{0.4em}
\hrule
\end{minipage}

\caption{Conditional and feature-space implementations of PI-LSFM training.}
\label{fig:conditional_feature_algorithms}
\end{figure}

\subsection{Stabilization via Feature Spaces}\label{appedix:feature_space}
As discussed above, directly computing the kernel $k_t^\epsilon(y,x)$ in the high-dimensional data space may suffer from the curse of dimensionality, making the kernel weights less informative and the estimated direction unstable. To alleviate this issue, we compute the kernel in a lower-dimensional feature space.

Let $\phi:\mathbb R^n\to\mathbb R^m$ be an encoder with $m<n$, and define the feature variable $\mathcal{Z}=\phi(X)$. Let $Z_1=\phi(X_1)$, where $X_1\sim p_1$, and denote the density of $Z_1$ by $p_1^{\mathcal \mathcal{Z}}$. Equivalently, for any test function $f$,
\begin{equation}
\mathbb E_{Z_1\sim p_1^{\mathcal \mathcal{Z}}}[f(Z_1)]
=
\mathbb E_{X_1\sim p_1}[f(\phi(X_1))].
\end{equation}

If Flow Matching is trained in the feature space $\mathcal \mathcal{Z}$, then, similarly to \autoref{thm:potential_close_form}, the corresponding dynamics can be interpreted as a potential-driven system $\frac{d}{dt}z_t=-\nabla_z U_{\mathrm{data}}^{\mathcal \mathcal{Z}}(z_t,t)$, with potential $U_{\mathrm{data}}^{\mathcal \mathcal{Z}}(z,t)$
\begin{equation}
U_{\mathrm{data}}^{\mathcal \mathcal{Z}}(z,t)
=
-\frac{1-t}{t}
\log
\mathbb E_{Z_1\sim p_1^{\mathcal \mathcal{Z}}}
[
k_t^{\mathcal \mathcal{Z}}(Z_1,z)
]
-\frac{1}{2t}\|z\|^2+C_t^{\mathcal \mathcal{Z}},\qquad k_t^{\mathcal \mathcal{Z}}(y,z)
=
e^{(
-\frac{\|ty-z\|^2}{2(1-t)^2}
)}.
\end{equation}
Equivalently,
\begin{equation}
U_{\mathrm{data}}^{\mathcal \mathcal{Z}}(z,t)
=
-\frac{1-t}{t}
\log
\mathbb E_{X_1\sim p_1}
[
k_t^{\mathcal \mathcal{Z}}(\phi(X_1),z)
]
-\frac{1}{2t}\|z\|^2+C_t^{\mathcal \mathcal{Z}}.
\end{equation}
The kernel in $U_{\mathrm{data}}^{\mathcal \mathcal{Z}}(z,t)$ is computed in the low-dimensional feature space, where the kernel weights are more stable. We next relate this feature-space potential to the original data-space potential, so that $U_{\mathrm{data}}^{\mathcal \mathcal{Z}}(z,t)$ can be used to characterize the dynamics on $\mathcal{X}$.

\begin{theorem}[Local-linear feature-data potential gradient relation]
Let $z=\phi(x)$, and let $p_{x|z}(x| z)$ denote the conditional density induced by $z=\phi(x)$. Assume that, near $t=1$, we approximately have $\phi(X_t)=Z_t$, where $X_t$ and $Z_t$ denote the conditional paths in the spaces $\mathcal{X}$ and $\mathcal{Z}$, respectively. Then, in the near-endpoint regime $t\to 1$, we have the following approximation:
\begin{equation}
\nabla_z U_{\mathrm{data}}^{\mathcal \mathcal{Z}}(z,t)
=
\mathbb E_{x\sim p_{x|z}}
[
J_{\phi}(x)\nabla_x U_{\mathrm{data}}(x,t)
].
\end{equation}
\end{theorem}
\begin{proof}
By the definition of the potential-driven dynamics, we have $u_t^{x}(x)=-\nabla_x U_{\mathrm{data}}(x,t)$ and $u_t^{z}(z)=-\nabla_z U_{\mathrm{data}}^{\mathcal \mathcal{Z}}(z,t)$, where $u_t^{x}(x)$ and $u_t^{z}(z)$ denote the Flow Matching velocities on $\mathcal{X}$ and $\mathcal{Z}$, respectively. We now relate $u_t^{x}(x)$ and $u_t^{z}(z)$. Near $t=1$, we approximately have $\phi(X_t)={\color{black}Z_t}$, and then we have
\begin{equation}
\begin{aligned}
u_t^{z,*}(z)
&=
\mathbb E[\frac{Z_1-z}{1-t}| Z_t=z] \\
&=
\mathbb E[
\mathbb E[
\frac{Z_1-z}{1-t}
| Z_t=z, X_t
]
| Z_t=z
] \\
&=
\mathbb E_{x\sim p_{x|z}}
[
\mathbb E[
\frac{Z_1-z}{1-t}
| Z_t=z, X_t=x
]
] \\
&\overset{Z_t={\color{black}\phi}(X_t)}{=}
\mathbb E_{x\sim p_{x|z}}
[
\mathbb E[
\frac{Z_1-{\color{black}\phi}(x)}{1-t}
| X_t=x
]
] \\
&=
\mathbb E_{x\sim p_{x|z}}
[
\frac{\mathbb E[Z_1| X_t=x]-{\color{black}\phi}(x)}{1-t}
] \\
&\overset{\textcircled{1}}{=}
\mathbb E_{x\sim p_{x|z}}
[
\frac{\mathbb E[{\color{black}\phi}(X_1)| X_t=x]-{\color{black}\phi}(x)}{1-t}
] \\
&=
\mathbb E_{x\sim p_{x|z}}
[
\mathbb E[
\frac{{\color{black}\phi}(X_1)-{\color{black}\phi}(x)}{1-t}
| X_t=x
]
] \\
&\overset{\textcircled{2}}{\approx}
\mathbb E_{x\sim p_{x|z}}
[
J_{{\color{black}\phi}}(x)
\mathbb E[
\frac{X_1-x}{1-t}
| X_t=x
]
] \\
&=
\mathbb E_{x\sim p_{x|z}}
[
J_{{\color{black}\phi}}(x)u_t^{x,*}(x)
].
\end{aligned}
\end{equation}
where $\textcircled{2}$ follows from a Taylor expansion and $\textcircled{1}$ holds because
\begin{equation}
\begin{aligned}
\mathbb E[Z_1| X_t=x]
&=
\mathbb E\big[\mathbb E[Z_1| X_t=x, X_1]| X_t=x\big] \\
&=
\mathbb E\big[{\color{black}\phi}(X_1)| X_t=x\big].
\end{aligned}
\end{equation}
Substituting $u_t^{x,*}(x)=-\nabla_x U_{\mathrm{data}}(x,t)$ and $u_t^{z,*}(z)=-\nabla_z U_{\mathrm{data}}^{\mathcal \mathcal{Z}}(z,t)$, we obtain
\begin{equation}
-\nabla_z U_{\mathrm{data}}^{\mathcal \mathcal{Z}}(z,t)
=
\mathbb E_{x\sim p_{x|z}}
[
J_{\phi}(x)(-\nabla_x U_{\mathrm{data}}(x,t))
].
\end{equation}
This completes the proof.
\end{proof}

We then consider learning trajectory consistency in the feature space using \autoref{eq:decomposition}, namely
$\|\phi(\psi_{0\to 1}(x_0))-\phi(\psi_{1-\Delta t\to 1}\circ \psi_{0\to 1-\Delta t}(x_0))\|^2$.
We assume that the encoder $\phi$ is well-behaved and approximately invertible. In this case, \autoref{eq:overview_loss} becomes
\begin{equation}\label{eq:featire_overview_loss}
    \begin{aligned}
        &\mathcal{L}(\theta)\\
        &\!=\!
        \mathbb{E}_{x_0\sim p_0}
        \|
        \phi(\psi_{0\to 1}^\theta(x_0))
        \!-\!
        sg\!(\phi(\psi^\theta_{0\to 1}(x_0)\!+\!\lim_{\Delta t\to 0}
        \mathcal{I}[\frac{d}{dt}\psi_{1\!-\!\Delta t\to t}(x);1\!-\!\Delta t, 1, \psi^\theta_{0\to 1\!-\!\Delta t}(x_0)]
        ))
        \|^2\\
&\!=\!
        \mathbb{E}_{x_0\sim p_0}
        \|
        \phi(\psi_{0\to 1}^\theta(x_0))
        \!-\!
        sg\!(\phi(\psi^\theta_{0\to 1}(x_0)\!+\mathcal{T}_{phys}(x_0)-\!\lim_{\Delta t\to 0}
        \mathcal{I}[\nabla U_{data}(x);1\!-\!\Delta t, 1, \psi^\theta_{0\to 1\!-\!\Delta t}(x_0)]
        ))
        \|^2\\
&\!\overset{\textcircled{1}}{\approx}\!
        \mathbb{E}_{x_0\sim p_0}
        \|
        \phi(\psi_{0\to 1}^\theta(x_0))
        \!-\!
        sg\!(\phi(\psi^\theta_{0\to 1}(x_0)\!+\mathcal{T}_{phys}(x_0))-\!\lim_{\Delta t\to 0}
        \mathcal{I}[J_{\phi}(x)\nabla U_{data}(x);1\!-\!\Delta t, 1, \psi^\theta_{0\to 1\!-\!\Delta t}(x_0)]
        )
        \|^2\\
        &=\!
        \mathbb{E}_{x_0\sim p_0}
        \|
        \phi(\psi_{0\to 1}^\theta(x_0))
        \!-\!
        sg\!(\phi(\psi^\theta_{0\to 1}(x_0)\!+\mathcal{T}_{phys}(x_0))-\!\lim_{\Delta t\to 0}
        \mathcal{I}[\nabla_z U_{data}^{\mathcal{Z}};1\!-\!\Delta t, 1, \phi(\psi^\theta_{0\to 1\!-\!\Delta t}(x_0))]
        )
        \|^2\\
        &=\!
        \mathbb{E}_{x_0\sim p_0}
        \|
        \phi(\psi_{0\to 1}^\theta(x_0))
        \!-\!
        sg\!(\phi(\psi^\theta_{0\to 1}(x_0)\!+\mathcal{T}_{phys}(x_0))+\mathcal{T}^{\mathcal{Z}}_{data}(x_0)
        )
        \|^2\\
    \end{aligned}
\end{equation}
where $\textcircled{1}$ follows from a Taylor expansion. This is exactly the formula stated in the main text as \autoref{eq:feature_loss}, where
$\mathcal{T}^{\mathcal{Z}}_{data}(x_0) = \lim_{\Delta t\to 0}
        \mathcal{I}[\nabla_z U_{data}^{\mathcal{Z}};1\!-\!\Delta t, 1, \phi(\psi^\theta_{0\to 1\!-\!\Delta t}(x_0))]$
is numerically computed in the feature space. Note that the physics target $\mathcal{T}_{phys}(x_0)$ does not require any modification; only the data target $\mathcal{T}^{\mathcal{Z}}_{data}(x_0)$ needs to be computed in the feature space. The corresponding targets obtained by different numerical schemes in the feature space are given in \autoref{tab:feature_data_potential_discretization}. Our derivation relies on the assumption that the encoder $\phi$ is well defined and well behaved. Empirically, we find that the encoders used in our experiments work effectively under this formulation.  

It is worth noting that the use of the encoder and feature space only affects the training process. During inference, generation is still performed in the original space $\mathcal{X}$. Therefore, after learning the flow map $\psi_{0\to 1}^\theta$,  the inference procedure is the same as in \autoref{sec:algorithm}.
\begin{equation}
    \begin{aligned}
        x^\theta_1 = \psi_{0\to 1}^\theta(x_0) \Rightarrow x^\theta_1\sim p_1, \mathcal{R}(x_1^\theta) = 0
    \end{aligned}
\end{equation}

\begin{table}[H]
\vspace{-5mm}
\caption{Three standard discretizations of the data potential-driven dynamics in feature space.}
\label{tab:feature_data_potential_discretization}
\centering
\small
\setlength{\tabcolsep}{8pt}
\renewcommand{\arraystretch}{1.4}
\begin{tabular}{p{0.08\linewidth} p{0.29\linewidth} p{0.53\linewidth}}
\toprule
\textbf{Scheme} & \textbf{Update formula} & \textbf{Target} \\
\begin{tabular}[t]{@{}l@{}}
FE
\end{tabular}
&
$
\begin{aligned}
&z_{n+1}
=
z_n
\\
&-\Delta t\, \nabla U^\mathcal{Z}_{\mathrm{data}}(z_n,t_n)
\end{aligned}
$
&
$
\begin{aligned}
\mathcal{T}_{data}^{\mathcal{Z}}(x_0)&=\frac{\mathbb{E}_{z_1\sim p^\mathcal{Z}_1}[(z_1-\phi(\psi^{\theta}_{0\to 1}(x_0)))k^\epsilon_{1}(z_1,\phi(\psi^{\theta}_{0\to 1}(x_0)))]}{\mathbb{E}_{z_1\sim p^\mathcal{Z}_1}[k^\epsilon_{1}(z_1,\phi(\psi^{\theta}_{0\to 1}(x_0)))]}
\end{aligned}
$
\\

\begin{tabular}[t]{@{}l@{}}
BE
\end{tabular}
&
$
\begin{aligned}
&z_{n+1}
=
z_n\\
&-\Delta t \nabla U^\mathcal{Z}_{\mathrm{data}}(z_{n+1},t_{n+1})
\end{aligned}
$
&
$
\begin{aligned}
\mathcal{T}_{data}^{\mathcal{Z}}(x_0)&=\frac{\mathbb{E}_{\tilde z_{1}\sim [\phi\circ\psi_{0\to 1}^\theta]_\sharp p_0}[(\phi(\psi_{0 \to 1}^\theta(x_0))-\tilde z_{1})k^\epsilon_{1}(\phi(\psi_{0 \to 1}^\theta(x_0)),\tilde z_{1})]}{\mathbb{E}_{\tilde z_{1}\sim [\phi\circ\psi_{0\to 1}^\theta]_\sharp p_0}[k^\epsilon_{1}(\phi(\psi_{0 \to 1}^\theta(x_0)),\tilde z_{1})]}
\end{aligned}
$
\\

\begin{tabular}[t]{@{}l@{}}
TR
\end{tabular}
&
$
\begin{aligned}
&z_{n+1}
=
z_n
-\frac{\Delta t}{2}
(
\nabla U^\mathcal{Z}_{\mathrm{data}}(z_n,t_n)
\\&+
\nabla U^\mathcal{Z}_{\mathrm{data}}(z_{n+1},t_{n+1})
)
\end{aligned}
$
&
$
\begin{aligned}
\mathcal{T}_{data}^{\mathcal{Z}}(x_0)&= \frac{1}{2}\frac{\mathbb{E}_{z_1\sim p^\mathcal{Z}_1}[(z_1-\phi(\psi^{\theta}_{0\to 1}(x_0)))k^\epsilon_{1}(z_1,\phi(\psi^{\theta}_{0\to 1}(x_0)))]}{\mathbb{E}_{z_1\sim p^\mathcal{Z}_1}[k^\epsilon_{1}(z_1,\phi(\psi^{\theta}_{0\to 1}(x_0)))]}\\
&+\frac{1}{2}\frac{\mathbb{E}_{\tilde z_{1}\sim [\phi\circ\psi_{0\to 1}^\theta]_\sharp p_0}[(\phi(\psi_{0 \to 1}^\theta(x_0))-\tilde z_{1})k^\epsilon_{1}(\phi(\psi_{0 \to 1}^\theta(x_0)),\tilde z_{1})]}{\mathbb{E}_{\tilde z_{1}\sim [\phi\circ\psi_{0\to 1}^\theta]_\sharp p_0}[k^\epsilon_{1}(\phi(\psi_{0 \to 1}^\theta(x_0)),\tilde z_{1})]}
\end{aligned}
$
\\
\bottomrule
\end{tabular}%
\vspace{-5mm}
\end{table}
\subsection{Pytorch Code}\label{appedix:code}
Here, we take the trapezoidal rule of \autoref{tab:data_potential_discretization} as an example and provide the parallel PyTorch implementation of the expectation computation in \autoref{eq:expectation_calculation}. In each update, we sample a data batch $\mathcal{X}_1$ and compute a generated batch $\mathcal{X}_1^\theta$ by mapping noise samples through the learned flow map. The generated batch $\mathcal{X}_1^\theta$ is also used as samples from the distribution $[\psi_{0\to 1}^\theta]_\sharp p_0$, denoted by $\tilde{\mathcal{X}}_1=\mathcal{X}_1^\theta$.

For each generated sample $x_1^{\theta,i}\in\mathcal{X}_1^\theta$, we compute kernel-weighted averages over all samples in the data batch $\mathcal{X}_1$ and the generated batch $\tilde{\mathcal{X}}_1$. These kernel-weighted averages provide empirical estimates of the expectations used to construct the target $\mathcal{T}_{data}(x_0^i)$ for $x_1^{\theta,i}$. The expectation computation for all samples in $\mathcal{X}_1^\theta$ is fully parallelized, and the corresponding PyTorch implementation is shown below.
\begin{lstlisting}
import torch

def compute_kernel_weighted_targets(
    x1_theta: torch.Tensor, #[B...]
    x1: torch.Tensor, #[B...]
    tilde_x1: torch.Tensor, #[B...]
    epsilon: float,
    eps: float = 1e-12,
):
    dist2_data = torch.cdist(x1_theta, x1, p=2) ** 2
    weights_data = torch.exp(-dist2_data / epsilon).clamp_min(eps)
    V_forward = (weights_data @ x1) / weights_data.sum(dim=1, keepdim=True)

    dist2_tilde = torch.cdist(x1_theta, tilde_x1, p=2) ** 2
    weights_tilde = torch.exp(-dist2_tilde / epsilon).clamp_min(eps)
    V_backward = -(weights_tilde @ tilde_x1) / weights_tilde.sum(dim=1, keepdim=True)

    return 0.5*(V_forward+V_backward)
\end{lstlisting}

\section{Design Study and Additional Results}\label{appendix:design_study}

\subsection{Physics Term Numerical Integration Study}\label{appendix:physics_design_study}

We next study the impact of different numerical discretizations in \autoref{tab:physics_potential_discretization} for the physics term. Specifically, we compare Forward Euler, Backward Euler, and Backward Euler(limit), while fixing the data term to its best-performing configuration, namely the trapezoidal rule.

\begin{figure}[H]
\centering

\hspace{-5mm}
\begin{subfigure}[t]{0.32\linewidth}
    \centering
    \includegraphics[width=1.2\linewidth]{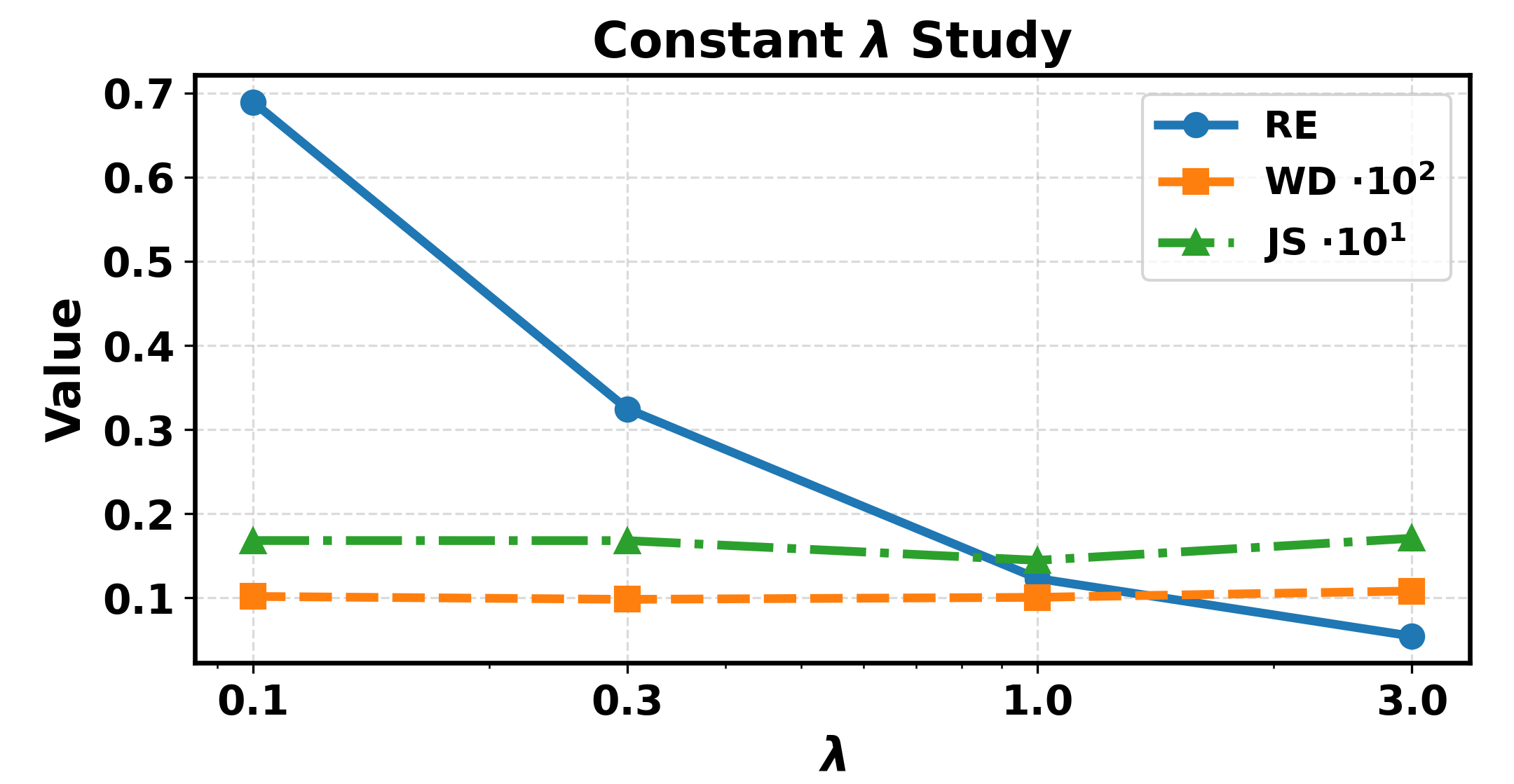}
    \caption{Forward: trend.}
    \label{fig:physics_forward_trend}
\end{subfigure}
\hfill
\begin{subfigure}[t]{0.32\linewidth}
    \centering
    \includegraphics[width=1.2\linewidth]{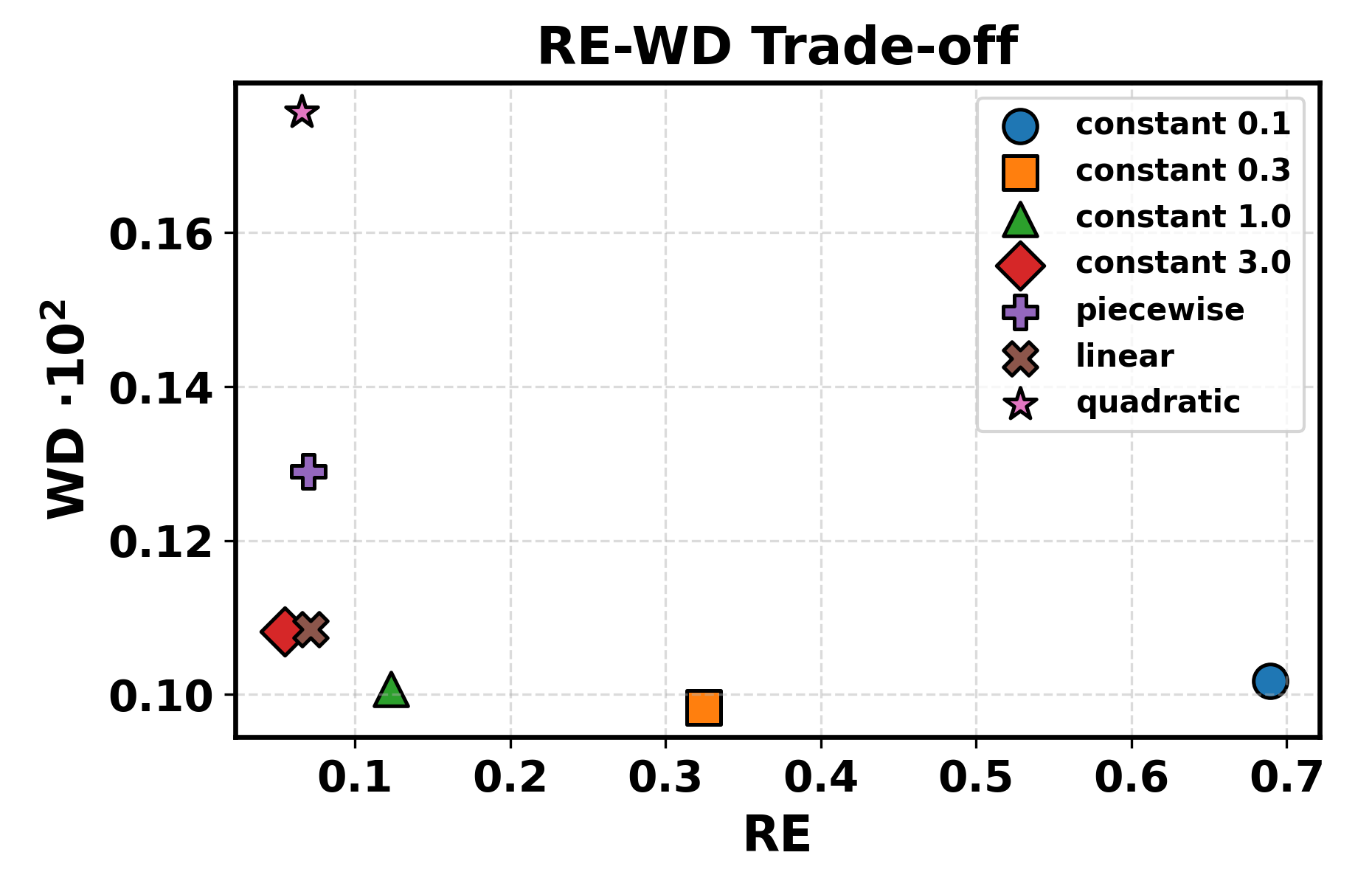}
    \caption{Forward: RE vs. WD.}
    \label{fig:physics_forward_re_wd}
\end{subfigure}
\hfill
\begin{subfigure}[t]{0.32\linewidth}
    \centering
    \includegraphics[width=1.2\linewidth]{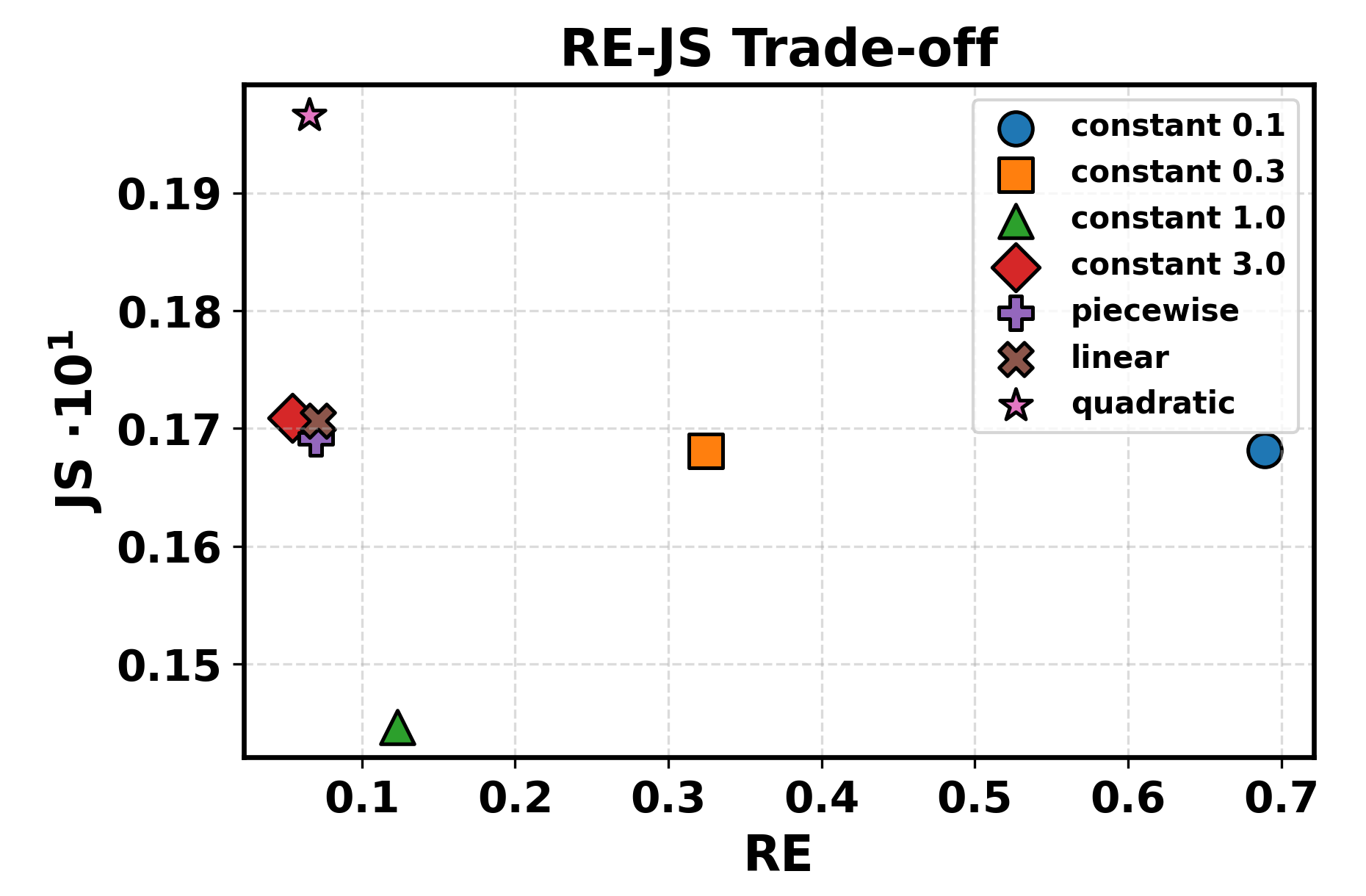}
    \caption{Forward: RE vs. JS.}
    \label{fig:physics_forward_re_js}
\end{subfigure}

\vspace{2mm}

\hspace{-5mm}
\begin{subfigure}[t]{0.32\linewidth}
    \centering
    \includegraphics[width=1.2\linewidth]{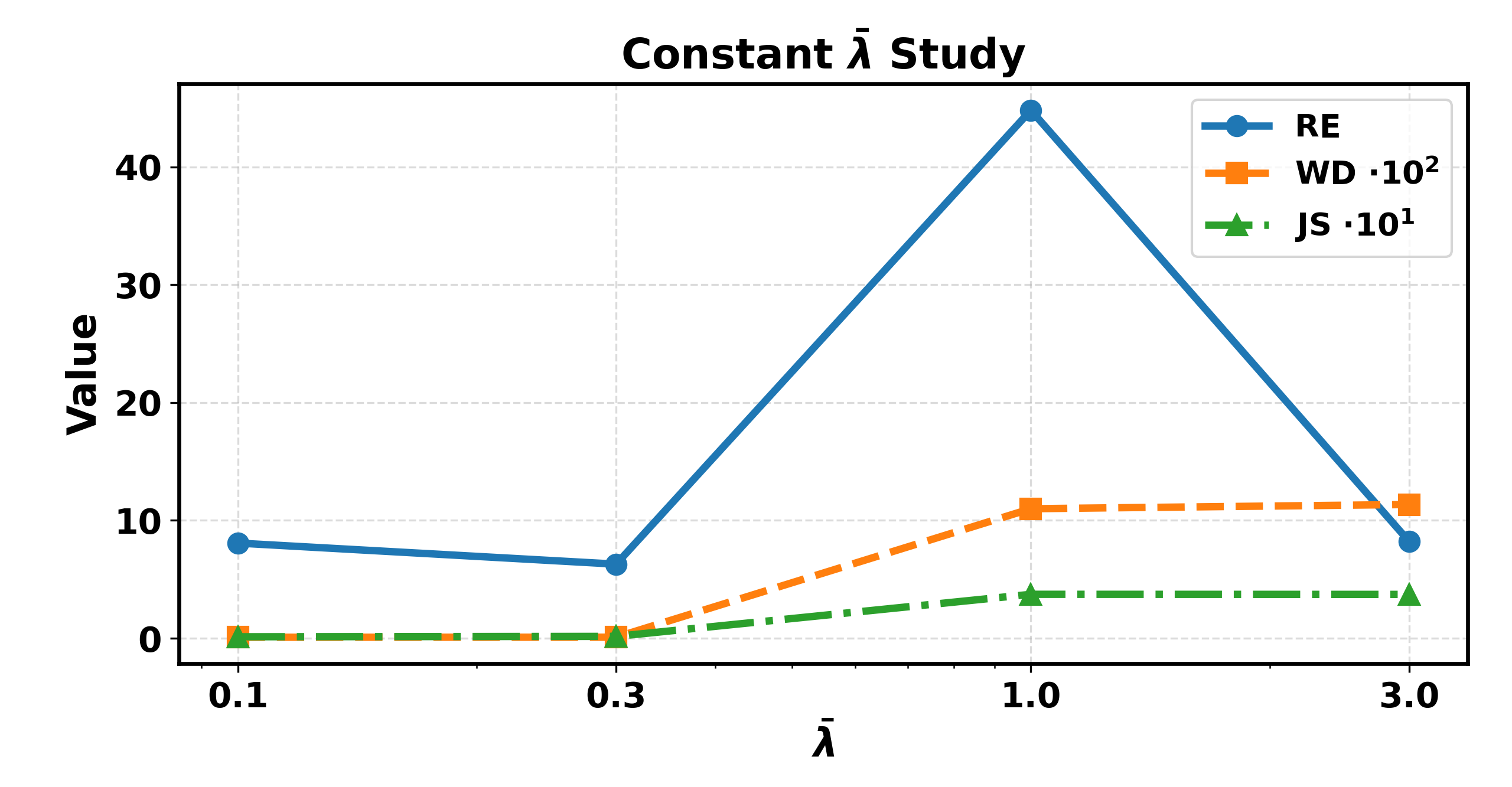}
    \caption{Backward: trend.}
    \label{fig:physics_backward_trend}
\end{subfigure}
\hfill
\begin{subfigure}[t]{0.32\linewidth}
    \centering
    \includegraphics[width=1.2\linewidth]{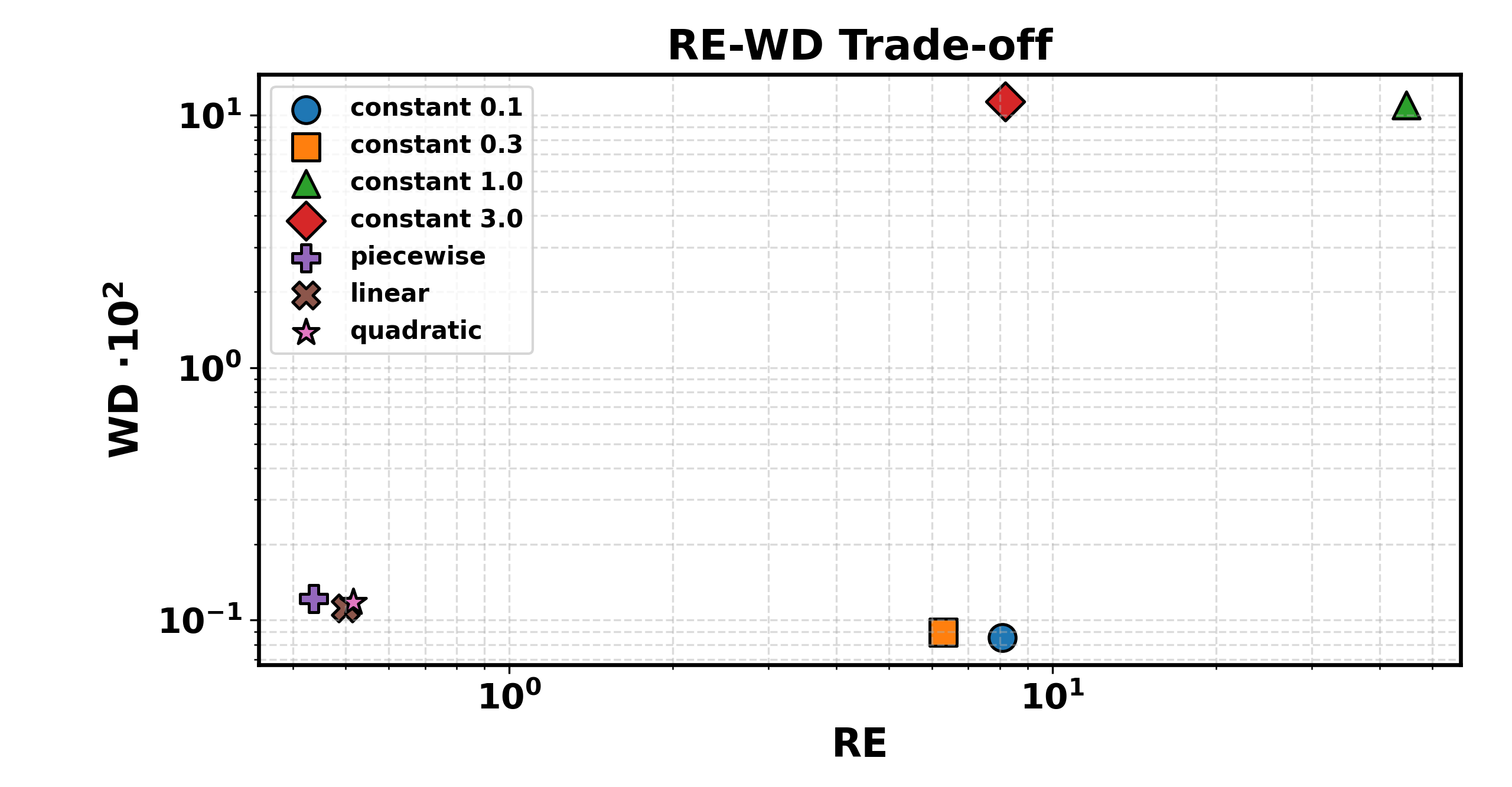}
    \caption{Backward: RE vs. WD.}
    \label{fig:physics_backward_re_wd}
\end{subfigure}
\hfill
\begin{subfigure}[t]{0.32\linewidth}
    \centering
    \includegraphics[width=1.2\linewidth]{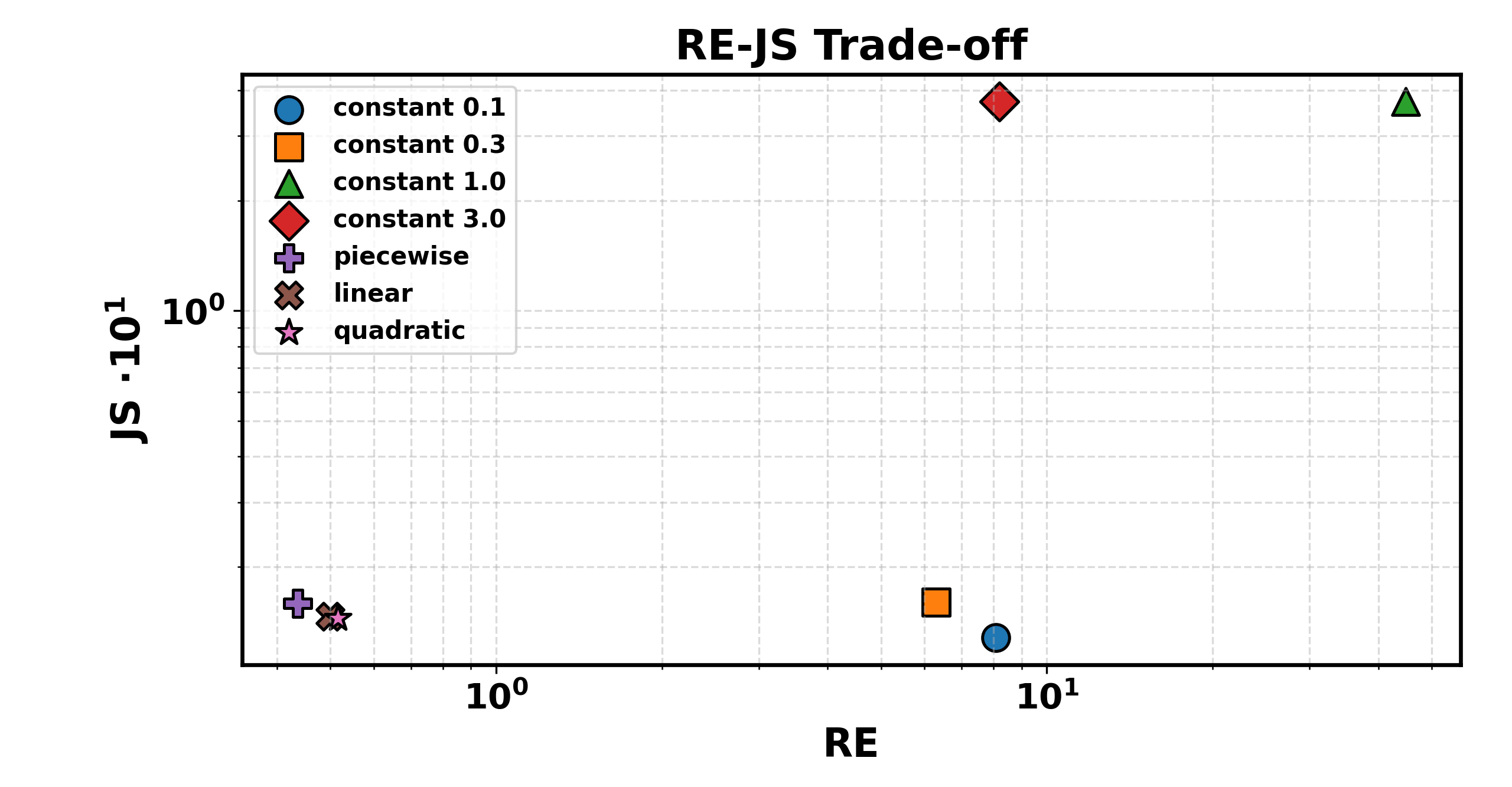}
    \caption{Backward: RE vs. JS.}
    \label{fig:physics_backward_re_js}
\end{subfigure}

\vspace{2mm}

\hspace{-5mm}
\begin{subfigure}[t]{0.32\linewidth}
    \centering
    \includegraphics[width=1.2\linewidth]{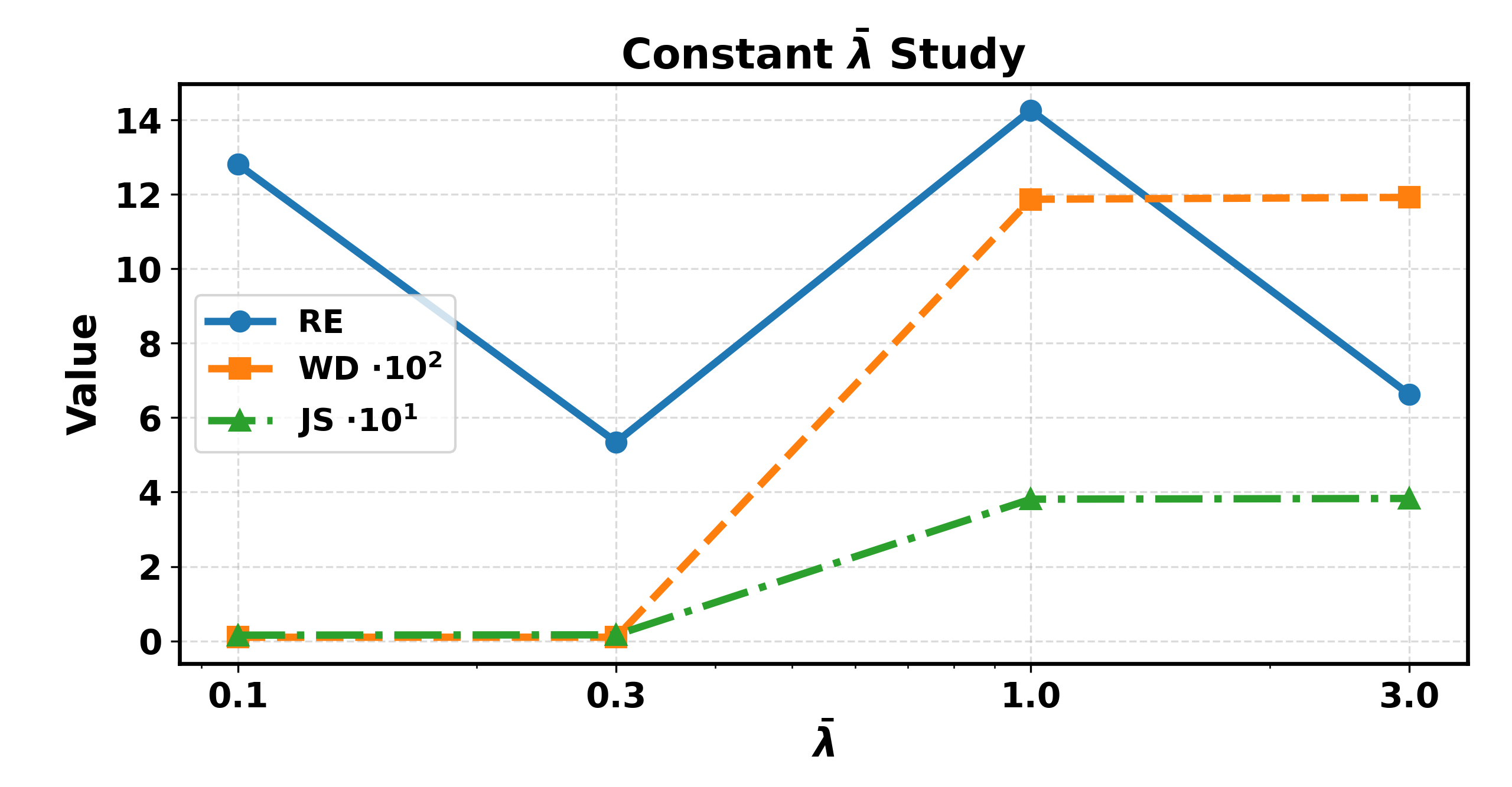}
    \caption{Backward (Limit): trend.}
    \label{fig:physics_backward_inf_trend}
\end{subfigure}
\hfill
\begin{subfigure}[t]{0.32\linewidth}
    \centering
    \includegraphics[width=1.2\linewidth]{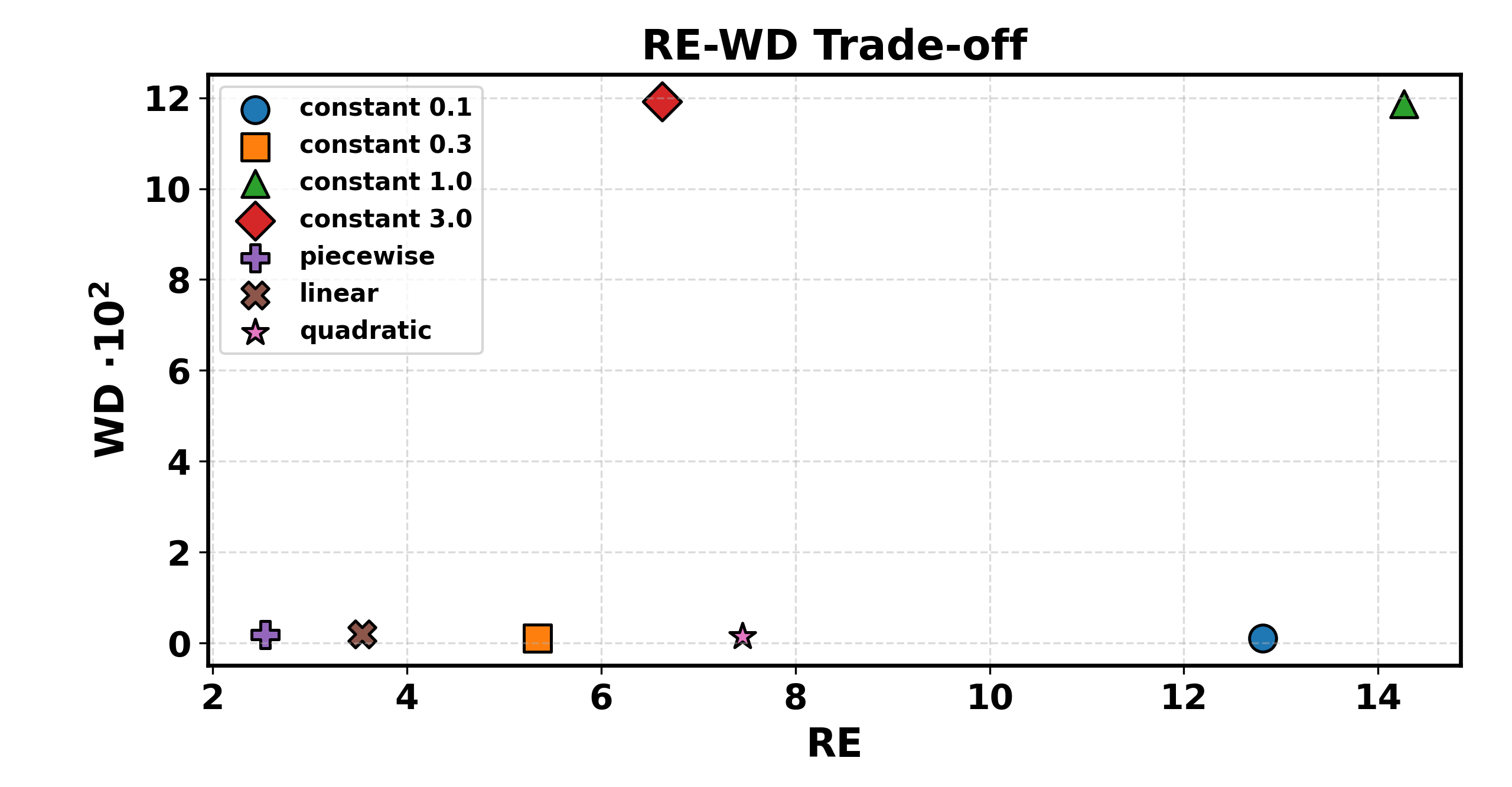}
    \caption{Backward (Limit): RE vs. WD.}
    \label{fig:physics_backward_inf_re_wd}
\end{subfigure}
\hfill
\begin{subfigure}[t]{0.32\linewidth}
    \centering
    \includegraphics[width=1.2\linewidth]{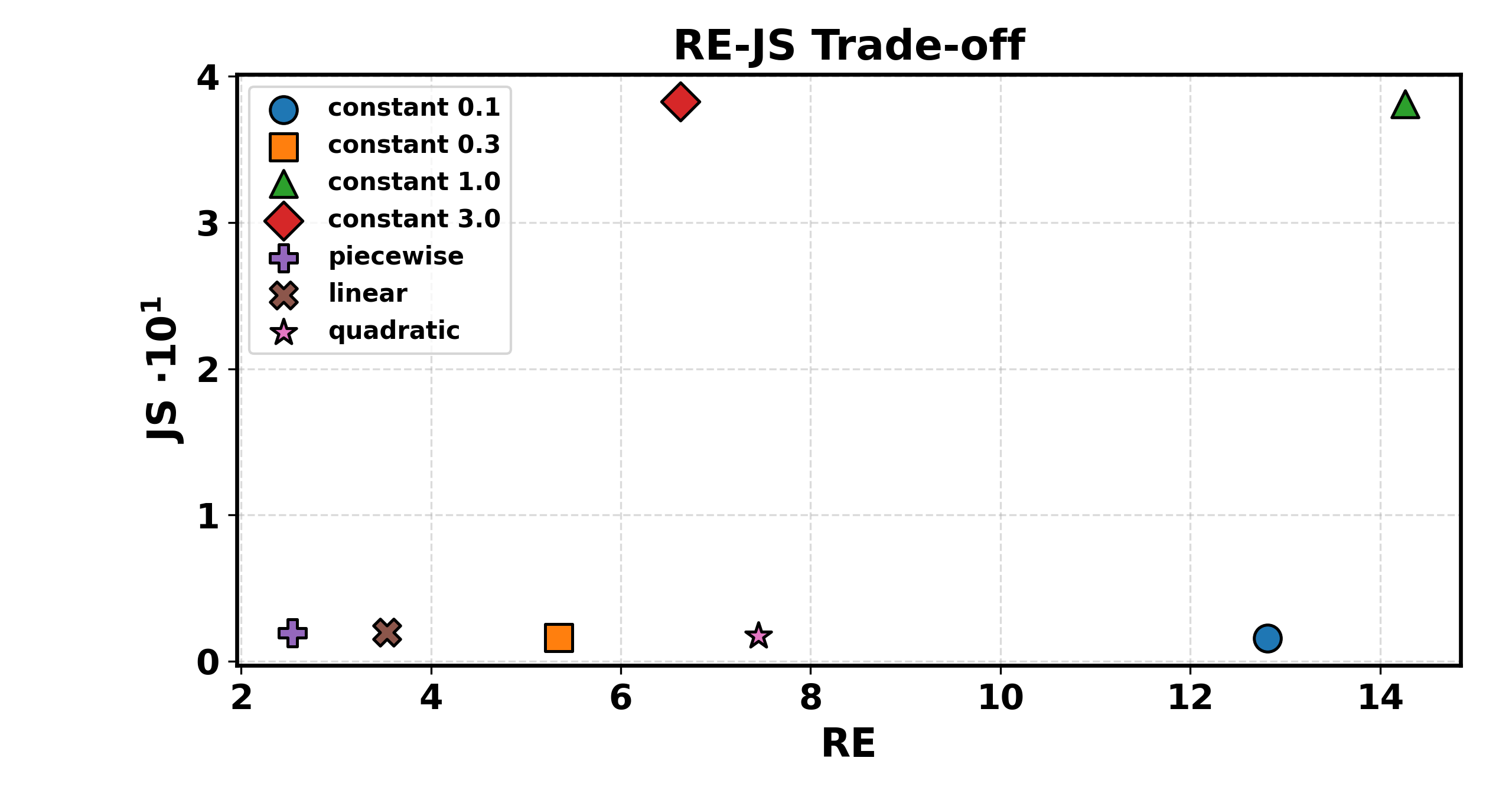}
    \caption{Backward (Limit): RE vs. JS.}
    \label{fig:physics_backward_inf_re_js}
\end{subfigure}

\caption{Comparison of constant step sizes and scheduling strategies for different physics-term discretizations. Each row corresponds to one numerical scheme, while columns show metric trends, RE--WD trade-offs, and RE--JS trade-offs. Due to the large dynamic range of the metrics for the Backward Euler methods, logarithmic scales are used in the scatter plots for better visualization.}
\label{fig:physics_step_size_comparison}
\end{figure}

In the experiments, we evaluate four constant values of the coefficient $\bar{\lambda}$ from $0.1$ to $3.0$, which controls the relative strength between the data term and the physics term. For each numerical scheme, we plot the trends of RE, WD, and JS as $\bar{\lambda}$ varies, as shown in \autoref{fig:physics_forward_trend},\autoref{fig:physics_backward_trend} and \autoref{fig:physics_backward_inf_trend}. Unlike Forward Euler, Backward Euler does not exhibit a simple smooth trend. In particular, as shown in \autoref{fig:physics_backward_trend}, WD and JS increase sharply when $\bar{\lambda}$ changes from $0.3$ to $0.1$. To address this sensitivity, we additionally consider three scheduling strategies for $\bar{\lambda}$: piecewise, linear, and quadratic schedules. Let $s$ denote the training step, and let $s_{\mathrm{warm}}$ and $s_{\mathrm{ramp}}$ denote the end of the warm-up stage and ramping stage, respectively. The piecewise schedule is defined as $\bar{\lambda}(s)
=
\begin{cases}
0.1, & s < s_{\mathrm{warm}},\\
1.0, & s_{\mathrm{warm}}\le s < s_{\mathrm{ramp}},\\
3.0, & s\ge s_{\mathrm{ramp}}.
\end{cases}$ with $s_{\mathrm{ramp}} = 200k$ and ${\color{black}s_{\mathrm{warm}}} = 50k$.  For the linear and quadratic schedules, we set $\bar{\lambda}_{\mathrm{start}}=0.1$ and $\bar{\lambda}_{\mathrm{end}}=3.0$, define $\tau(s) = \frac{s-s_{\mathrm{warm}}}{s_{\mathrm{ramp}}-s_{\mathrm{warm}}}$ and then the linear schedule is $\bar{\lambda}_{\mathrm{lin}}(s) =\bar{\lambda}_{\mathrm{start}}+ (\bar{\lambda}_{\mathrm{end}}-\bar{\lambda}_{\mathrm{start}})\tau(s)$  and the quadratic schedule is $\bar{\lambda}_{\mathrm{quad}}(s)
=\bar{\lambda}_{\mathrm{start}}+(\bar{\lambda}_{\mathrm{end}}-\bar{\lambda}_{\mathrm{start}})\tau(s)^2$.  For $s<s_{\mathrm{warm}}$, both schedules use $\bar{\lambda}_{\mathrm{start}}$, and for $s\ge s_{\mathrm{ramp}}$, both schedules use $\bar{\lambda}_{\mathrm{end}}$.

To better compare the trade-off between the data term and the physics term under different constant and scheduled choices of $\bar{\lambda}$, we also plot RE-WD and RE-JS scatter plots. These plots allow us to select the best configuration for each numerical scheme. We observe that Forward Euler consistently achieves better trade-offs across RE, WD, and JS compared with the two implicit variants. We then select the best-performing $\bar{\lambda}$ for each method and report the corresponding RE, WD, and JS in {\color{black}Figure~\ref{tab:physics_term_methods_comparison}}. The selected configurations are $\bar{\lambda}=3.0$ for Forward Euler, the piecewise schedule for Backward Euler, and the piecewise schedule for Backward Euler (limit).

All experiments use the same model architecture and training configuration as in \autoref{appedix:darcy_flow}. We train each variant for 400K steps. Overall, Forward Euler achieves the best performance for the physics term. We hypothesize that this is because the physics term has an explicit analytical form, so the steepest-descent direction provided by Forward Euler gives the most direct and stable correction.

\subsection{Data Term Numerical Integration Study}

We next study the effect of different numerical schemes in \autoref{tab:data_potential_discretization} for the data term. Similar to the physics-term study, we compare Forward Euler, Backward Euler, and the trapezoidal rule, while fixing the physics term to its best-performing configuration, namely the Forward Euler method.

\begin{figure}[H]
\centering
\begin{subfigure}[t]{0.47\linewidth}
    \centering
    \includegraphics[width=\linewidth]{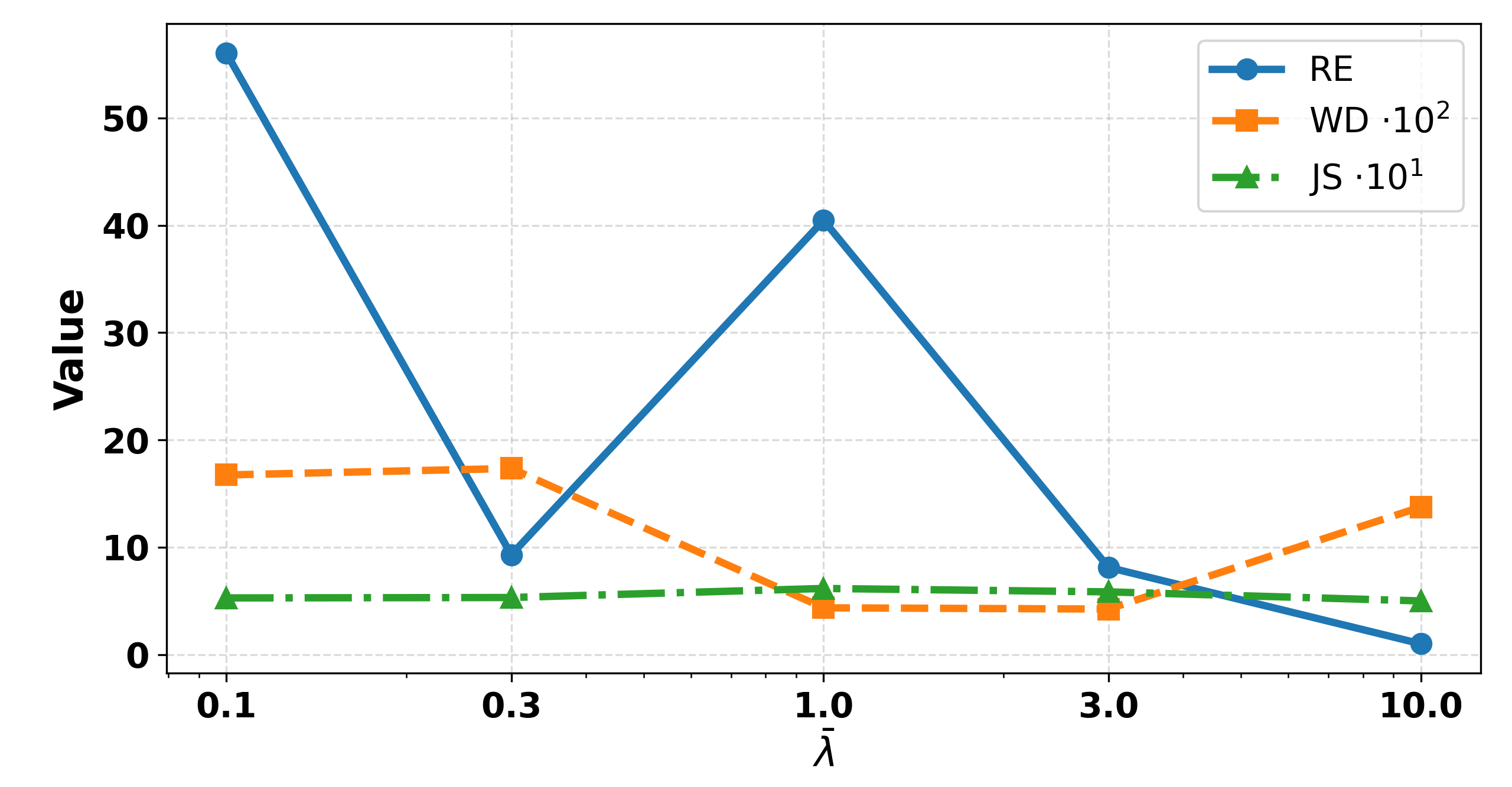}
    \caption{Forward Euler.}
    \label{fig:data_study_forward}
\end{subfigure}
\hfill
\begin{subfigure}[t]{0.47\linewidth}
    \centering
    \includegraphics[width=\linewidth]{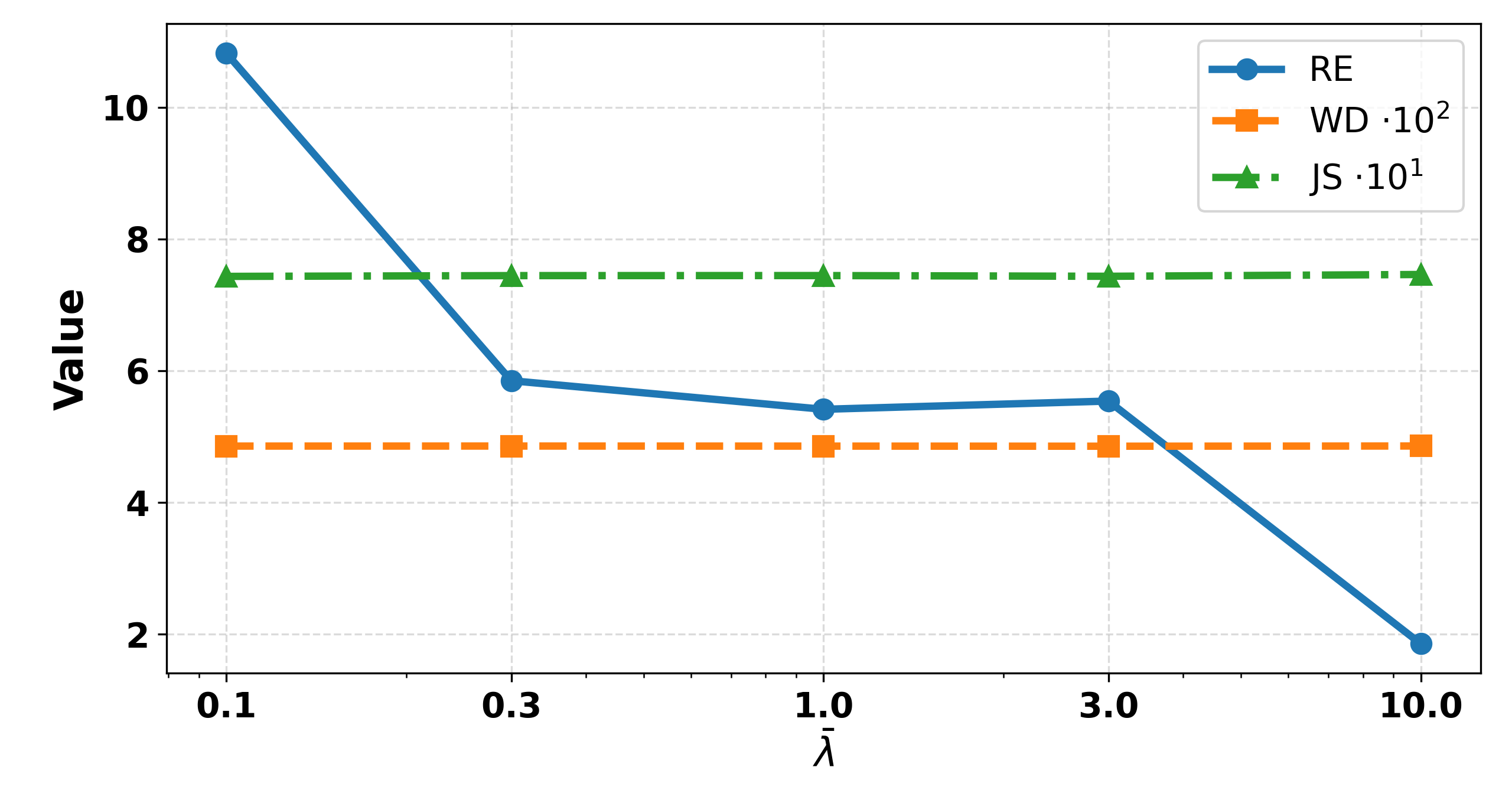}
    \caption{Backward Euler.}
    \label{fig:data_study_backward}
\end{subfigure}

\begin{subfigure}[t]{0.47\linewidth}
    \centering
    \includegraphics[width=\linewidth]{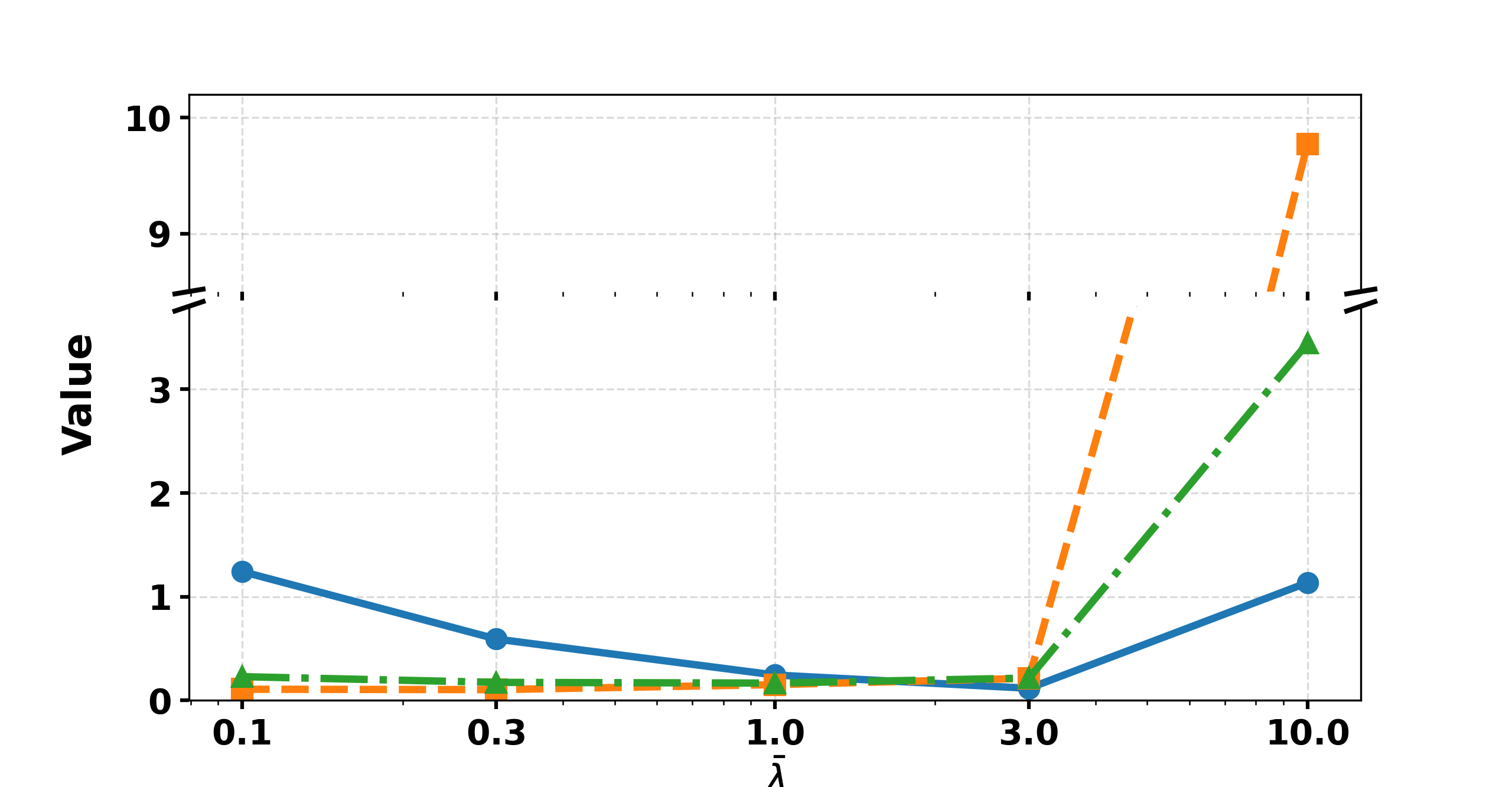}
    \caption{Trapezoidal rule.}
    \label{fig:data_study_trapezoidal}
\end{subfigure}

\caption{Effect of the coefficient $\bar \lambda$ under different numerical discretizations.}
\label{fig:data_term_lambda_study}
\end{figure}

In the experiments, we evaluate five values of the coefficient $\bar \lambda$ from $0.3$ to $10.0$, which controls the relative strength between the data term and the physics term. For each numerical scheme, we plot the trends of RE, WD, and JS as $\bar \lambda$ varies, as shown in \autoref{fig:data_term_lambda_study}. The trapezoidal rule consistently yields smaller values across all three metrics compared with the two Euler schemes. We then select the best-performing ${\color{black}\bar\lambda}$ for each method and report the corresponding RE, WD, and JS in {\color{black}Figure~\ref{tab:data_term_methods_comparison}}. The selected values are ${\color{black}\bar\lambda}=3.0$ for Forward Euler, $\bar\lambda=10.0$ for Backward Euler, and $\bar \lambda=3.0$ for the trapezoidal rule.

All experiments use the same model architecture and training configuration as in \autoref{appedix:darcy_flow}. To reduce computational cost, we train each variant for 100K steps. In contrast to the physics term, we find that the second-order trapezoidal rule consistently achieves the best performance for the data term. This can be attributed to the kernel-based expectation in the data target, where a higher-order discretization provides a more accurate approximation of the near-terminal dynamics.

\subsection{Encoder Study}\label{appendix:encoder_study}

We investigate the effect of different feature encoders used in the feature-space stabilization strategy described in \autoref{sec:algorithm}. The corresponding training procedure is summarized in \autoref{alg:feature_space_training}. We consider several representative encoder choices, including a multi-resolution MAE \cite{he2022masked} trained on the target dataset, the pretrained U-Net-based physics foundation model Poseidon \cite{herde2024poseidon}, and the transformer-based physics foundation models DPOT \cite{hao2024dpot} and MPP \cite{mccabe2024multiple}. For a fair comparison, we keep the encoder parameter counts approximately matched at around 30M parameters. For Poseidon, we use the basic variant, while for DPOT and MPP, we use their small variants to ensure comparable model sizes.

The MAE encoder is implemented as a ResNet--U-Net architecture trained directly on the Darcy dataset (for Kolmogorov flow, we use the same architecture and training setup, only adapting the input and output channels). Each input consists of two channels, which are first patchified using $2\times2$ patches, resulting in an 8-channel representation on a $32\times32$ grid. The encoder follows a ResNet34-style \cite{he2016deep} design with Group Normalization, producing multi-resolution feature maps with channel widths $64,128,256,$ and $512$. The decoder adopts a U-Net structure with bilinear upsampling and skip connections to reconstruct the input. During training, we randomly mask 50\% of the patches and optimize a mean-squared reconstruction loss on the masked regions. The model is trained with batch size 256 using AdamW, with an initial learning rate of $5\times10^{-4}$ decayed to $1.25\times10^{-4}$, weight decay 0, and $\beta=(0.9,0.999)$, together with an exponential moving average with decay 0.9995. The MAE is trained for 400K steps, after which it is frozen and used as the feature encoder.

\begin{figure}[t]
\centering
\begin{subfigure}[t]{0.48\linewidth}
    \centering
    \includegraphics[width=1.1\linewidth]{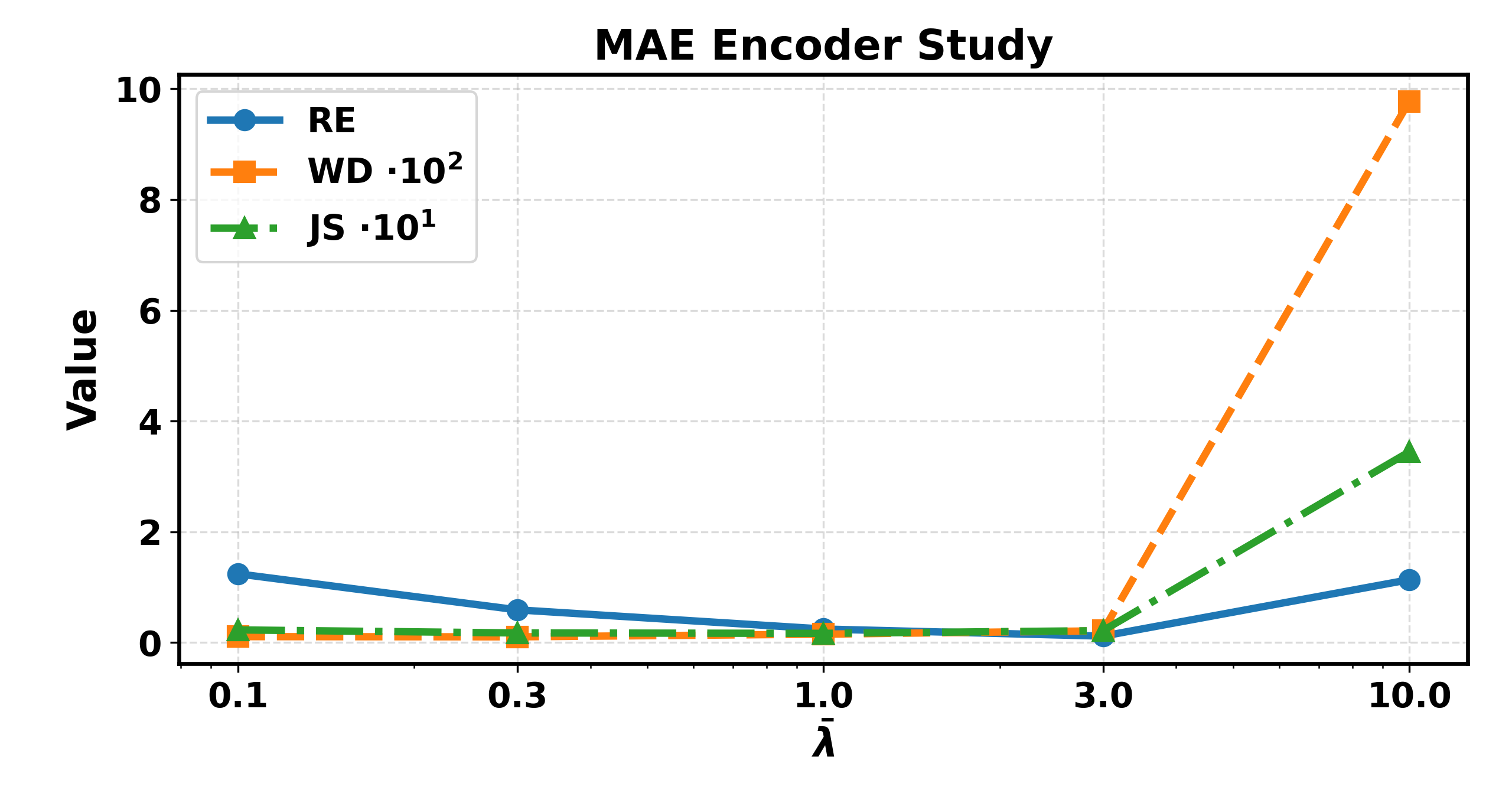}
    \caption{MAE.}
    \label{fig:encoder_mae_lambda}
\end{subfigure}
\hfill
\begin{subfigure}[t]{0.48\linewidth}
    \centering
    \includegraphics[width=1.1\linewidth]{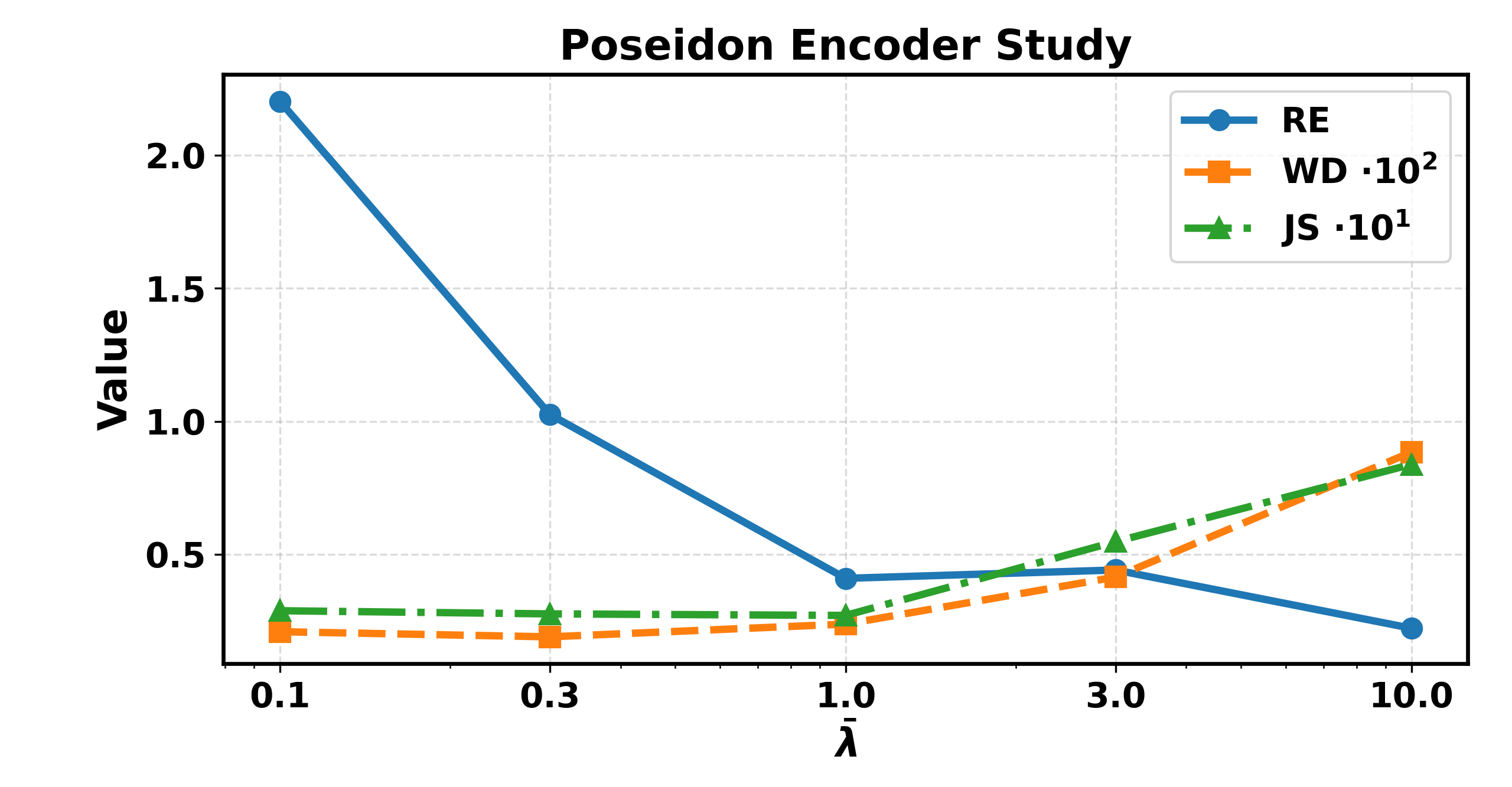}
    \caption{Poseidon.}
    \label{fig:encoder_poseidon_lambda}
\end{subfigure}

\vspace{2mm}

\begin{subfigure}[t]{0.48\linewidth}
    \centering
    \includegraphics[width=1.1\linewidth]{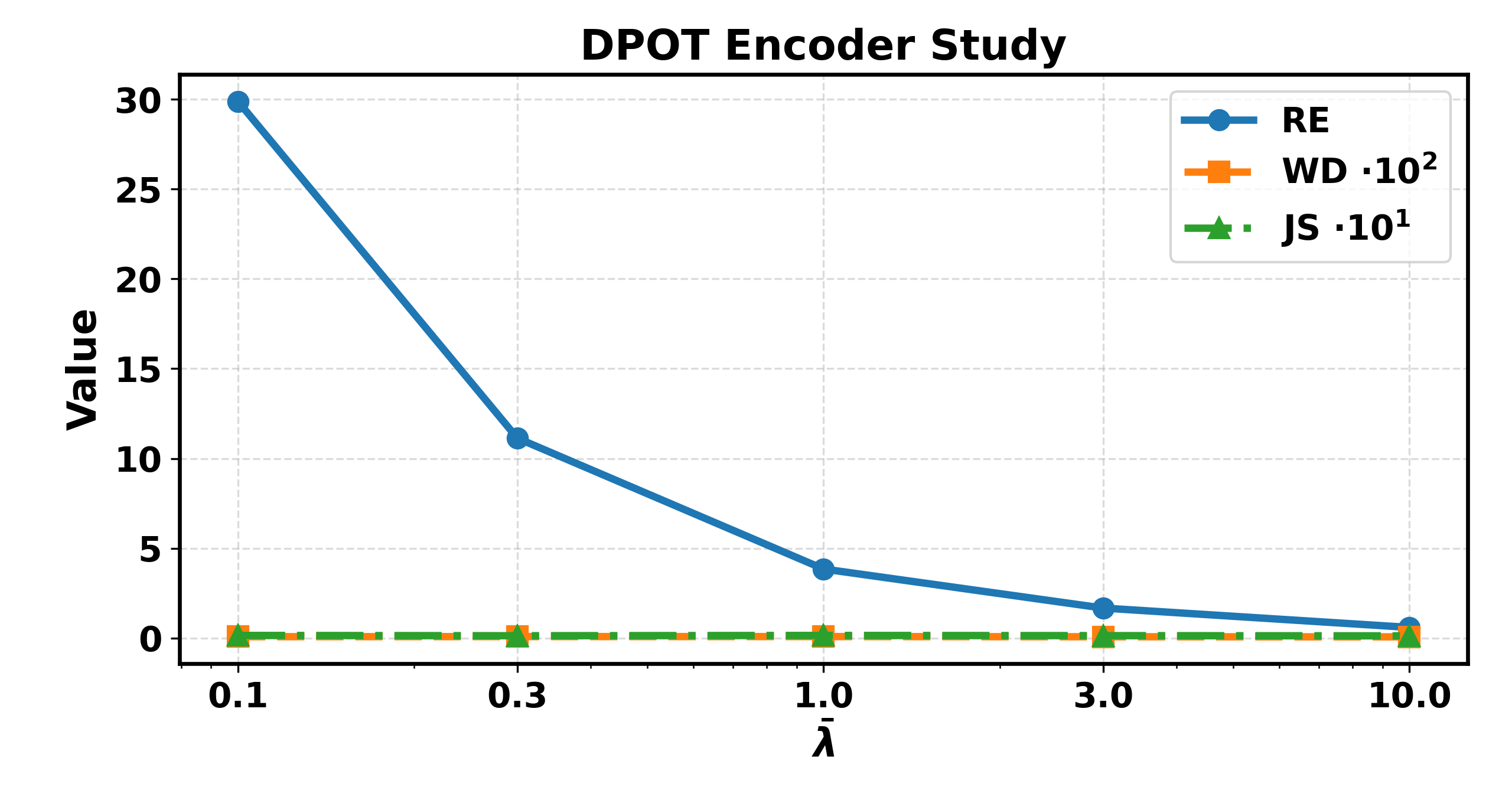}
    \caption{DPOT.}
    \label{fig:encoder_dpot_lambda}
\end{subfigure}
\hfill
\begin{subfigure}[t]{0.48\linewidth}
    \centering
    \includegraphics[width=1.1\linewidth]{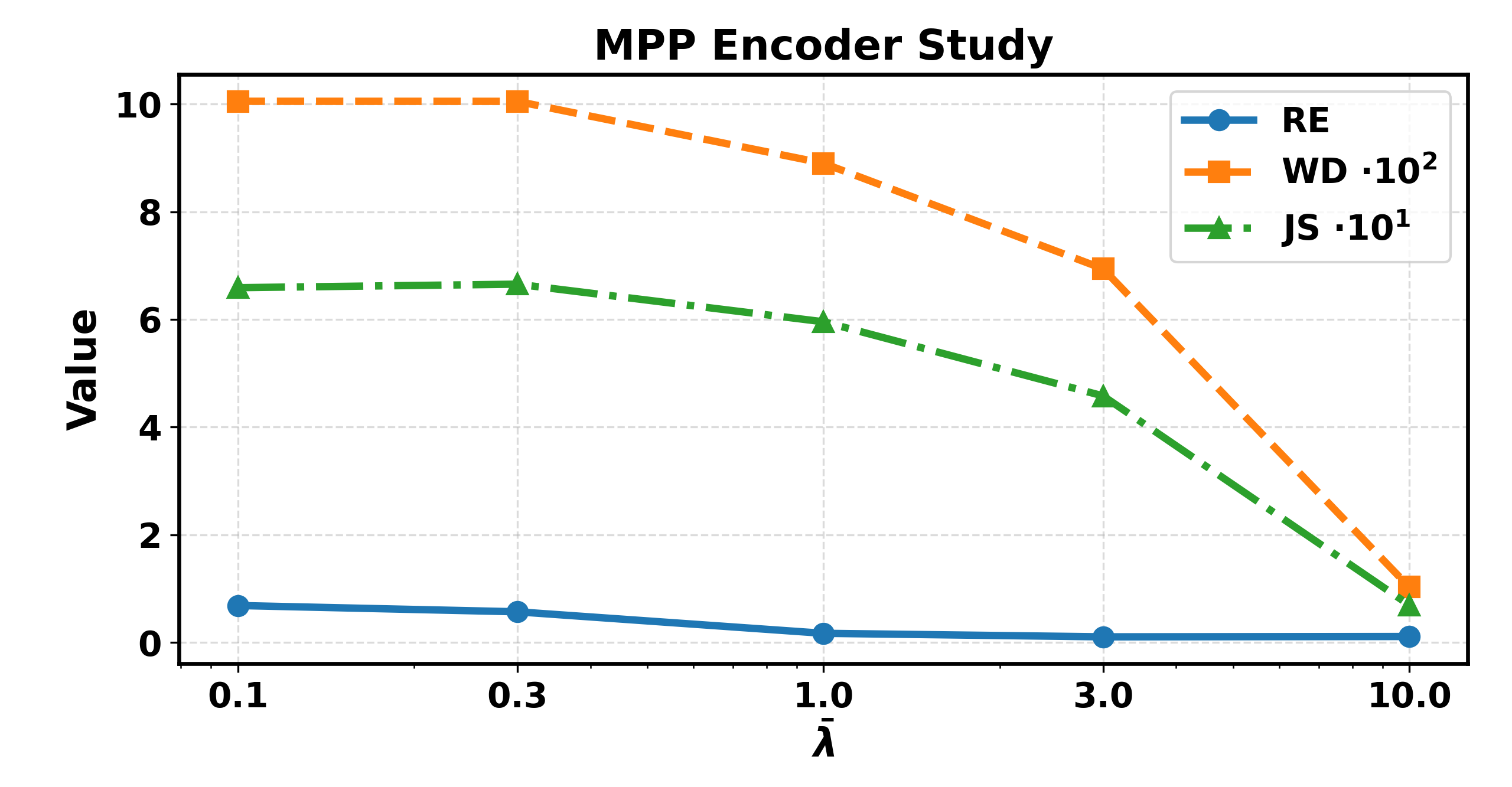}
    \caption{MPP.}
    \label{fig:encoder_mpp_lambda}
\end{subfigure}

\caption{Effect of the coefficient $\bar{\lambda}$ for different feature encoders. Each plot shows RE, WD, and JS as $\bar{\lambda}$ varies.}
\label{fig:encoder_lambda_study}
\end{figure}

We keep the neural representation of the flow map and the training configuration the same as in \autoref{appedix:darcy_flow}. To reduce computational cost, each encoder variant is trained for 100K steps within the PI-LSFM framework. In the experiments, we evaluate five constant values of the coefficient $\bar{\lambda}$ from $0.1$ to $10.0$, which controls the relative strength between the data term and the physics term. For each encoder, we plot the trends of RE, WD, and JS as $\bar{\lambda}$ varies, as shown in \autoref{fig:encoder_lambda_study}. These plots allow us to select the best configuration for each encoder and compare different encoders under their respective best-performing settings. We then report the corresponding RE, WD, JS, and per-step training time in {\color{black}Figure~\ref{tab:feature_extractor}}. The selected configurations use $\bar{\lambda}=3.0$ for MAE, $\bar{\lambda}=1.0$ for Poseidon, and $\bar{\lambda}=10.0$ for DPOT and MPP.

We observe that all encoders improve the performance of our model, demonstrating the effectiveness and robustness of feature-space training. Among them, MAE, Poseidon, and DPOT achieve the strongest overall results. Since MAE provides the fastest training while maintaining strong performance, we use it as the default encoder in our subsequent experiments. Nevertheless, for settings where training a task-specific encoder is undesirable, pretrained encoders such as Poseidon and DPOT remain strong alternatives.

To construct the feature space, we avoid directly flattening the full encoder feature maps, as this would lead to prohibitively high dimensionality. Instead, following \cite{deng2026generative}, we aggregate the encoder outputs into a compact set of multi-scale statistics. For each feature map of size $H_i\times W_i\times C_i$, we retain location-wise features, global channel-wise mean and standard deviation, and channel-wise mean and standard deviation computed over spatial patches at multiple resolutions. This design preserves both local and global information while significantly reducing the dimensionality of the feature representation.

\begin{figure}[H]
\centering

\newcommand{\darcyrow}[1]{
\begin{subfigure}[t]{0.31\linewidth}
    \centering
    \includegraphics[width=\linewidth]{images/darcy_failure/#1_pressure.png}
\end{subfigure}
\hfill
\begin{subfigure}[t]{0.31\linewidth}
    \centering
    \includegraphics[width=1.07\linewidth]{images/darcy_failure/#1_permeability.png}
\end{subfigure}
\hfill
\begin{subfigure}[t]{0.31\linewidth}
    \centering
    \includegraphics[width=1.05\linewidth]{images/darcy_failure/#1_residual.png}
\end{subfigure}

\vspace{2mm}
}

\darcyrow{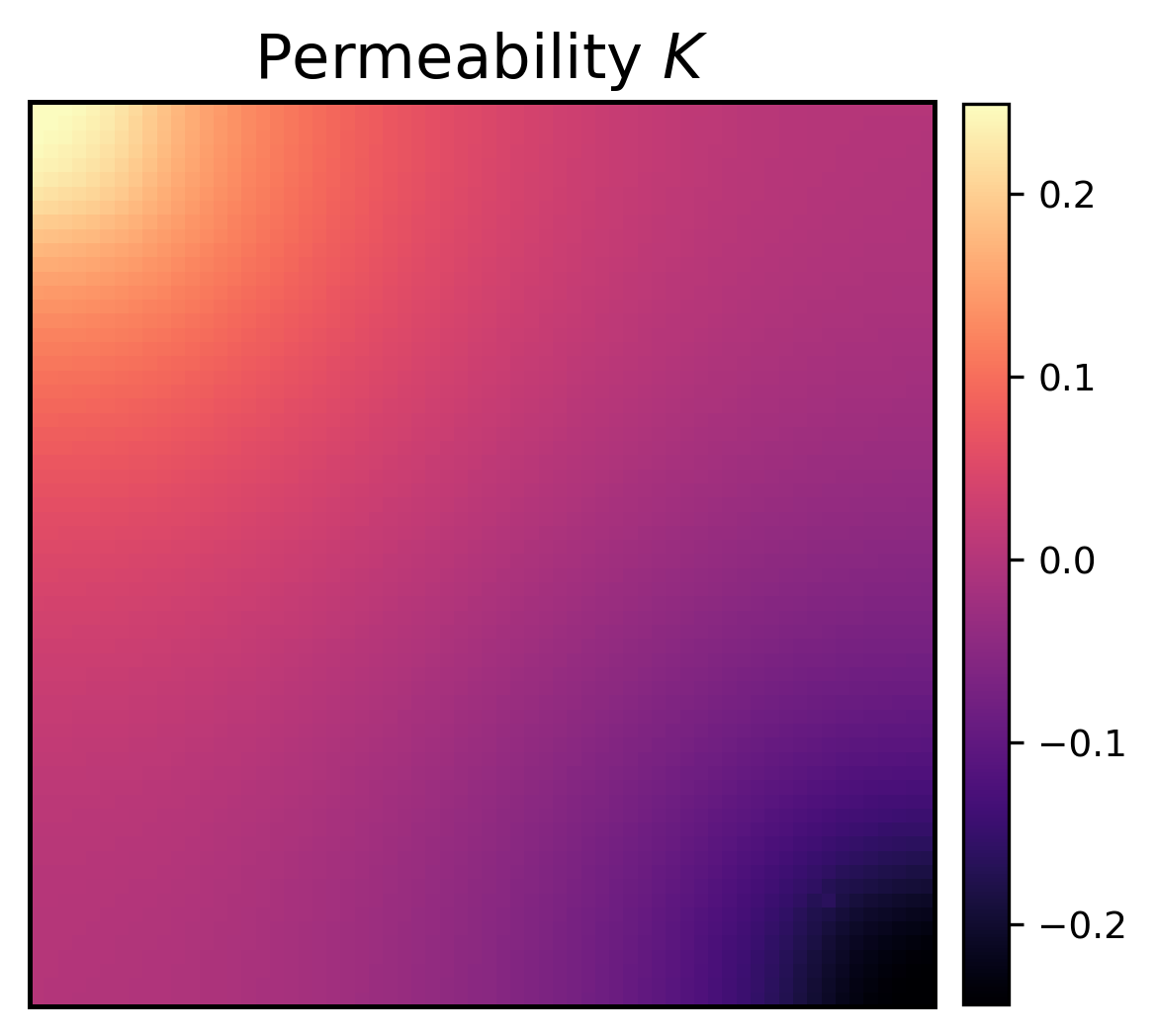}
\darcyrow{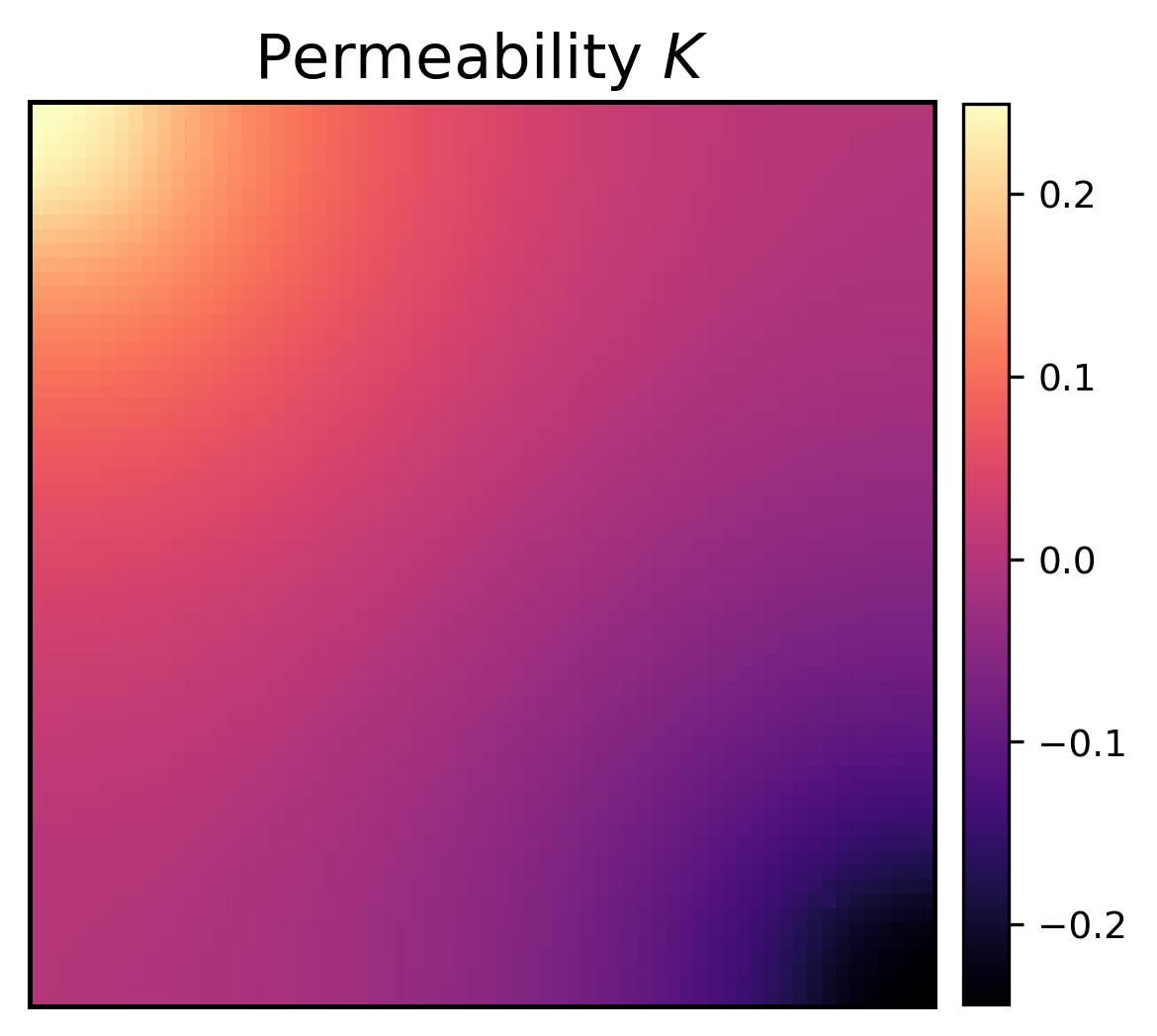}

\caption{Failure cases without feature-space optimization on Darcy flow. Each row shows one generated sample, with columns corresponding to pressure $p$, permeability $K$, and residual. Without a feature encoder, optimization fails to produce physically meaningful results.}
\label{fig:darcy_no_feature_encoder}
\end{figure}

In \autoref{fig:darcy_no_feature_encoder}, we demonstrate the necessity of feature-space optimization: without a feature encoder, training fails, in contrast to the successful feature-space results in \autoref{fig:darcy_generated_5x3}.

\section{Experiment Details and Additional Results}

\subsection{Toy Example}\label{appedix:toy_example}

\paragraph{Dataset} To clearly demonstrate the advantage of our direct loss in jointly controlling the physics constraint and the data distribution, we consider a two-dimensional toy dataset for intuitive visualization. Specifically, we construct $\mathcal{D}=\{(x_i,y_i)\}_{i=1}^n$ from a mixture of three Gaussian distributions whose centers are approximately located on the unit circle. The underlying physical constraint is $\|(x,y)\|^2=1$.

\begin{figure}[t]
\centering

\begin{subfigure}[t]{0.31\linewidth}
    \centering
    \includegraphics[width=\linewidth]{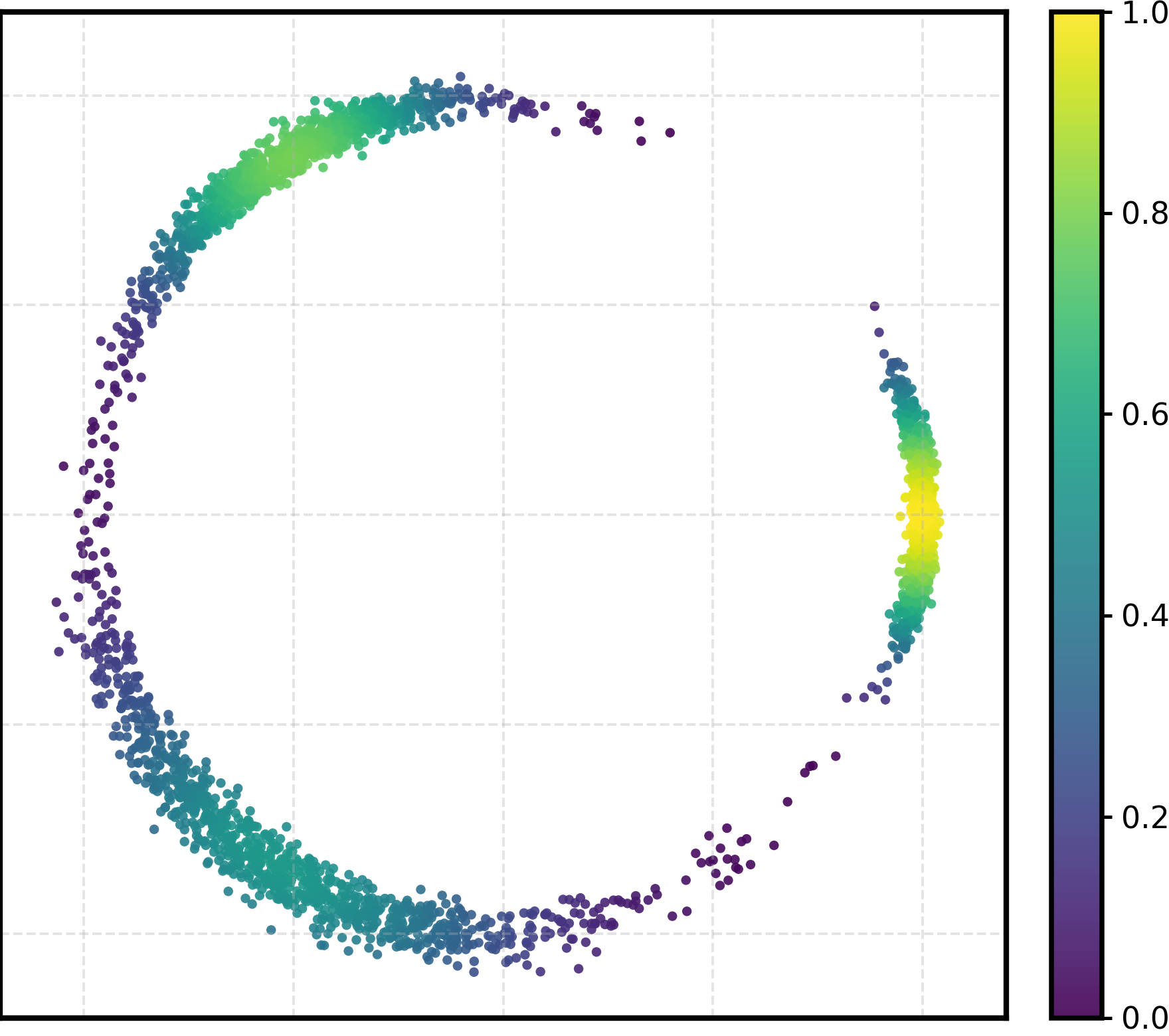}
    \caption{Dataset}\label{fig:toy_dataset}
\end{subfigure}
\hfill
\begin{subfigure}[t]{0.31\linewidth}
    \centering
    \includegraphics[width=\linewidth]{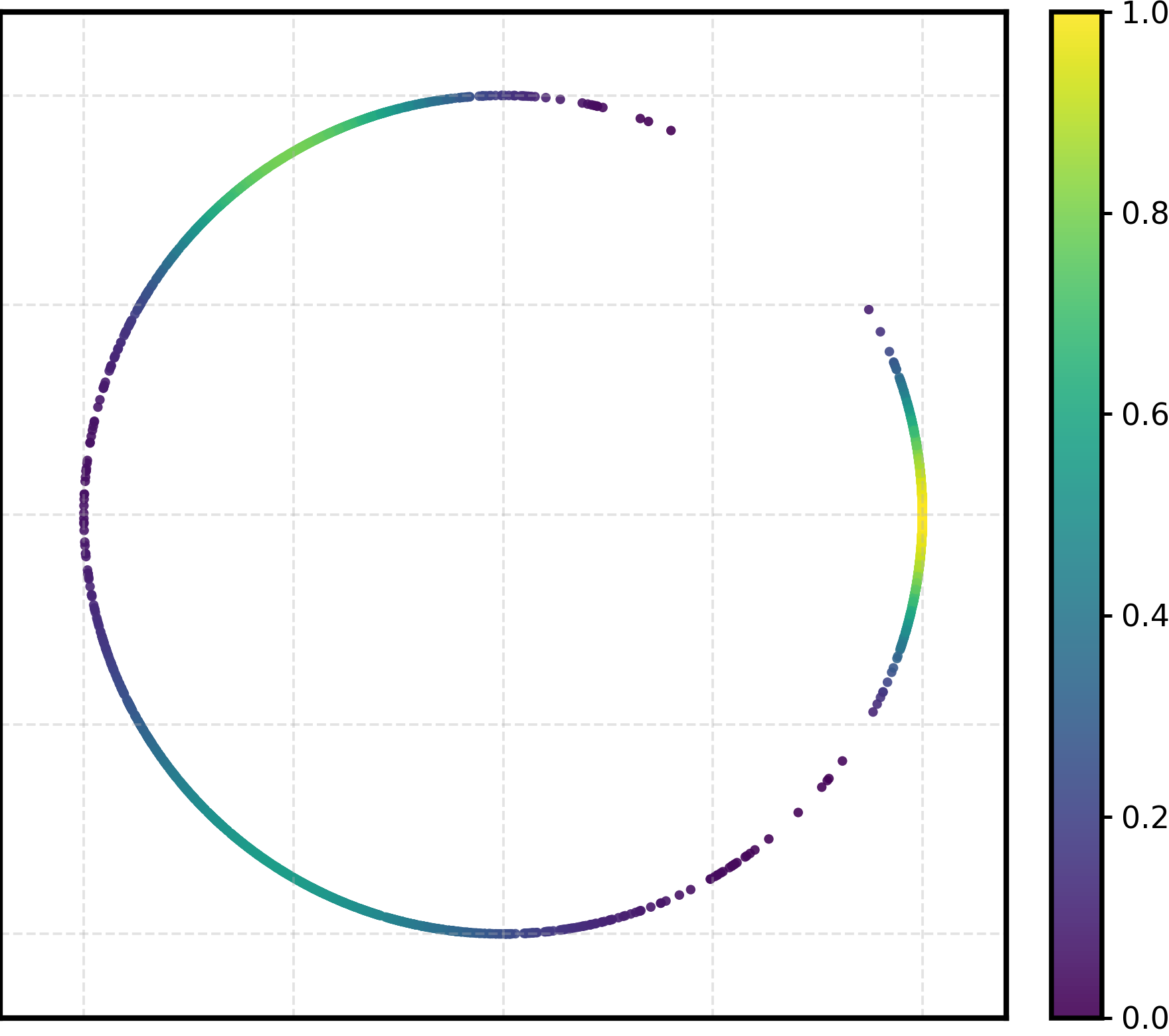}
    \caption{Full-projected dataset}\label{fig:toy_proj_dataset}
\end{subfigure}
\hfill
\begin{subfigure}[t]{0.31\linewidth}
    \centering
    \includegraphics[width=\linewidth]{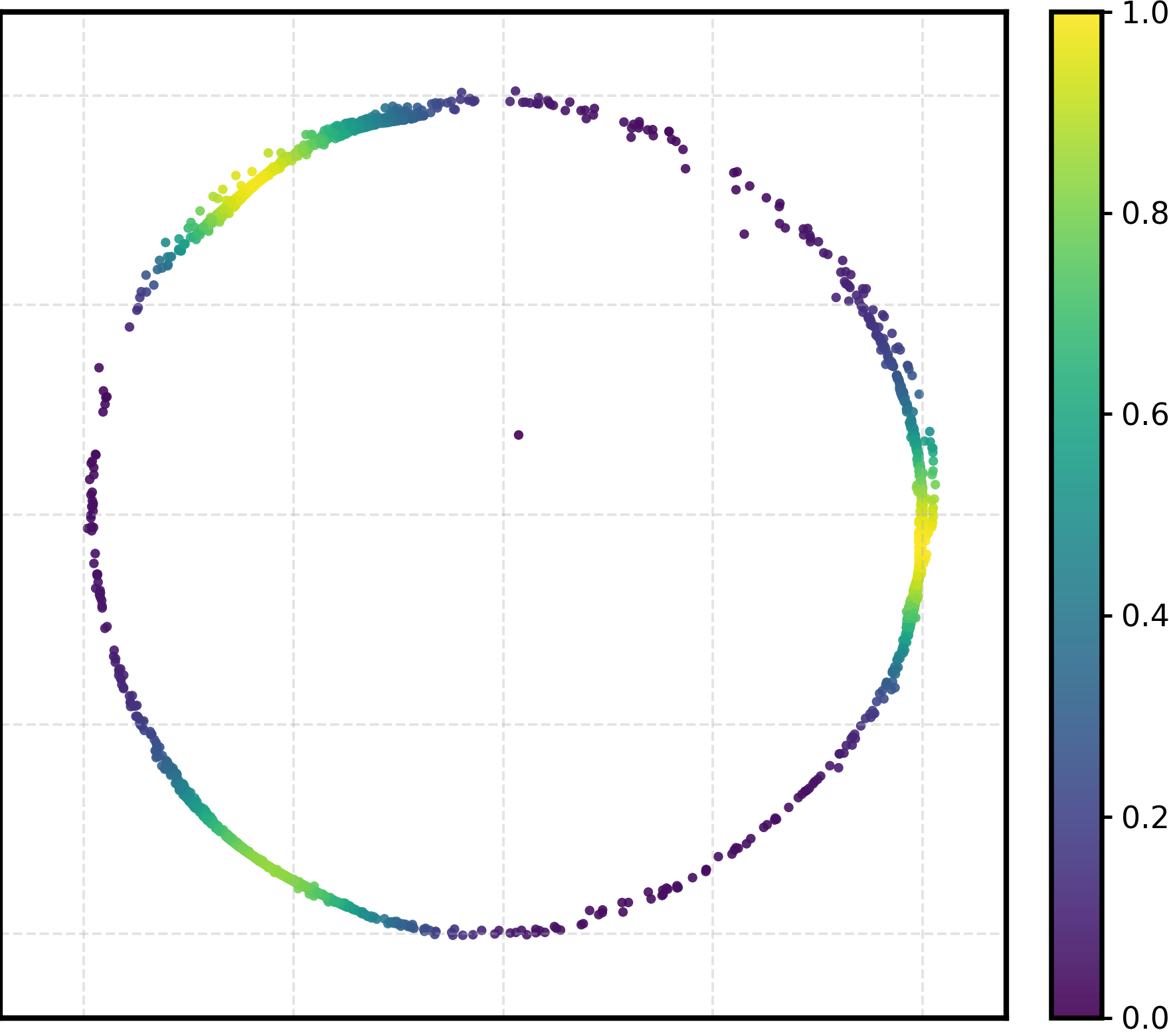}
    \caption{Our method}\label{fig:toy_our_method}
\end{subfigure}

\vspace{2mm}

\begin{subfigure}[t]{0.31\linewidth}
    \centering
    \includegraphics[width=\linewidth]{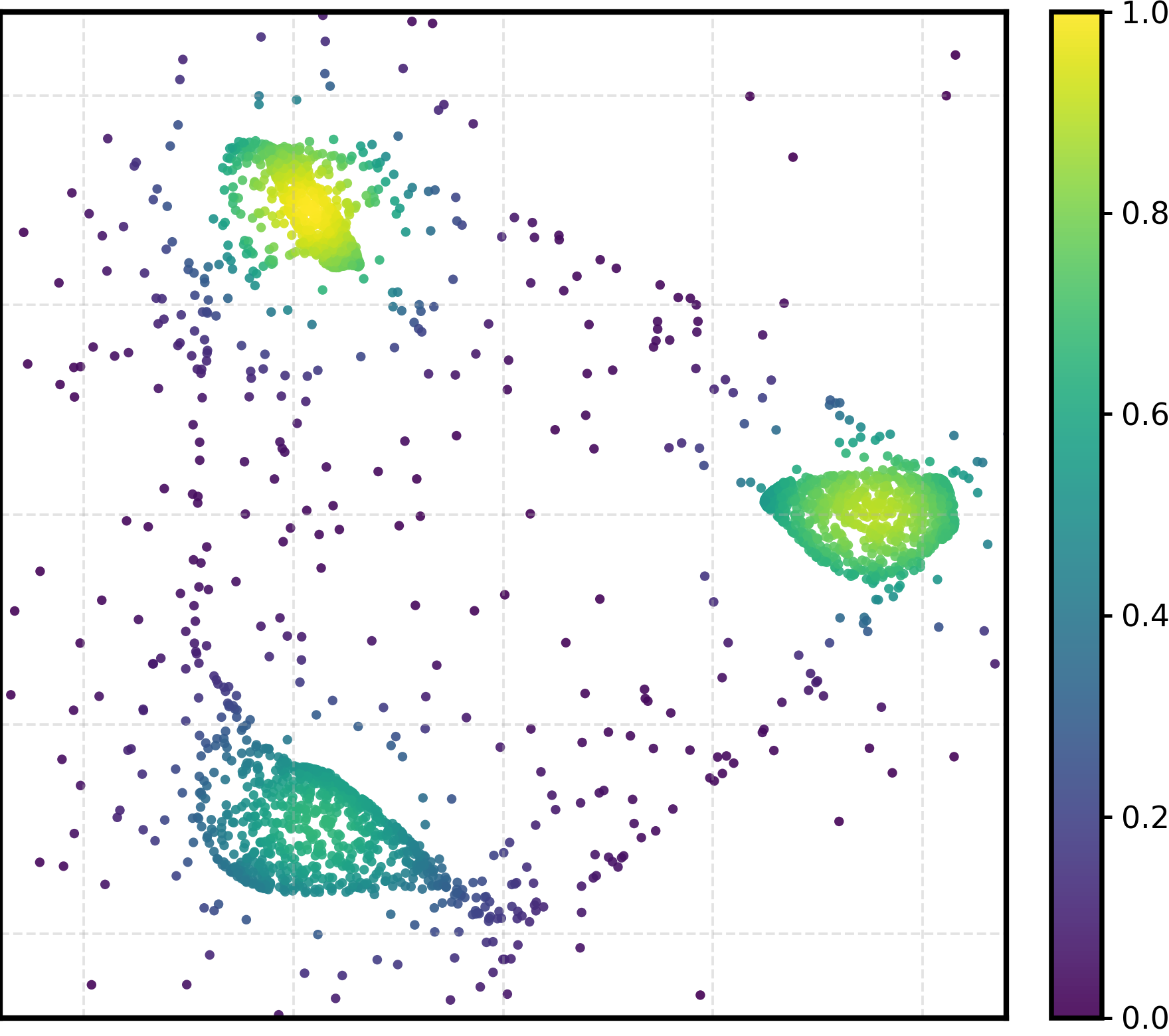}
    \caption{sCM ($\bar{\lambda}=0.0$)}\label{fig:toy_CM1}
\end{subfigure}
\hfill
\begin{subfigure}[t]{0.31\linewidth}
    \centering
    \includegraphics[width=\linewidth]{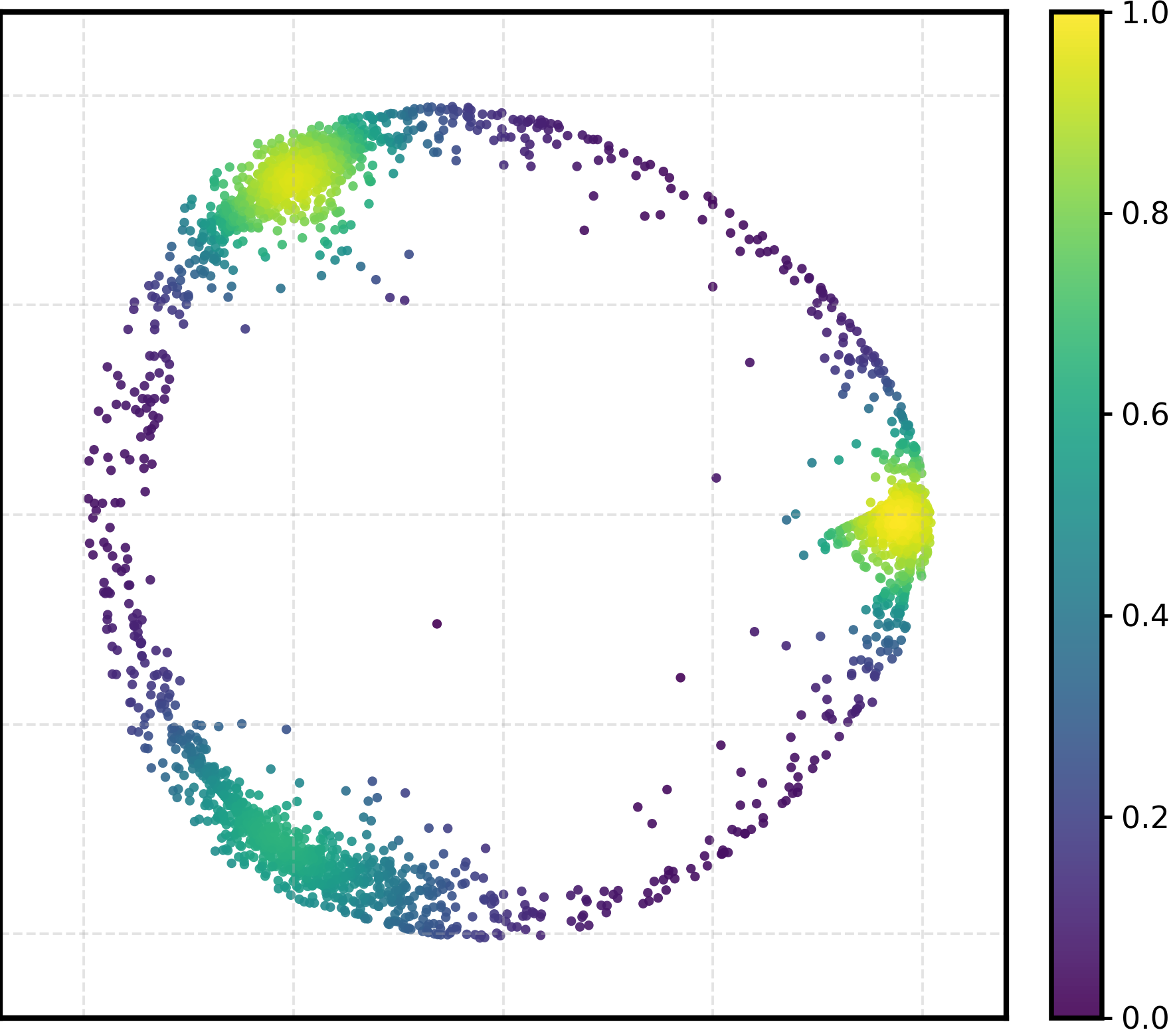}
    \caption{sCM ($\bar{\lambda}=1.0$)}\label{fig:toy_CM2}
\end{subfigure}
\hfill
\begin{subfigure}[t]{0.31\linewidth}
    \centering
    \includegraphics[width=\linewidth]{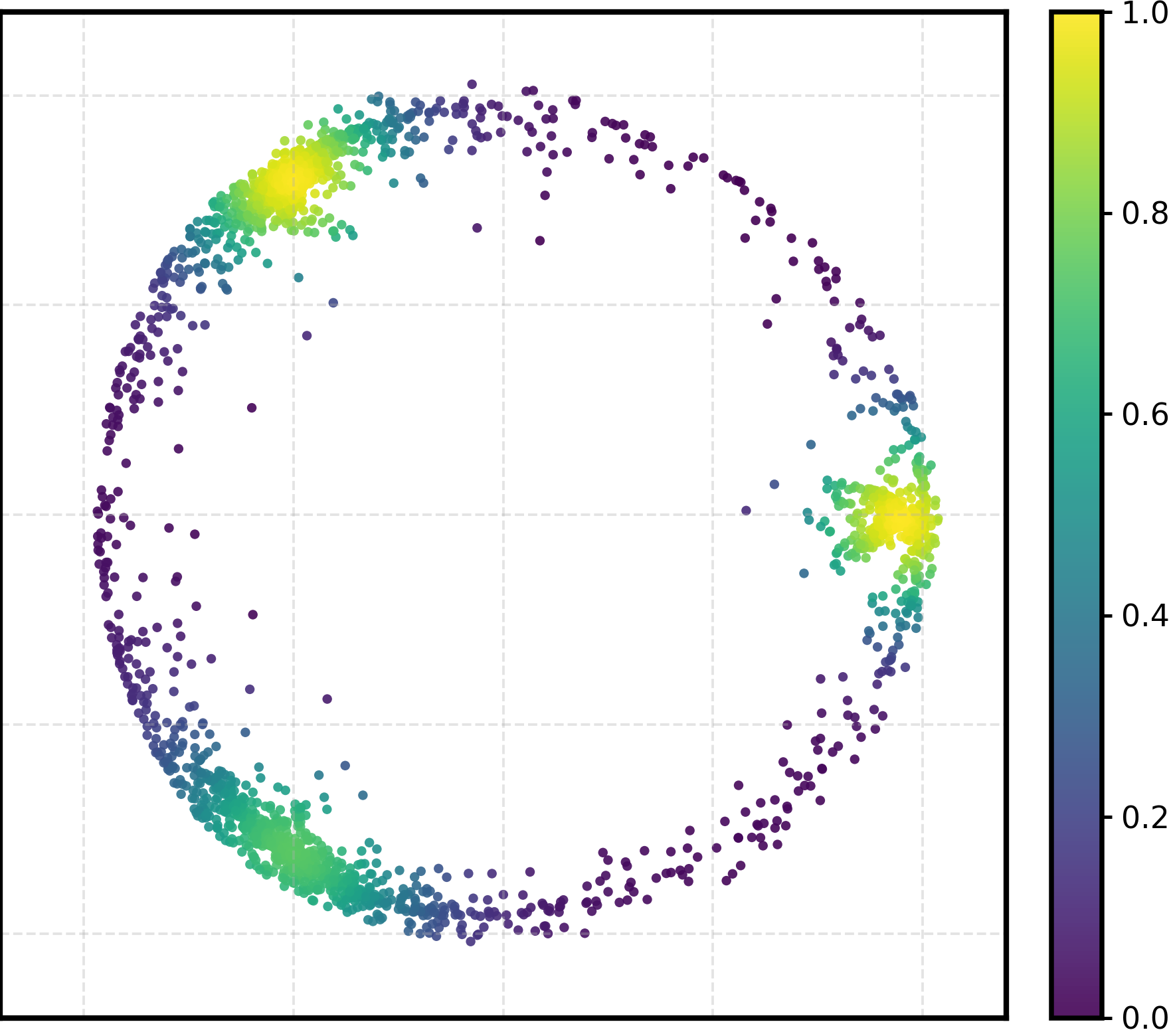}
    \caption{sCM ($\bar{\lambda}=3.0$)}\label{fig:toy_CM3}
\end{subfigure}

\vspace{2mm}

\begin{subfigure}[t]{0.31\linewidth}
    \centering
    \includegraphics[width=\linewidth]{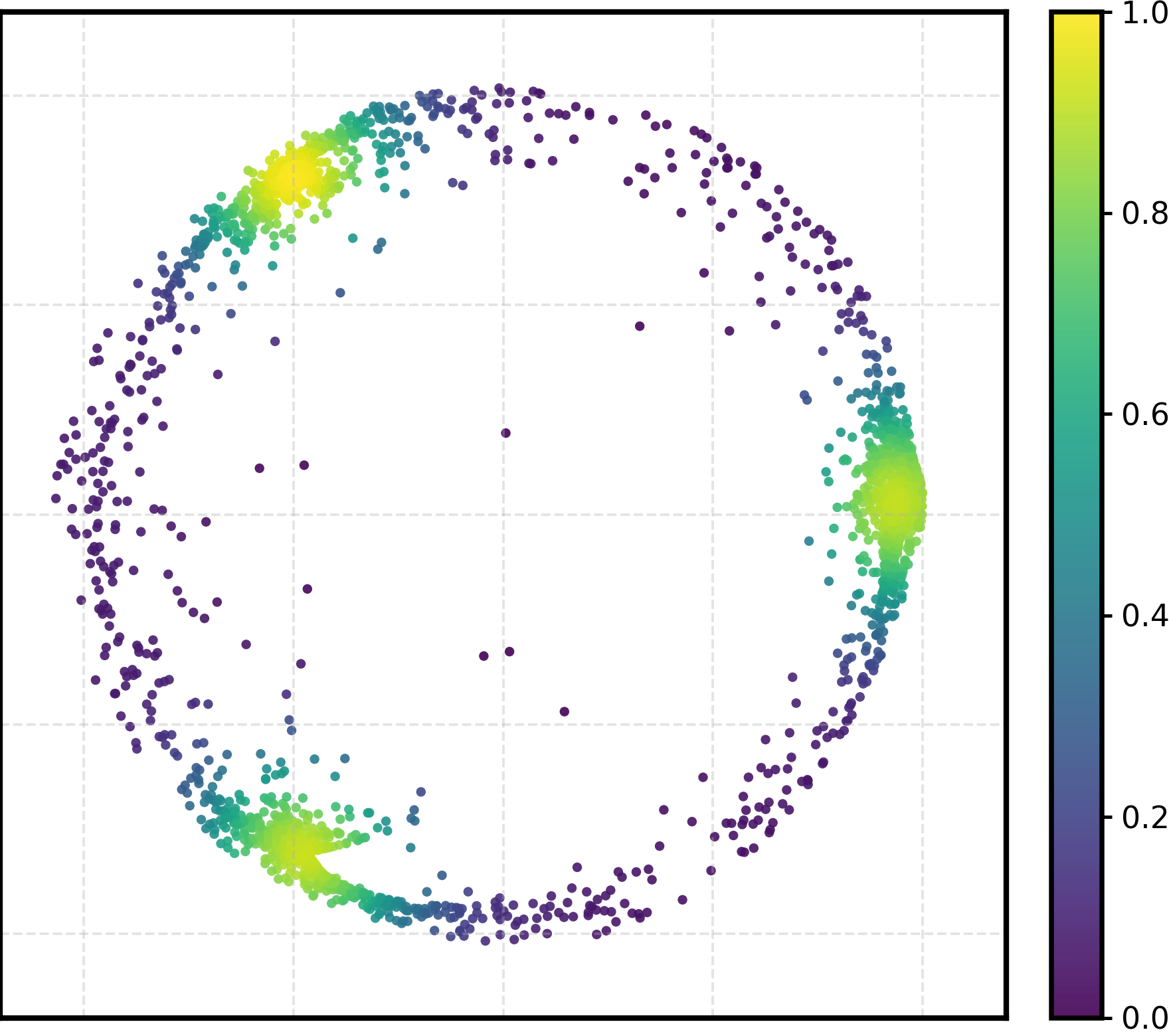}
    \caption{sCM ($\bar{\lambda}=10.0$)}\label{fig:toy_CM4}
\end{subfigure}
\hfill
\begin{subfigure}[t]{0.31\linewidth}
    \centering
    \includegraphics[width=\linewidth]{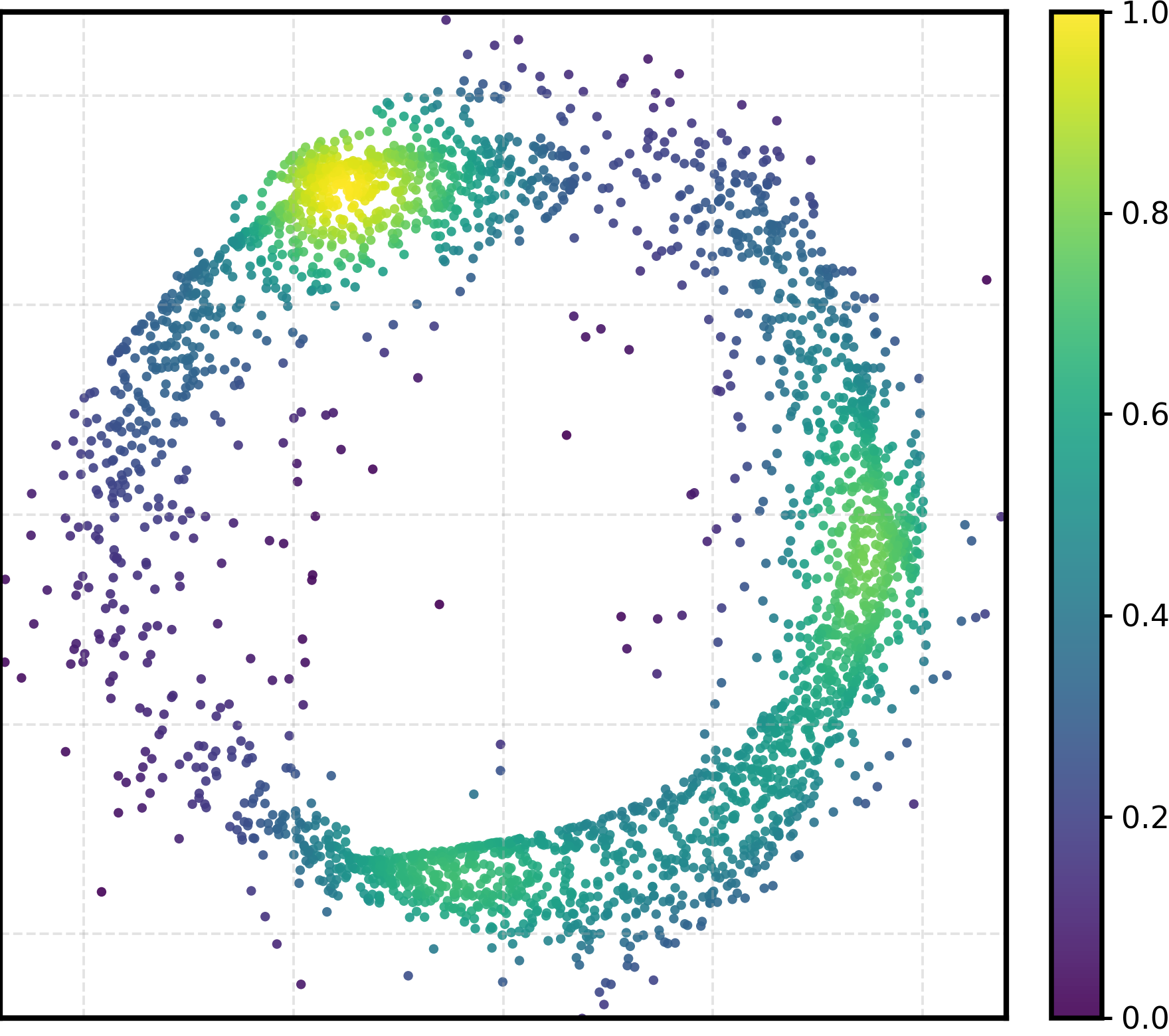}
    \caption{sCM ($\bar{\lambda}=30.0$)}\label{fig:toy_CM5}
\end{subfigure}
\hfill
\begin{subfigure}[t]{0.31\linewidth}
    \centering
    \includegraphics[width=\linewidth]{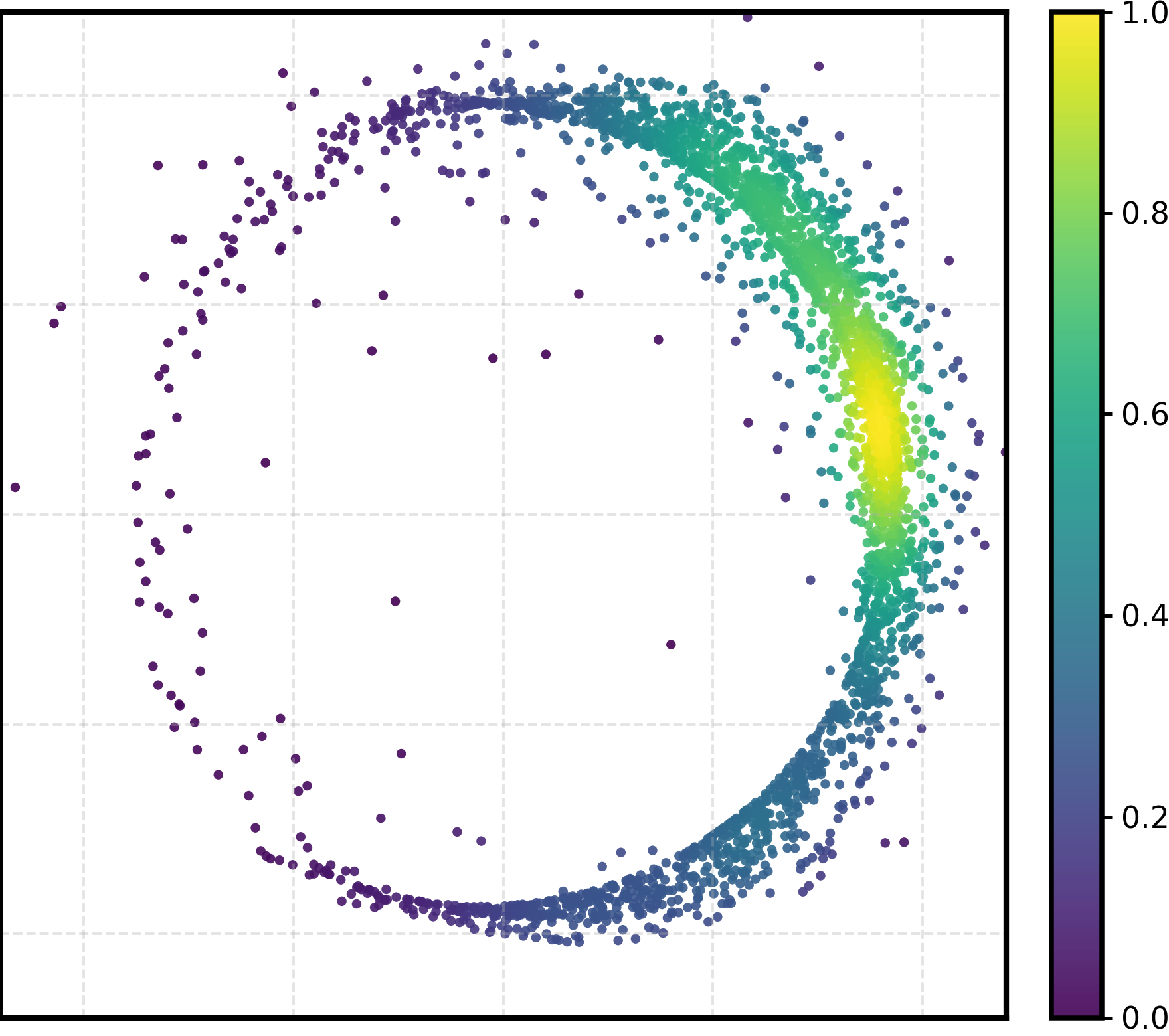}
    \caption{sCM ($\bar{\lambda}=100.0$)}\label{fig:toy_CM6}
\end{subfigure}
\caption{Toy example visualization comparing the training dataset, the fully projected target distribution, our method, and physics-informed sCM with different values of $\bar{\lambda}$. Our method preserves the three-mode structure while satisfying the circle constraint, and achieves up to approximately $30\times$ lower physical residual error than sCM, as reported in \autoref{tab:toy_constraint_satisfaction}.}
\label{fig:toy_3x3}
\end{figure}

To simulate a realistic setting where data samples are influenced by both the data distribution and the physical constraint, we do not fully project the samples onto the circle. Instead, we introduce controlled deviations by interpolating between the original Gaussian samples and their projections $x_{\mathrm{final}}=(1-\alpha)x_{\mathrm{raw}}+\alpha x_{\mathrm{proj}}$, where $\alpha=0.9$ in all experiments. This produces training samples that approximately satisfy the constraint while preserving the multimodal structure of the data. An example of this training dataset is shown in \autoref{fig:toy_dataset}, while the fully projected dataset in \autoref{fig:toy_proj_dataset} serves as the desired physics-consistent distribution.

\paragraph{Training Details}  On this dataset, we compare our method with a single-stage consistency model (sCM \cite{lu2024simplifying}). For sCM, we use the standard consistency objective $\mathcal{L}_{\mathrm{sCM}} = \mathbb{E} |[|\|F_\theta(x_{t_1},t_1)-\mathrm{sg}\!|(F_\theta(x_{t_2},t_2)|)|\|^2|]+\bar\lambda\mathbb{E}|[\|\mathcal{R}\!(f_\theta(x_t,t)|)\|^2]$, where $F_\theta$ denotes the raw neural-network output and $f_\theta$ denotes the corresponding consistency output used for generation. $\mathcal{R}(x,y)=x^2+y^2-1$ is the physics residual for the circle constraint. Except for the training loss, our method and sCM use the same model architecture and optimization setup. We use a MLP network as the backbone. The network has 8 layers with hidden dimension 128 and sinusoidal time embeddings of dimension 32. Both methods are trained for {\color{black}40K} steps using AdamW with learning rate $5\times10^{-5}$, weight decay $0$, and $\beta=(0.9,0.999)$. {\color{black}We use exponential moving average (EMA) with decay $0.9999$.}

\paragraph{Results} We visualize the learned distributions in \autoref{fig:toy_3x3} using a small batch size $B=3$, which highlights whether each method can simultaneously recover the multimodal distribution and satisfy the circle constraint. For sCM, we test multiple values of the physics weight $\lambda$ for a fair comparison.  As shown in {\color{black}Figures~\ref{fig:toy_CM1}--\ref{fig:toy_CM6}}, physics-informed consistency models struggle to balance the data distribution and the constraint: small $\lambda$ leads to poor constraint satisfaction, while large $\lambda$ causes the physics term to dominate and often results in mode collapse. In contrast, our method in \autoref{fig:toy_our_method} enforces the geometric constraint while preserving the three modes along the circle, demonstrating effective joint control of distributional fidelity and physical validity. We further report constraint satisfaction under different batch sizes, including $B=3,4,8$, in \autoref{tab:toy_constraint_satisfaction}. Our method achieves significantly lower residual errors, outperforming sCM by up to approximately $30\times$ across different batch sizes.

\begin{table}[t]
\centering
\caption{Constraint satisfaction on the toy example under different batch sizes. We report RE: physical Residual MSE $\times 10^{4}$.}
\label{tab:toy_constraint_satisfaction}
\small
\resizebox{\linewidth}{!}{%
\begin{tabular}{lccccccc}
\toprule
Batch size & Ours & sCM ($\bar{\lambda}=0$) & sCM ($\bar{\lambda}=1$) & sCM ($\bar{\lambda}=3$) & sCM ($\bar{\lambda}=10$) & sCM ($\bar{\lambda}=30$) & sCM ($\bar{\lambda}=100$) \\
\midrule
$B=3$ & 8.015 & 3996.970 & 220.986 & 176.003 & 190.425 & 773.622 & 240.384 \\
$B=4$ & 3.040 & 574.162 & 198.419 & 155.516 & 146.720 & 142.761 & 98.074 \\
$B=8$ & 0.722 & 366.265 & 158.029 & 142.076 & 68.378 & 38.710 & 33.033 \\
\bottomrule
\end{tabular}%
}
\end{table}

\subsection{Darcy Flow}\label{appedix:darcy_flow}

\paragraph{Dataset and constraints.}
We use the Darcy-flow dataset following \cite{baldan2026physics} to enable direct comparison with prior physics-informed generative baselines. The task models steady-state flow in a heterogeneous porous medium on a $64\times64$ grid, where each sample contains a permeability field $K$ and the corresponding pressure field $p$. The flow is governed by
\begin{equation}
\begin{aligned}
u(x) &= -K(x)\nabla p(x), \qquad x\in\Omega,\\
\nabla\cdot u(x) &= f(x), \qquad x\in\Omega,\\
u(x)\cdot \hat n(x) &= 0, \qquad x\in\partial\Omega,
\end{aligned}
\end{equation}
where $u$ is the Darcy velocity, $\hat n$ is the outward unit normal on the boundary, and $x=(x_1,x_2)$ denotes the spatial coordinate. Equivalently, the pressure satisfies $\nabla\cdot(K\nabla p)+f=0$ with a no-flux boundary condition.  Following \cite{baldan2026physics}, the forcing term is a localized source--sink pair:
\begin{equation}
\begin{aligned}
f(x)
=
\begin{cases}
r, & \text{if } |x_i-\frac{1}{2}w|\le \frac{1}{2}w, i=1,2\\
-r, & \text{if } |x_i-1+\frac{1}{2}w|\le \frac{1}{2}w, i=1,2\\
0, & \text{otherwise},
\end{cases}
\end{aligned}
\end{equation}
with $r=10$, $w=0.125$ and $x = (x_1,x_2)$ is the 2D coordination of $x$. The permeability field is generated as $K(x)=\exp(G(x))$, where $G(x)$ is a Gaussian random field. Given $K$ and $f$, the pressure field $p$ is obtained by solving a least-squares problem using a 64-term truncated Karhunen-Loève expansion, following \cite{jacobsen2025cocogen}. Since the no-flux boundary condition determines $p$ only up to an additive constant, the pressure field is normalized by removing its spatial mean.

For the physics constraint, we follow the residual-evaluation protocol of \cite{baldan2026physics} and use both the PDE residual and the no-flux boundary residual:
\begin{equation}
\begin{aligned}
\mathcal{R}_{\mathrm{pde}}(x)
&=
\nabla\cdot(K\nabla p)(x)+f(x),\\
\mathcal{R}_{\mathrm{bc}}(x)
&=
u(x)\cdot \hat n(x).
\end{aligned}
\end{equation}
In practice, the differential operators are evaluated using fixed finite-difference kernels implemented as 2D convolutional filters, and the forcing field is reconstructed with the same procedure used during dataset generation. The generated pressure field is likewise mean-normalized before evaluating the residuals. This benchmark is unconditional, so no external condition is provided to the generator.

\paragraph{Metrics.}
We evaluate both physical consistency and distributional fidelity. The residual error (RE) measures the mean squared violation of the governing equation and no-flux boundary condition:
\begin{equation}
\begin{aligned}
\mathrm{RE}
&=
\mathbb{E}_{x\in\Omega}\![\mathcal{R}_{\mathrm{pde}}(x)^2]
+
\mathbb{E}_{x\in\partial\Omega}\![\mathcal{R}_{\mathrm{bc}}(x)^2].
\end{aligned}
\end{equation}
For distributional fidelity, we follow \cite{baldan2026physics} and report Wasserstein distance (WD) \cite{villani2009optimal} and Jensen-Shannon divergence (JS) \cite{lin2002divergence}, computed using the same evaluation protocol and implementations with the SciPy library \cite{gommers2025scipy}.  We also report the number of function evaluations (NFE) and inference time (IT) to measure sampling efficiency.

\paragraph{Model and training.}
We use the basic architecture described in \cite{bastek2024physics} without additional architectural optimization. The number of parameters is kept consistent with \cite{baldan2026physics}, which serves as our main baseline. For experiments with feature-space stabilization, the encoder is used only during training to compute the data target, while generation is still performed in the original data space. The training details of the feature encoder are provided in \autoref{appendix:encoder_study}, and the architecture and training configuration of the main model after encoder pretraining are summarized in \autoref{tab:hyperparameters}. Note that our method does not require intermediate time inputs $t$; therefore, for unconditional generation, we remove the time embedder and conditional modulation modules.

\begin{figure}[H]
\centering

\newcommand{\darcyrow}[2]{
\begin{subfigure}[t]{0.31\linewidth}
    \centering
    \includegraphics[width=\linewidth]{images/darcy/#1_pressure.png}
\end{subfigure}
\hfill
\begin{subfigure}[t]{0.31\linewidth}
    \centering
    \includegraphics[width=1.07\linewidth]{images/darcy/#1_permeability.png}
\end{subfigure}
\hfill
\begin{subfigure}[t]{0.31\linewidth}
    \centering
    \includegraphics[width=1.05\linewidth]{images/darcy/#1_residual.png}
\end{subfigure}

\vspace{2mm}
}

\darcyrow{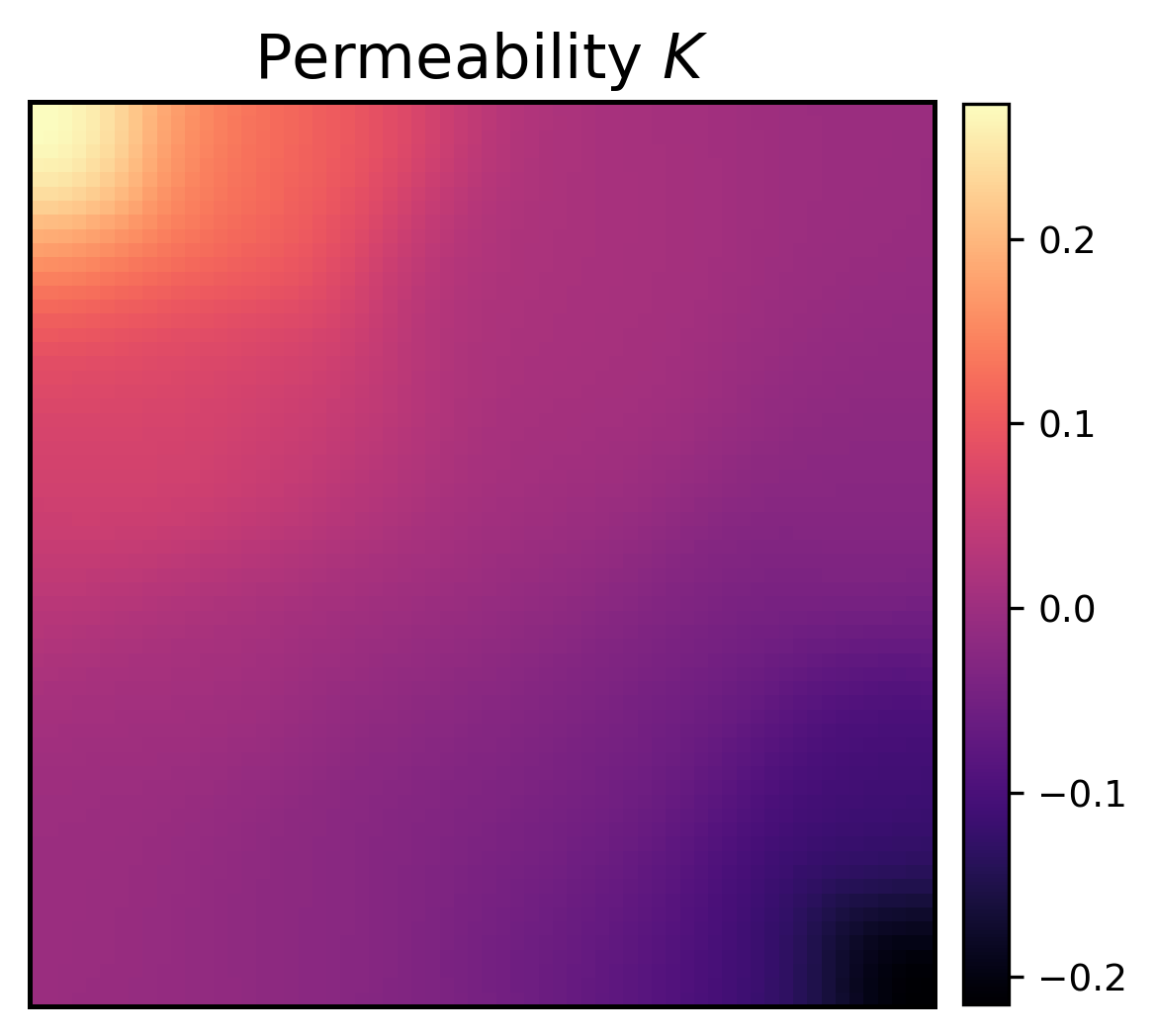}{Sample 0}
\darcyrow{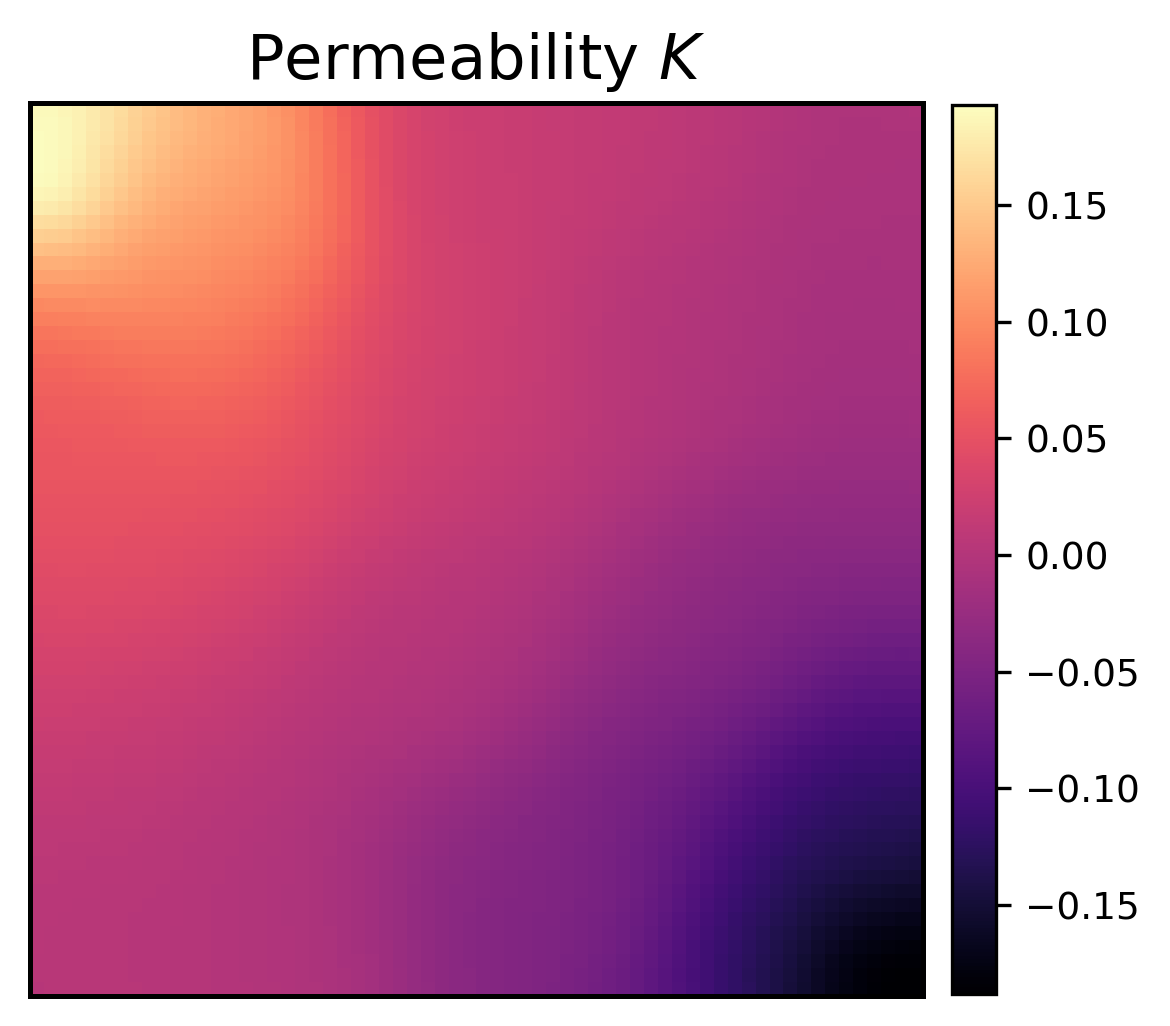}{Sample 3}
\darcyrow{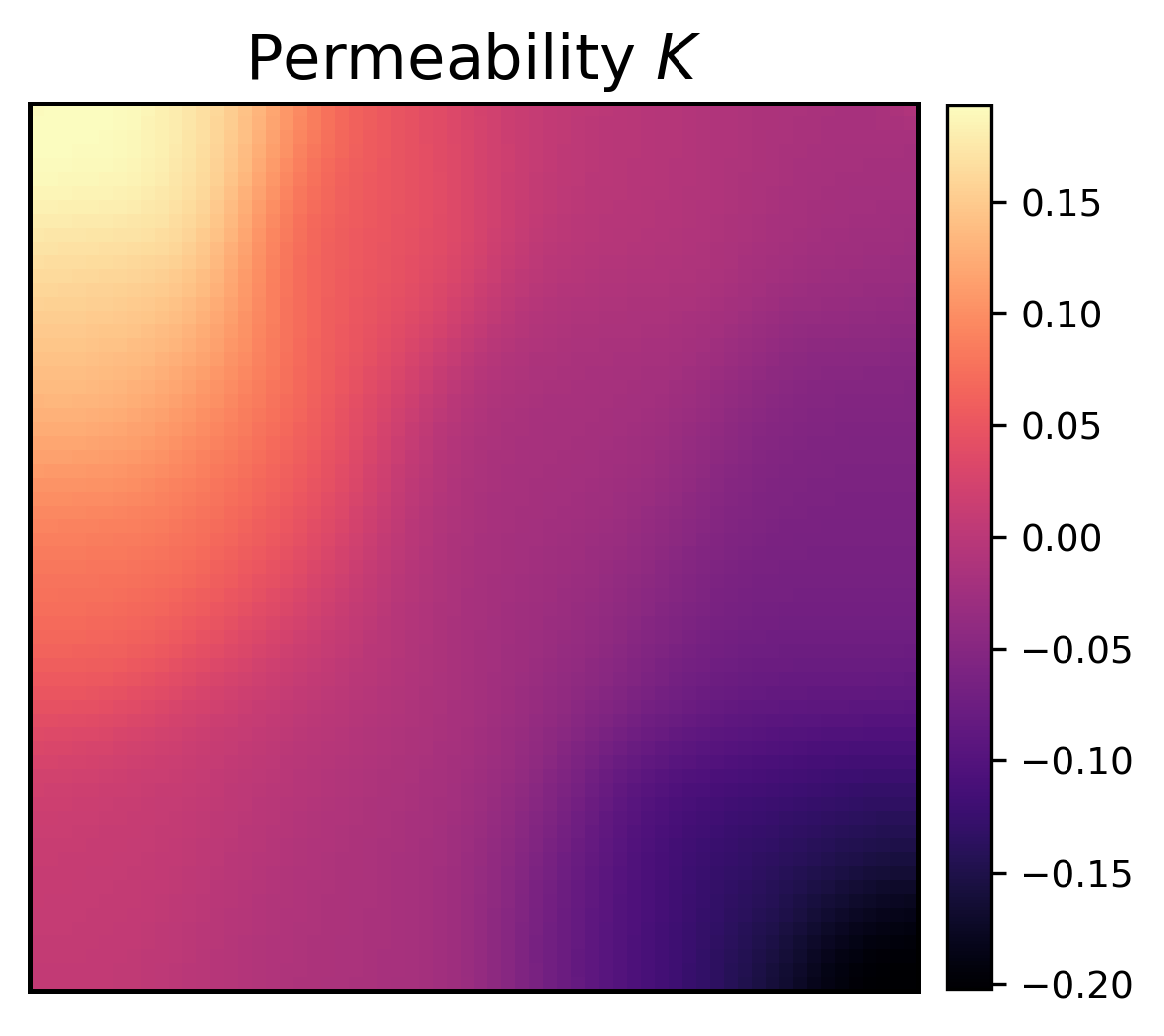}{Sample 6}
\darcyrow{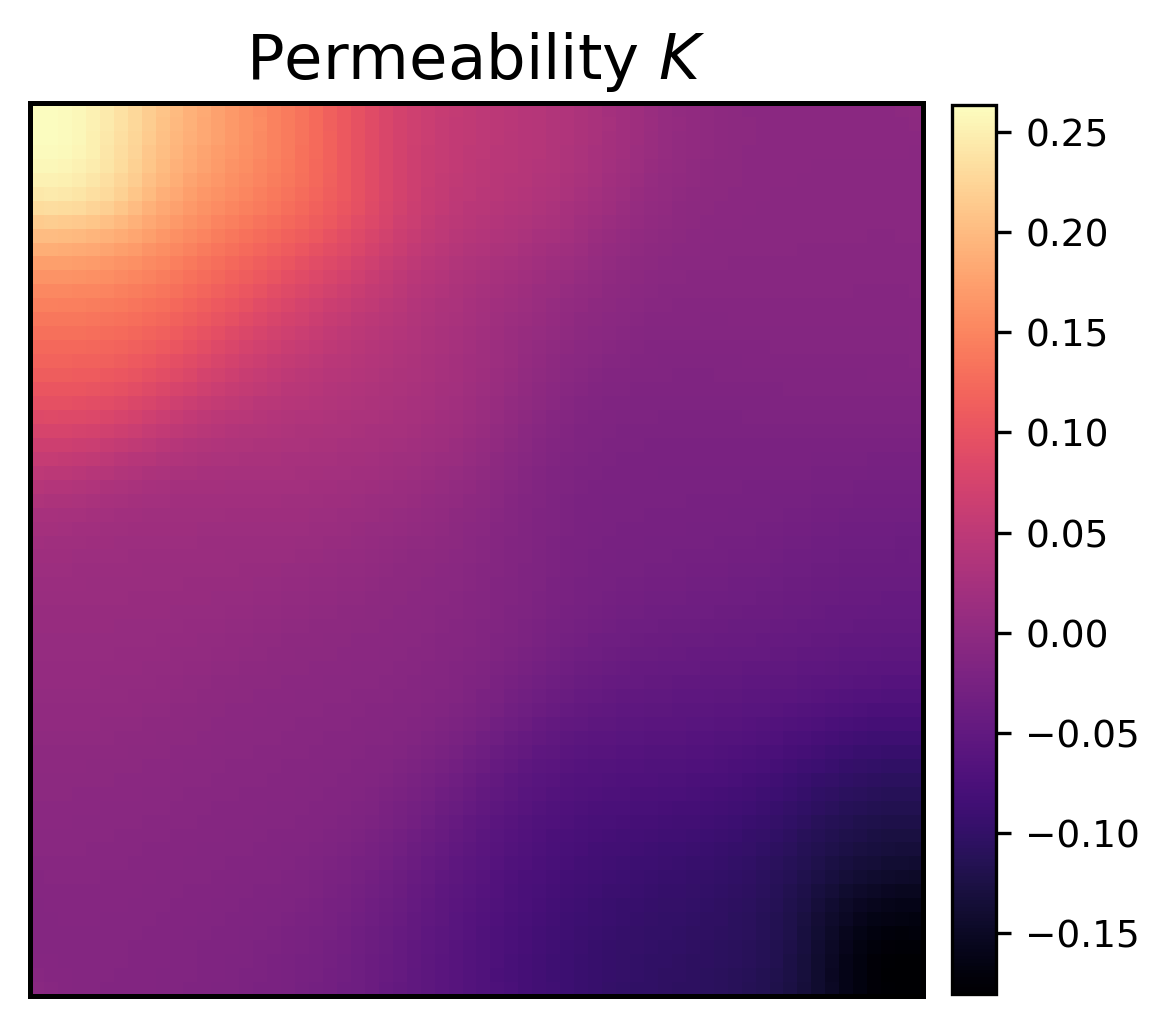}{Sample 7}

\caption{Generated Darcy flow samples. Each row shows one sample, with columns corresponding to pressure $p$, permeability $K$, and residual.}
\label{fig:darcy_generated_5x3}
\end{figure}

\paragraph{Visualization Results.} Qualitative results are shown in \autoref{fig:darcy_generated_5x3}. Our method produces physically consistent samples with coherent pressure and permeability fields, while maintaining low residual errors.

\subsection{Kolmogorov Flow}\label{appedix:kolmogorov_flow}

\paragraph{Dataset and constraints.} Kolmogorov flow is a standard benchmark for two-dimensional incompressible turbulence, where the dynamics are driven by a spatially periodic forcing. We follow the dataset construction in \cite{baldan2026physics}, where velocity fields are defined on a $128\times128$ periodic domain and conditioned on the Reynolds number, which determines the flow regime. The data are generated using a spectral solver based on the vorticity--streamfunction formulation of the incompressible Navier--Stokes equations. Specifically, the vorticity field is evolved in time, and the velocity field $U=(u,v)$ is recovered from the stream function, while the pressure is obtained by solving a Poisson equation. The simulations are performed for Reynolds numbers in the range $[100,500]$, with 32 distinct values used for training and 16 unseen values reserved for evaluation. For each Reynolds number, multiple trajectories are simulated from perturbed initial conditions, and snapshots are collected after the flow reaches a statistically steady regime, resulting in $1024$ samples per condition.

\begin{table}[t]
\centering
\caption{Architecture and training hyperparameters of Darcy Flow.}
\label{tab:hyperparameters}
\small
\resizebox{0.9\linewidth}{!}{%
\begin{tabular}{ll}
\toprule
\textbf{Hyperparameter} & \textbf{Value} \\
\midrule
\multicolumn{2}{c}{\textbf{Architecture hyperparameters}} \\
\midrule
In-, output channels & 2, 2 \\
ResNet blocks per down-/upsampling pass & 2 \\
ResNet block normalization & Group Normalization, 8 groups \\
ResNet block activation function & SiLU \\
Attention block normalization & Layer Normalization \\
Feature map resolutions & $64\!\times\!64 \to 32\!\times\!32 \to 16\!\times\!16 \to 8\!\times\!8$ \\
Latent dimensions & $32 \to 64 \to 128 \to 256$ \\
Attention head dimension & 32 \\
Number of attention heads & 8 \\
Parameters (M) & $9.85$ \\
Conditioning size & None \\
\midrule
\multicolumn{2}{c}{\textbf{Training hyperparameters}} \\
\midrule
Batch size (Darcy flow) & 64 \\
Iterations (Darcy flow) & 400k \\
Learning rate & $5e^{-5}$ \\
Optimization algorithm & Adam \\
EMA decay & 0.9999 \\
Dropout & none \\
Weight Decay & 0 \\
\bottomrule
\end{tabular}%
}
\end{table}

The governing physical constraint is incompressibility, corresponding to mass conservation:
\begin{equation}
\begin{aligned}
\nabla\cdot U(x) = \partial_x u(x) + \partial_y v(x) = 0, \qquad x\in\Omega.
\end{aligned}
\end{equation}
Because the domain is periodic, the velocity field satisfies periodic boundary conditions in both spatial directions. We therefore use circular padding in the neural-network convolutions, so the boundary conditions are respected without additional enforcement. For consistency with the data-generation process, the residual is computed using the same discretization scheme as the spectral solver, avoiding numerical discrepancies between training and evaluation.

This task is conditional: the Reynolds number specifies the operating regime, and the model is required to generate velocity fields that match both the conditional distribution and the incompressibility constraint.

\paragraph{Metrics.}
The physical residual error is defined by the divergence violation:
\begin{equation}
\begin{aligned}
\mathrm{RE}
&=
\mathbb{E}_{x\in\Omega}\![(\nabla\cdot U(x))^2].
\end{aligned}
\end{equation}
We use WD and JS to measure distributional fidelity as in Darcy flow (\autoref{appedix:darcy_flow}). In addition, we report the MSE of the conditional mean fields (MMSE) and the MSE of the conditional standard-deviation fields (SMSE):
\begin{equation}
\begin{aligned}
\mathrm{MMSE}
&=
\mathbb{E}_c\|\mathbb{E}[U| c]-\mathbb{E}[U_{\mathrm{data}}| c]\|_2^2,\\
\mathrm{SMSE}
&=
\mathbb{E}_c\|\mathrm{Std}(U| c)-\mathrm{Std}(U_{\mathrm{data}}| c)\|_2^2.
\end{aligned}
\end{equation}
MMSE evaluates whether the generated samples match the conditional mean flow, while SMSE evaluates whether they reproduce the conditional fluctuation level. We also report the number of function evaluations (NFE) and inference time (IT) to measure sampling efficiency.

\begin{figure}[H]
\centering

\newcommand{\kflowrow}[2]{
\begin{subfigure}[t]{0.31\linewidth}
    \centering
    \includegraphics[width=\linewidth]{images/kflow/#1_u_velocity.png}
\end{subfigure}
\hfill
\begin{subfigure}[t]{0.31\linewidth}
    \centering
    \includegraphics[width=\linewidth]{images/kflow/#1_v_velocity.png}
\end{subfigure}
\hfill
\begin{subfigure}[t]{0.31\linewidth}
    \centering
    \includegraphics[width=\linewidth]{images/kflow/#1_residual.png}
\end{subfigure}

\vspace{2mm}
}

\kflowrow{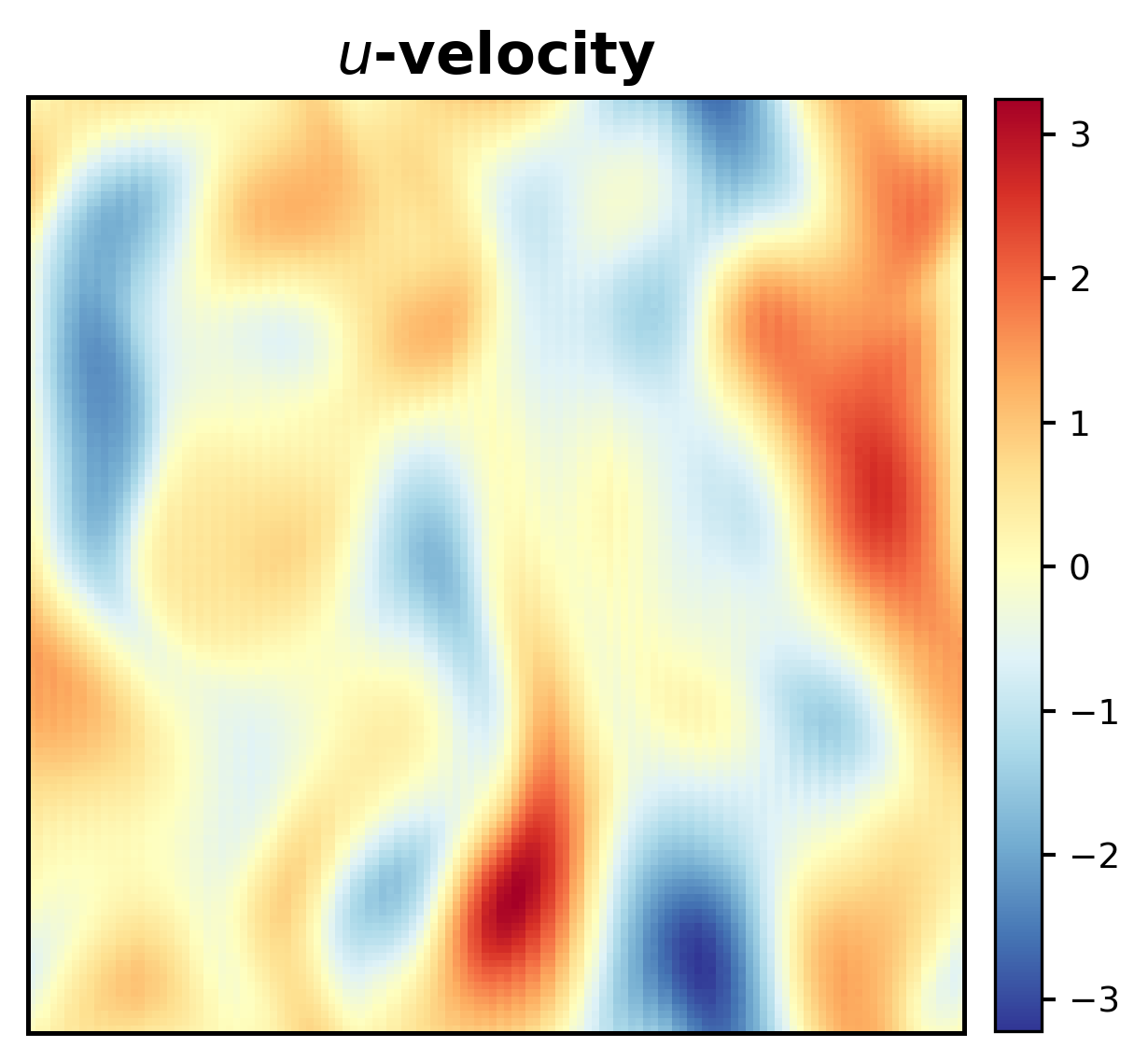}{Sample 0}
\kflowrow{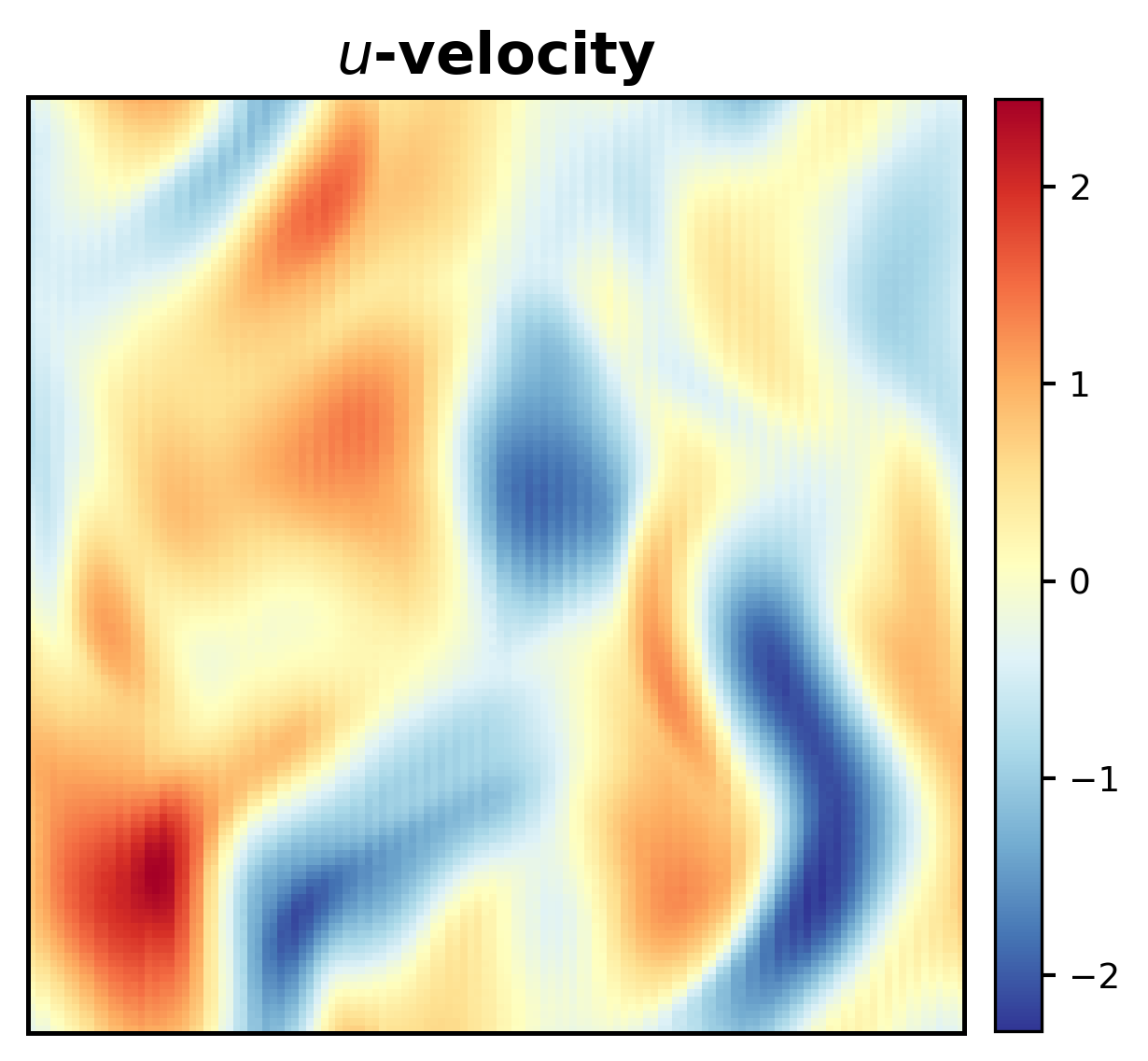}{Sample 3}
\kflowrow{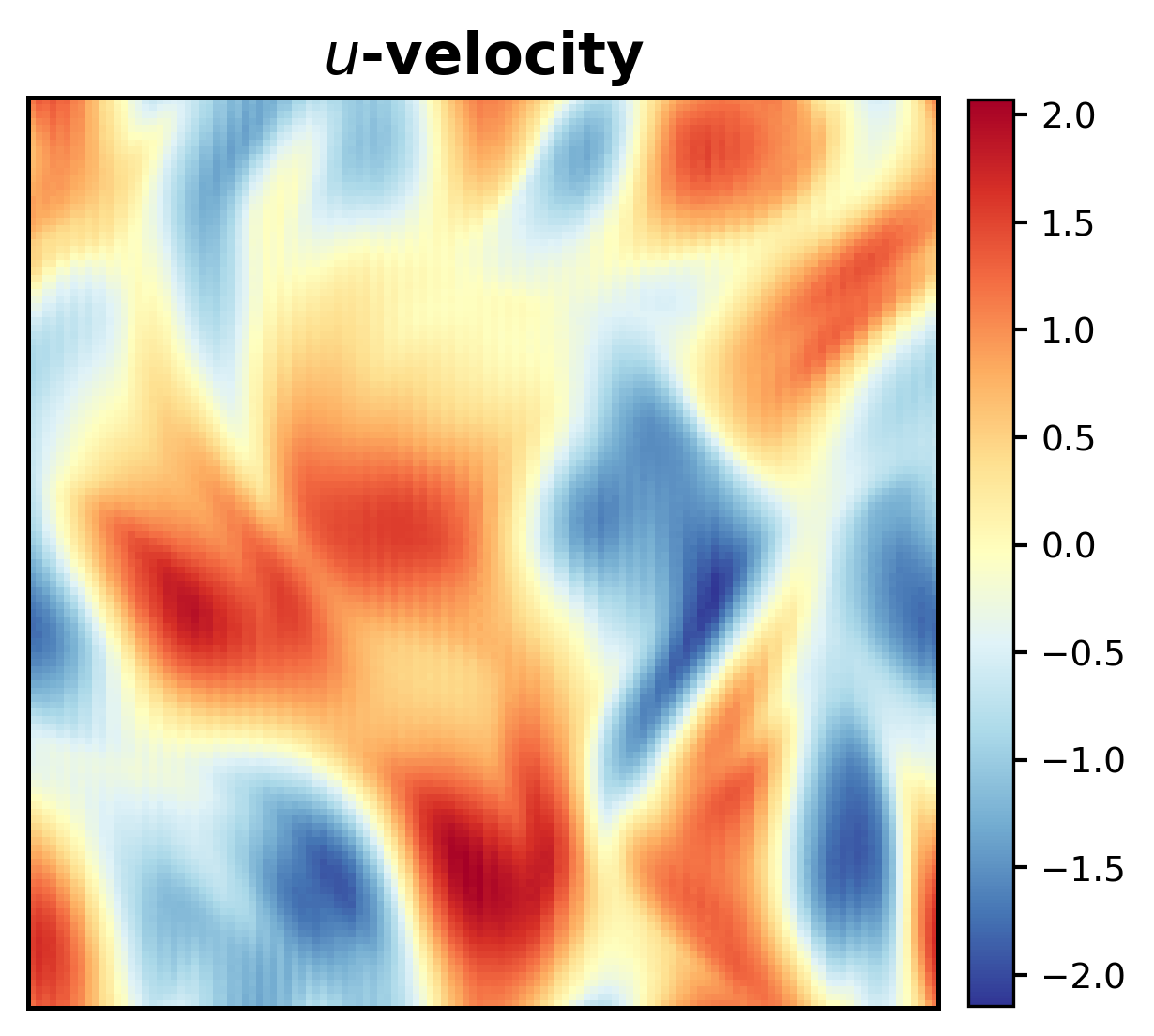}{Sample 6}
\kflowrow{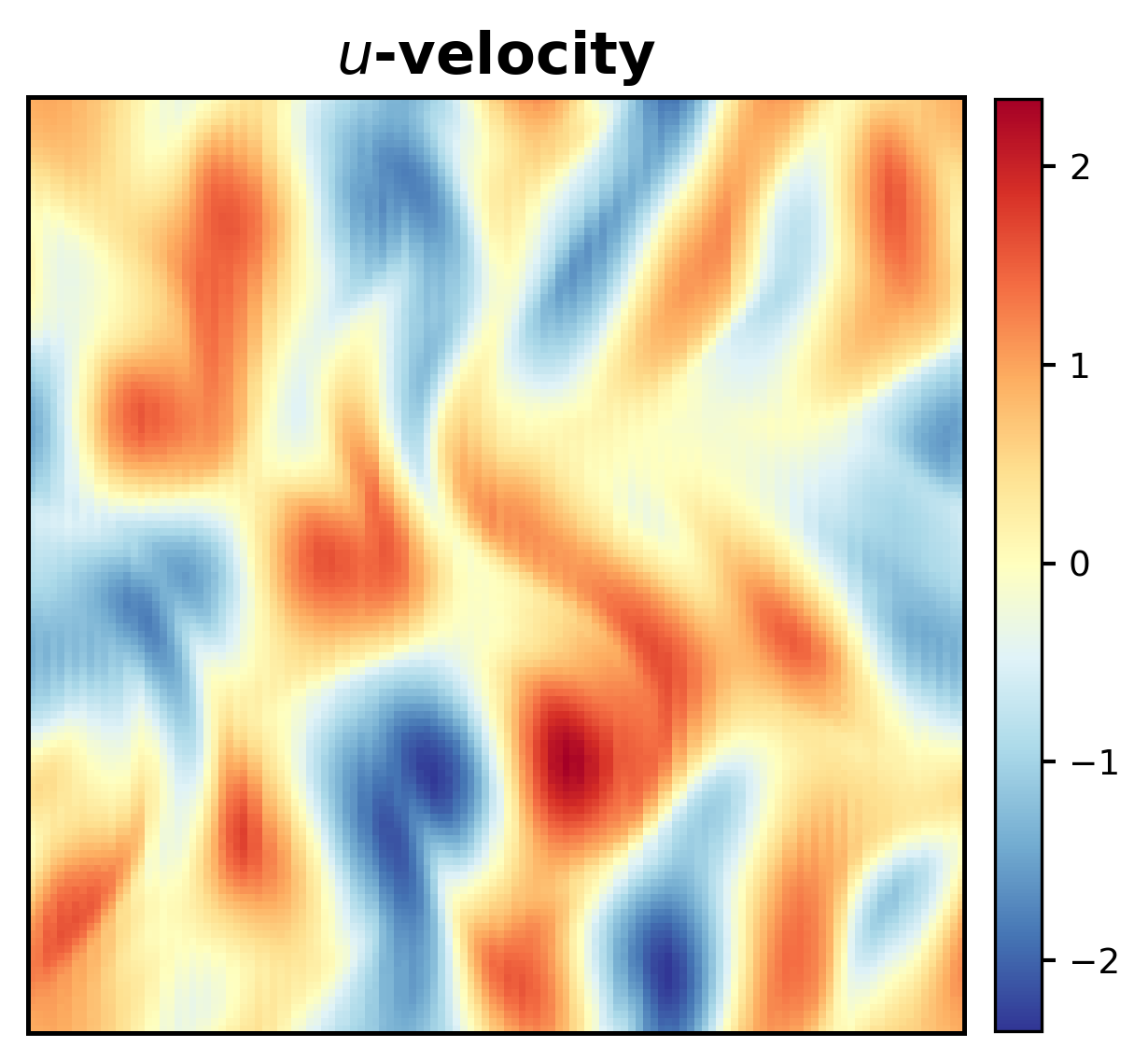}{Sample 7}

\caption{Generated Kolmogorov flow samples. Each row shows one sample, with columns corresponding to $u$-velocity, $v$-velocity, and divergence residual. From top to bottom, the Reynolds number increases.}
\label{fig:kflow_generated_4x3}
\end{figure}

\paragraph{Model and training.}
We use the same architecture as in \cite{bastek2024physics} without additional architectural modifications. For experiments with feature-space stabilization, the encoder is used only during training to compute the data target, while generation is still performed in the original data space. The training details of the feature encoder are provided in \autoref{appendix:encoder_study}, and the architecture and training configuration of the main model after encoder pretraining are summarized in \autoref{tab:hyperparameters_kflow}.

\paragraph{Visualization Results.} Qualitative results are shown in \autoref{fig:kflow_generated_4x3}. Our method produces coherent velocity fields while maintaining low divergence residuals.

\subsection{Dynamic Stall}\label{appedix:dynamic_stalling}

\paragraph{Dataset and constraints.}
Dynamic stall is a conditional generation benchmark for unsteady compressible flow around a sinusoidally pitching NACA0012 airfoil. We follow the dataset construction of \cite{baldan2026physics}, where the data are generated by solving the two-dimensional unsteady compressible Reynolds-averaged Navier-Stokes (RANS) equations on an O-type mesh around the airfoil. The simulations are performed with ANSYS Fluent \cite{ansys2024fluent} using an implicit finite-volume solver, and the resulting flow fields are downsampled to a $128\times128$ grid. The airfoil motion is prescribed by a sinusoidal pitching law, and each simulation is run until periodic convergence is reached, after which flow snapshots are collected over a full pitching cycle (see \cite{baldan2026physics} for further details).  Each sample contains six predicted fields, including absolute pressure $P$, temperature $T$, density $\rho$, skin friction $\tau_w$, and tangential velocity gradients along the airfoil surface. The model is conditioned on four operating parameters: the free-stream Mach number, the mean angle of attack, the pitching amplitude, and the reduced frequency. Following \cite{baldan2026physics}, training and testing configurations are sampled from this four-dimensional design space, with additional perturbations applied to each nominal configuration. {\color{black}The training set contains 128 base configurations, each perturbed 32 times, and the test set contains 16 unseen configurations.}

\begin{table}[t]
\centering
\caption{Architecture and training hyperparameters of Kolmogorov flow.}
\label{tab:hyperparameters_kflow}
\small
\resizebox{0.9\linewidth}{!}{%
\begin{tabular}{ll}
\toprule
\textbf{Hyperparameter} & \textbf{Value} \\
\midrule
\multicolumn{2}{c}{\textbf{Architecture hyperparameters}} \\
\midrule
In-, output channels & 2, 2 \\
Global conditioning dimension & 1 \\
ResNet blocks per down-/upsampling pass & 2 \\
ResNet block normalization & Group Normalization, 8 groups \\
ResNet block activation function & SiLU \\
Attention block normalization & Layer Normalization \\
Feature map resolutions & $128\!\times\!128 \to 64\!\times\!64 \to 32\!\times\!32 \to 16\!\times\!16 \to 8\!\times\!8$ \\
Latent dimensions & $32 \to 64 \to 128 \to 256 \to 512$ \\
Attention head dimension & 32 \\
Number of attention heads & 8 \\
Parameters (M) & 36.2 \\
Conditioning variable & 1 \\
Sigmoid output & False \\
\midrule
\multicolumn{2}{c}{\textbf{Training hyperparameters}} \\
\midrule
Batch size (K-flow) & 64 \\
Iterations (K-flow) & 400k \\
Learning rate & $5\times 10^{-5}$ \\
Optimization algorithm & Adam \\
EMA decay & 0.9999 \\
Dropout & none \\
Weight Decay & 0 \\
\bottomrule
\end{tabular}%
}
\end{table}

The physical constraints are imposed through two analytical point-wise residuals following \cite{baldan2026physics}. The first residual enforces the ideal gas relation:
\begin{equation}
\begin{aligned}
\mathcal{R}_{\mathrm{ig}}
&=
P-\rho R T,
\end{aligned}
\end{equation}
while the second constrains the skin friction using the temperature-dependent viscosity from Sutherland's law:
\begin{equation}
\begin{aligned}
\mathcal{R}_{\tau}
&=
\tau_w
-
\mu(T)
\sqrt{(\partial_x u)^2+(\partial_y u)^2},\\
\mu(T)
&=
\mu_0
\frac{T_0+S}{T+S}
|(\frac{T}{T_0}|)^{3/2}.
\end{aligned}
\end{equation}
Here, $R$ is the gas constant, and $\mu_0$, $T_0$, and $S$ are the reference viscosity, reference temperature, and Sutherland constant. These constraints couple the generated thermodynamic variables and near-wall quantities.

\paragraph{Metrics.}
The physical residual error combines the two point-wise residuals:
\begin{equation}
\begin{aligned}
\mathrm{RE}
&=
\mathbb{E}\![
\mathcal{R}_{\mathrm{ig}}^2
+
\mathcal{R}_{\tau}^2
].
\end{aligned}
\end{equation}
We use the same distributional and statistical metrics as in Kolmogorov flow, including WD, JS, MMSE, and SMSE. These metrics evaluate whether the generated samples match both the physical constraints and the conditional statistics of the target flow fields. NFE and IT are also reported to compare sampling efficiency.

\paragraph{Model and training.}
We use the same basic architecture as in \cite{bastek2024physics} without additional architectural modifications. The model is conditioned on the operating and pitching parameters of the dynamic-stall case. For experiments with feature-space stabilization, the encoder is used only during training to compute the data target, while generation is still performed in the original data space. The architecture and training configuration are summarized in \autoref{tab:hyperparameters_dynamic_stall}.

\begin{table}[t]
\centering
\caption{Architecture and training hyperparameters of dynamic stall.}
\label{tab:hyperparameters_dynamic_stall}
\small
\resizebox{0.9\linewidth}{!}{%
\begin{tabular}{ll}
\toprule
\textbf{Hyperparameter} & \textbf{Value} \\
\midrule
\multicolumn{2}{c}{\textbf{Architecture hyperparameters}} \\
\midrule
In-, output channels & 6, 6 \\
Global conditioning dimension & 4 \\
ResNet blocks per down-/upsampling pass & 2 \\
ResNet block normalization & Group Normalization, 6 groups \\
ResNet block activation function & SiLU \\
Attention block normalization & Layer Normalization \\
Feature map resolutions & $128\!\times\!128 \to 64\!\times\!64 \to 32\!\times\!32 \to 16\!\times\!16$ \\
Latent dimensions & $12 \to 24 \to 48 \to 96$ \\
Attention head dimension & 32 \\
Number of attention heads & 8 \\
Parameters (M) & 1.9 \\
Conditioning variables & 4 \\
Sigmoid output & False \\
\midrule
\multicolumn{2}{c}{\textbf{Training hyperparameters}} \\
\midrule
Batch size (dynamic stall) & 64 \\
Iterations (dynamic stall) & 400k \\
Learning rate & $5\times 10^{-5}$ \\
Optimization algorithm & Adam \\
EMA decay & 0.9999 \\
Dropout & none \\
Weight Decay & 0 \\
\bottomrule
\end{tabular}%
}
\end{table}

\begin{figure}[H]
\centering

\newcommand{\stallrow}[1]{
\begin{subfigure}[t]{0.16\linewidth}
    \centering
    \includegraphics[width=\linewidth]{images/dynamic_stall/#1_P.png}
\end{subfigure}
\hfill
\begin{subfigure}[t]{0.16\linewidth}
    \centering
    \includegraphics[width=\linewidth]{images/dynamic_stall/#1_T.png}
\end{subfigure}
\hfill
\begin{subfigure}[t]{0.16\linewidth}
    \centering
    \includegraphics[width=\linewidth]{images/dynamic_stall/#1_rho.png}
\end{subfigure}
\hfill
\begin{subfigure}[t]{0.16\linewidth}
    \centering
    \includegraphics[width=\linewidth]{images/dynamic_stall/#1_tau_w.png}
\end{subfigure}
\hfill
\begin{subfigure}[t]{0.16\linewidth}
    \centering
    \includegraphics[width=\linewidth]{images/dynamic_stall/#1_du_dx.png}
\end{subfigure}
\hfill
\begin{subfigure}[t]{0.16\linewidth}
    \centering
    \includegraphics[width=\linewidth]{images/dynamic_stall/#1_du_dy.png}
\end{subfigure}

\vspace{2mm}
}

\stallrow{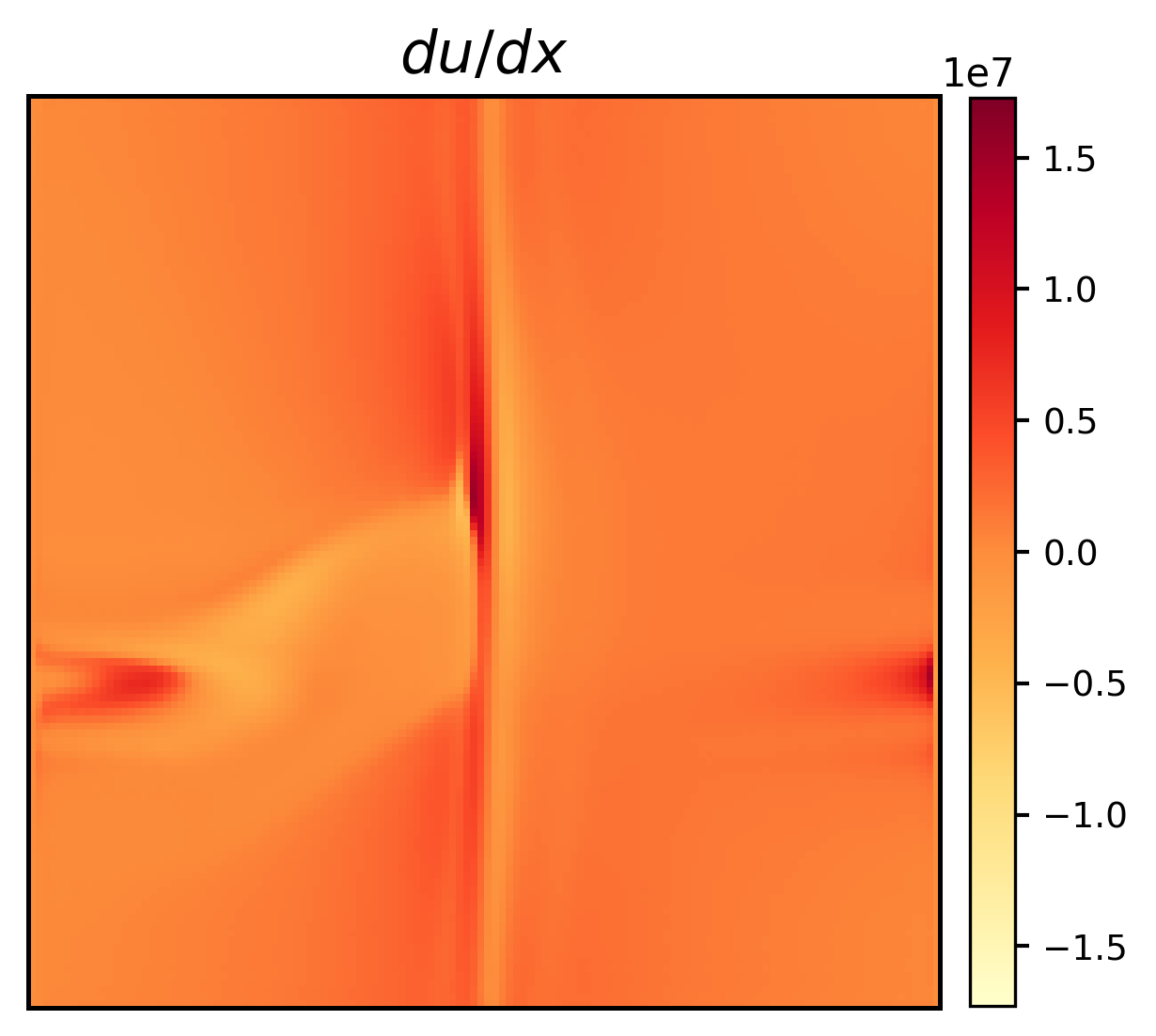}
\stallrow{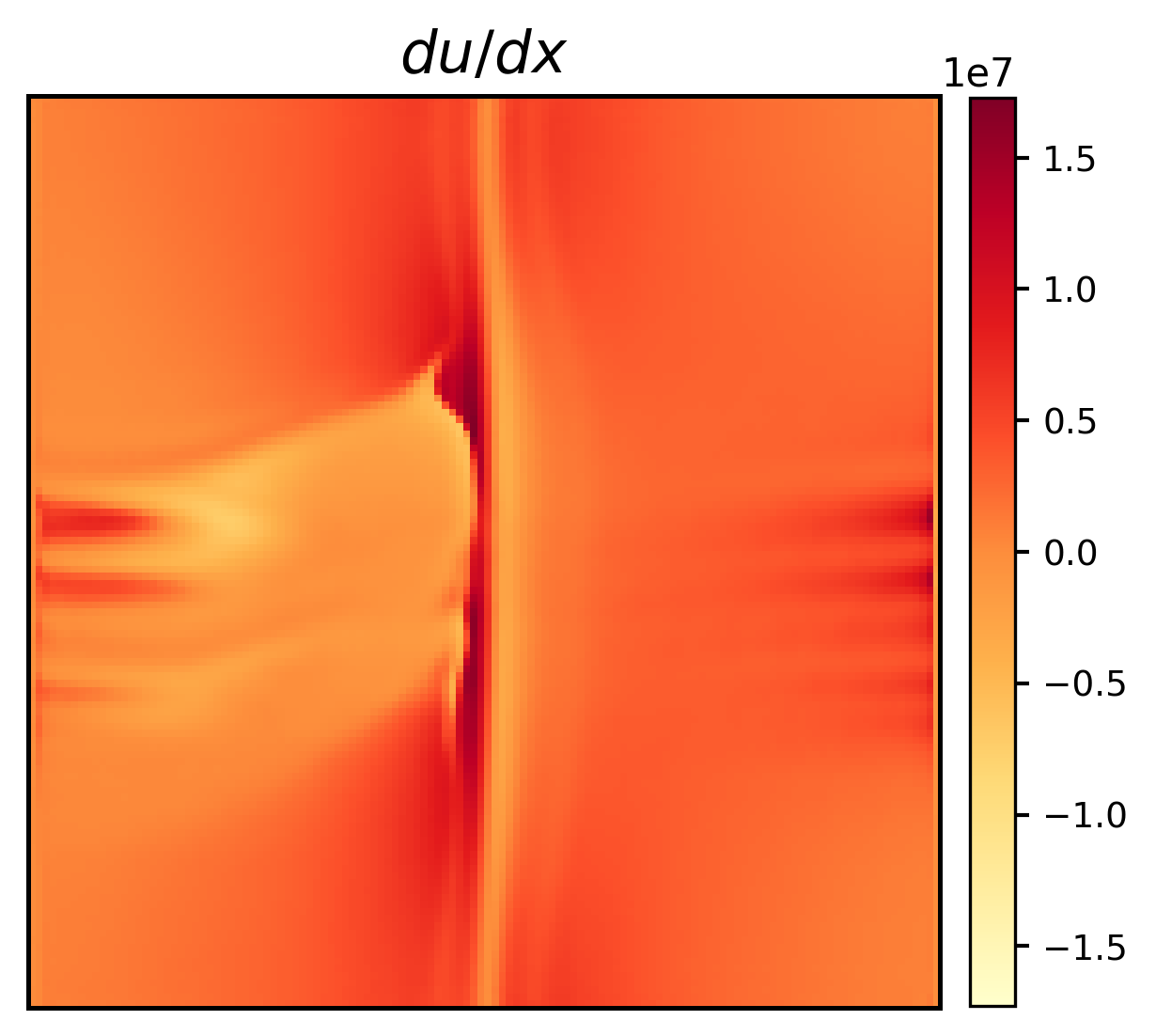}
\stallrow{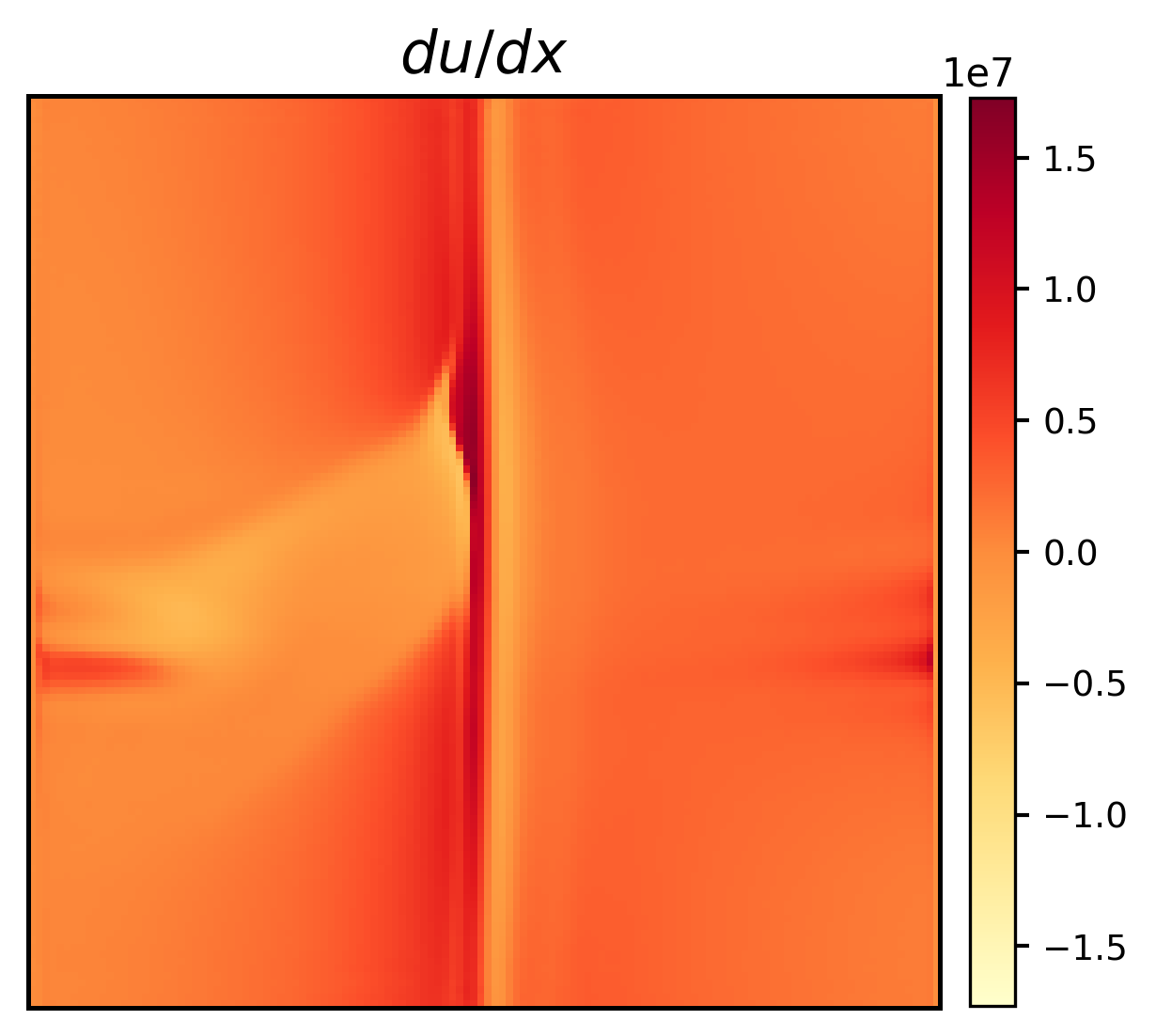}
\stallrow{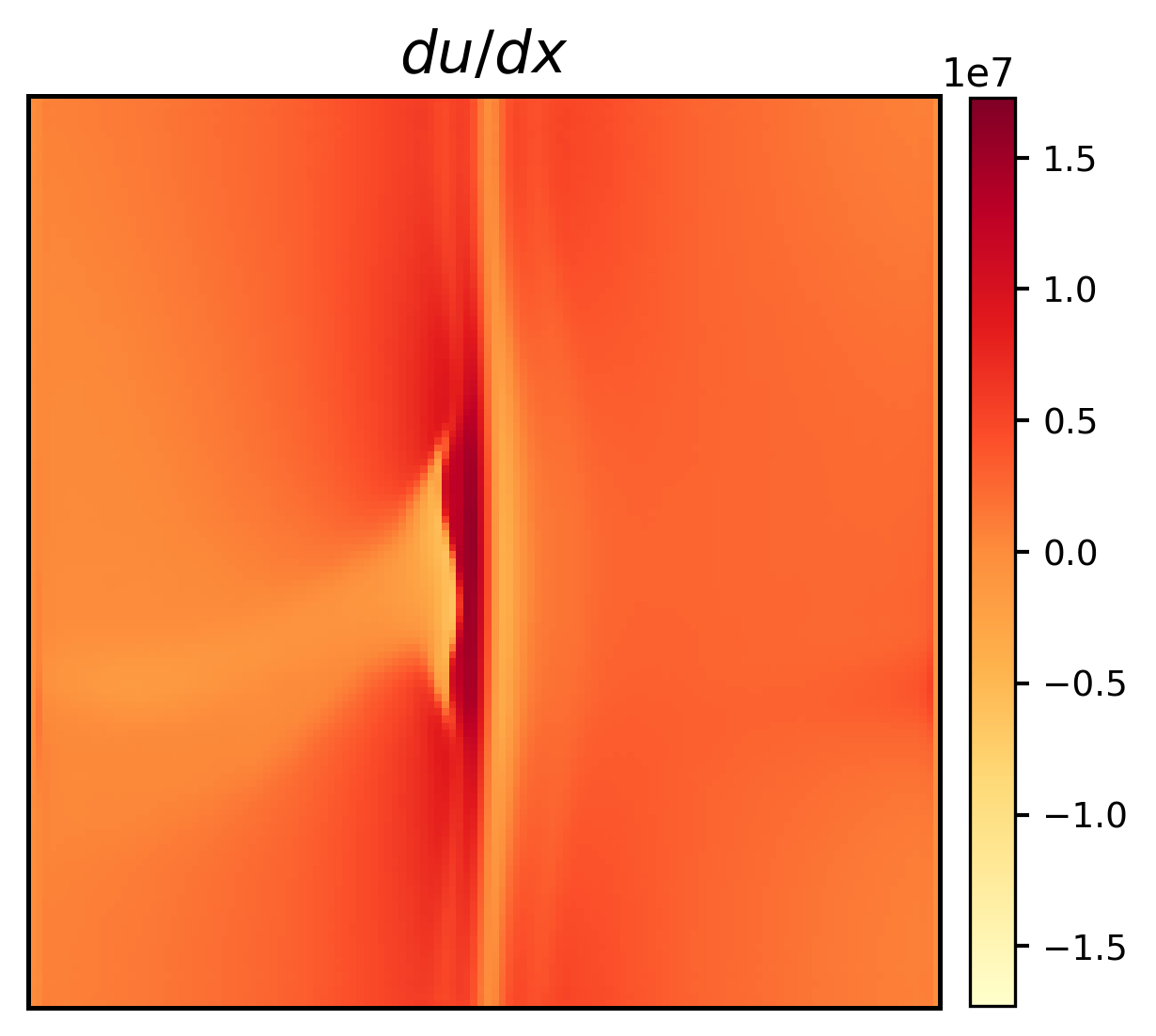}

\caption{Generated dynamic-stall samples. Each row corresponds to one generated sample, and the columns show pressure $P$, temperature $T$, density $\rho$, wall shear stress $\tau_w$, and tangential velocity gradients $\partial_x u$ and $\partial_y u$, respectively.}
\label{fig:dynamic_stall_generated_4x6}
\end{figure}

\paragraph{Visualization Results.} Qualitative results are shown in \autoref{fig:dynamic_stall_generated_4x6}. Our method generates coherent dynamic-stall fields across all predicted quantities.

\subsection{Topology Optimization}\label{appedix:topology}

\paragraph{Dataset and constraints.}
Topology optimization is a conditional inverse-design benchmark, where the objective is to generate a material distribution that achieves high structural stiffness under prescribed loading and boundary conditions. We follow the setup of \cite{bastek2024physics}. Each sample is defined on a $65\times65$ grid representing a discretized design domain. The model is conditioned on the external load field $\mathbf{f}=(f_x,f_y)$, boundary conditions, auxiliary physical fields (e.g., strain energy and von Mises stress), and a target volume constraint $V_{\max}$. These conditioning fields are concatenated with noise and provided as input to the network, which predicts both the displacement field $u=(u_1,u_2)$ and the material density field $\rho$. 

The problem corresponds to minimum-compliance topology optimization under linear elasticity \cite{bastek2024physics}. The objective is to minimize the total elastic energy
\begin{equation}
\begin{aligned}
\min_{\rho}
&\;
\int_{\Omega}
\frac{1}{2}\,
\sigma(\xi):\varepsilon(\xi)\,d\Omega,
\end{aligned}
\end{equation}
subject to the box constraint $\rho(\xi)\in[0,1]$, the volume constraint $\int_{\Omega}\rho(\xi)\,d\Omega \le V_{\max}$, and the equilibrium condition
$\nabla\cdot\sigma(\xi)+\mathbf{f}(\xi)=0$.
Dirichlet boundary conditions are imposed on fixed supports, while Neumann boundary conditions specify the applied loads.  The stress and strain fields are related through a density-dependent constitutive model
$\sigma=\mathbb{C}(\rho):\varepsilon(u)$, where $\varepsilon =\frac{1}{2}(\nabla u+\nabla u^\top)$.
Following \cite{bastek2024physics}, the material stiffness tensor is parameterized using the SIMP formulation, $\mathbb{C}(\rho)=\rho^p \mathbb{C}_0$ with $p>1$, which encourages near-binary material distributions.  The displacement field $u$ is discretized on the underlying grid, leading to a finite-dimensional representation $U$. Under a standard finite-element formulation, the equilibrium equation is converted into a linear system
\begin{equation}
\begin{aligned}
K(\rho) U = F,
\end{aligned}
\end{equation}
where $K(\rho)$ is the global stiffness matrix assembled from element-wise contributions, $U$ denotes the nodal displacement vector of $u$, and $F$ is the external force vector of $f$. Consistent with \cite{bastek2024physics}, we evaluate the physical residual by assembling the FEM stiffness system directly on the grid representation, rather than relying on finite-difference approximations, to ensure consistency with the data generation process.

The dataset is defined on a $65\times65$ nodal grid, which we interpolate to a $64\times64$ grid for compatibility with the convolutional architecture, following \cite{bastek2024physics}. We adopt the same data split as \cite{bastek2024physics}, where the test set is divided into in-distribution and out-of-distribution cases. The in-distribution setting contains samples with loading and boundary configurations similar to those seen during training, while the out-of-distribution setting introduces unseen configurations to evaluate generalization under novel physical conditions.

\paragraph{Metrics.} We evaluate the generated solutions from three complementary perspectives: physical consistency, structural optimality, and volume-constraint satisfaction.  For physical consistency, we measure the violation of the discretized equilibrium equation using the residual mean absolute error. Specifically, for each sample we compute the $\ell_1$ norm of the FEM residual and report its median over the test set:
\begin{equation}
\begin{aligned}
R_{\mathrm{MAE}}
&=
\mathrm{median}
|(
\frac{1}{N}
|\|
K(\rho)\,U - F
|\|_1
|),
\end{aligned}
\end{equation}
where $K(\rho)$ denotes the global stiffness matrix assembled from the predicted density field, $U$ is the corresponding nodal displacement obtained from the model, $F$ is the external force vector, and $N$ is the number of grid points. This metric quantifies how well the generated fields satisfy mechanical equilibrium.  To assess structural performance, we report the median compliance error (CE). Following standard practice \cite{bastek2024physics}, the predicted density field is first binarized, and the corresponding displacement is recomputed by solving the FEM system. The compliance of the generated structure is then defined as
\begin{equation}
\begin{aligned}
C_{\mathrm{gen}}
=
U_{\mathrm{gen}}^\top F,
\end{aligned}
\end{equation}
and compared against the reference compliance $C_{\mathrm{ref}}$, yielding
\begin{equation}
\begin{aligned}
\mathrm{CE}
&=
\frac{C_{\mathrm{gen}} - C_{\mathrm{ref}}}{C_{\mathrm{ref}}}.
\end{aligned}
\end{equation}

Finally, we evaluate satisfaction of the volume constraint using the volume fraction error (VFE). After binarization, the average density is compared with the target volume fraction $v_f$:
\begin{equation}
\begin{aligned}
\mathrm{VFE}
&=
\frac{
||
\bar{\rho}_{\mathrm{bin}} - v_f
||
}{v_f},
\end{aligned}
\end{equation}
where $\bar{\rho}_{\mathrm{bin}}$ denotes the mean value of the binarized density field. Lower values of $R_{\mathrm{MAE}}$, CE, and VFE indicate better physical consistency, structural optimality, and constraint satisfaction, respectively.

\paragraph{Model and training.}
We use the same basic architecture as in \cite{bastek2024physics} without additional architectural modifications. The model is conditioned on the loading, boundary-condition, and volume-constraint information of the topology-optimization problem. For experiments with feature-space stabilization, the encoder is used only during training to compute the data target, while generation is still performed in the original data space. The architecture and training configuration are summarized in \autoref{tab:hyperparameters_topology}.

\begin{table}[t]
\centering
\caption{Architecture and training hyperparameters of topology optimization.}
\label{tab:hyperparameters_topology}
\small
\resizebox{0.9\linewidth}{!}{%
\begin{tabular}{ll}
\toprule
\textbf{Hyperparameter} & \textbf{Value} \\
\midrule
\multicolumn{2}{c}{\textbf{Architecture hyperparameters}} \\
\midrule
In-, output channels & 10, 3 \\
ResNet blocks per down-/upsampling pass & 2 \\
ResNet block normalization & Group Normalization, 8 groups \\
ResNet block activation function & SiLU \\
Attention block normalization & Layer Normalization \\
Feature map resolutions & $64\!\times\!64 \to 32\!\times\!32 \to 16\!\times\!16 \to 8\!\times\!8$ \\
Latent dimensions & $128 \to 256 \to 512 \to 1024$ \\
Attention head dimension & 32 \\
Number of attention heads & 8 \\
Parameters (M) & 131.2 \\
Sigmoid output & True for the last channel \\
\midrule
\multicolumn{2}{c}{\textbf{Training hyperparameters}} \\
\midrule
Batch size (topology optimization) & 64 \\
Iterations (topology optimization) & 600k \\
Learning rate & $5\times 10^{-5}$ \\
Optimization algorithm & Adam \\
EMA decay & 0.9999 \\
Dropout & none \\
Weight Decay & 0 \\
\bottomrule
\end{tabular}%
}
\end{table}

\begin{figure}[H]
\centering

\newcommand{\toporow}[2]{
\begin{subfigure}[t]{0.24\linewidth}
    \centering
    \includegraphics[width=\linewidth]{images/topology/#1_rho.png}
\end{subfigure}
\hfill
\begin{subfigure}[t]{0.24\linewidth}
    \centering
    \includegraphics[width=\linewidth]{images/topology/#1_residual.png}
\end{subfigure}
\hfill
\begin{subfigure}[t]{0.24\linewidth}
    \centering
    \includegraphics[width=\linewidth]{images/topology/#2_rho.png}
\end{subfigure}
\hfill
\begin{subfigure}[t]{0.24\linewidth}
    \centering
    \includegraphics[width=\linewidth]{images/topology/#2_residual.png}
\end{subfigure}

\vspace{2mm}
}

\toporow{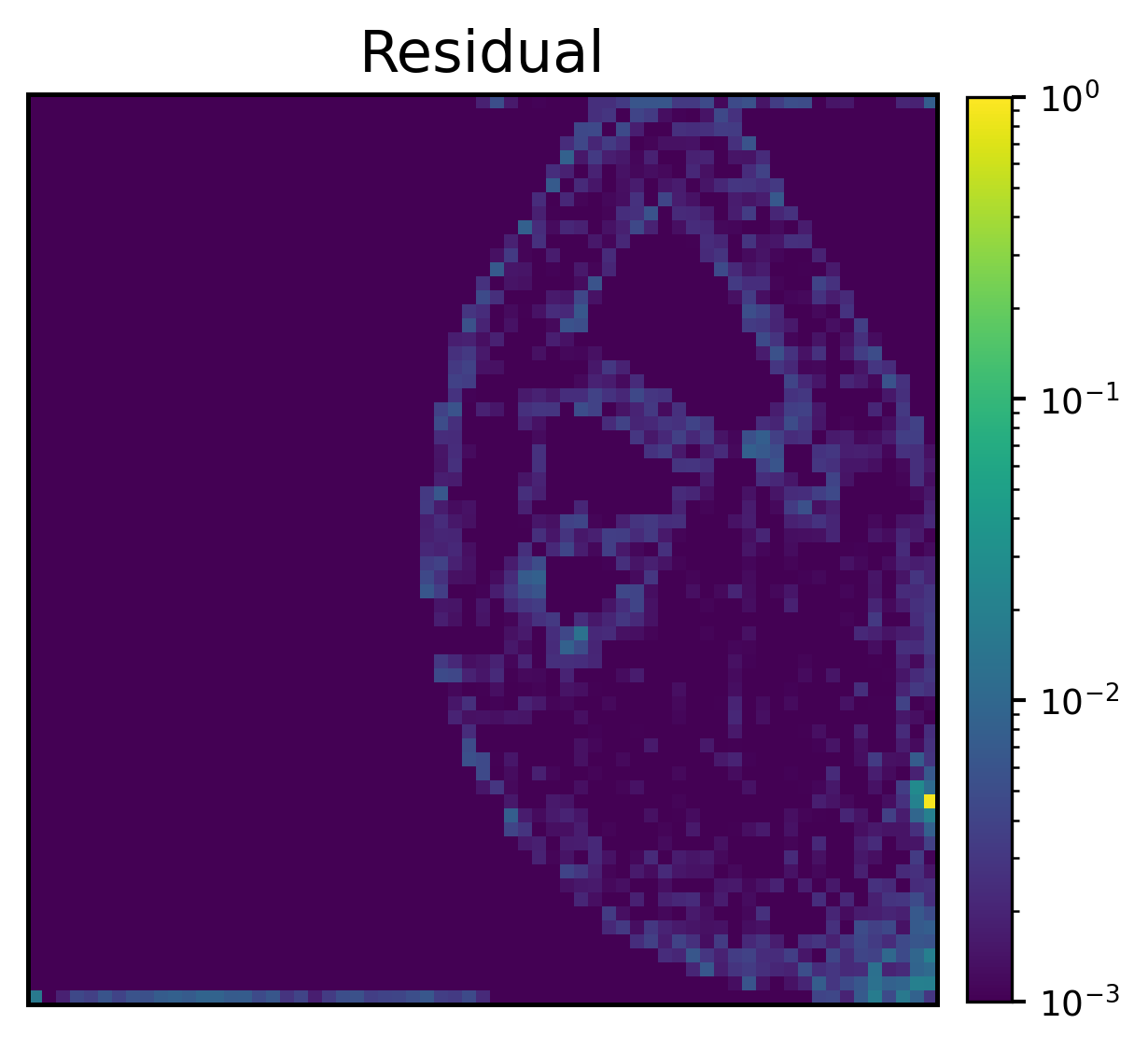}{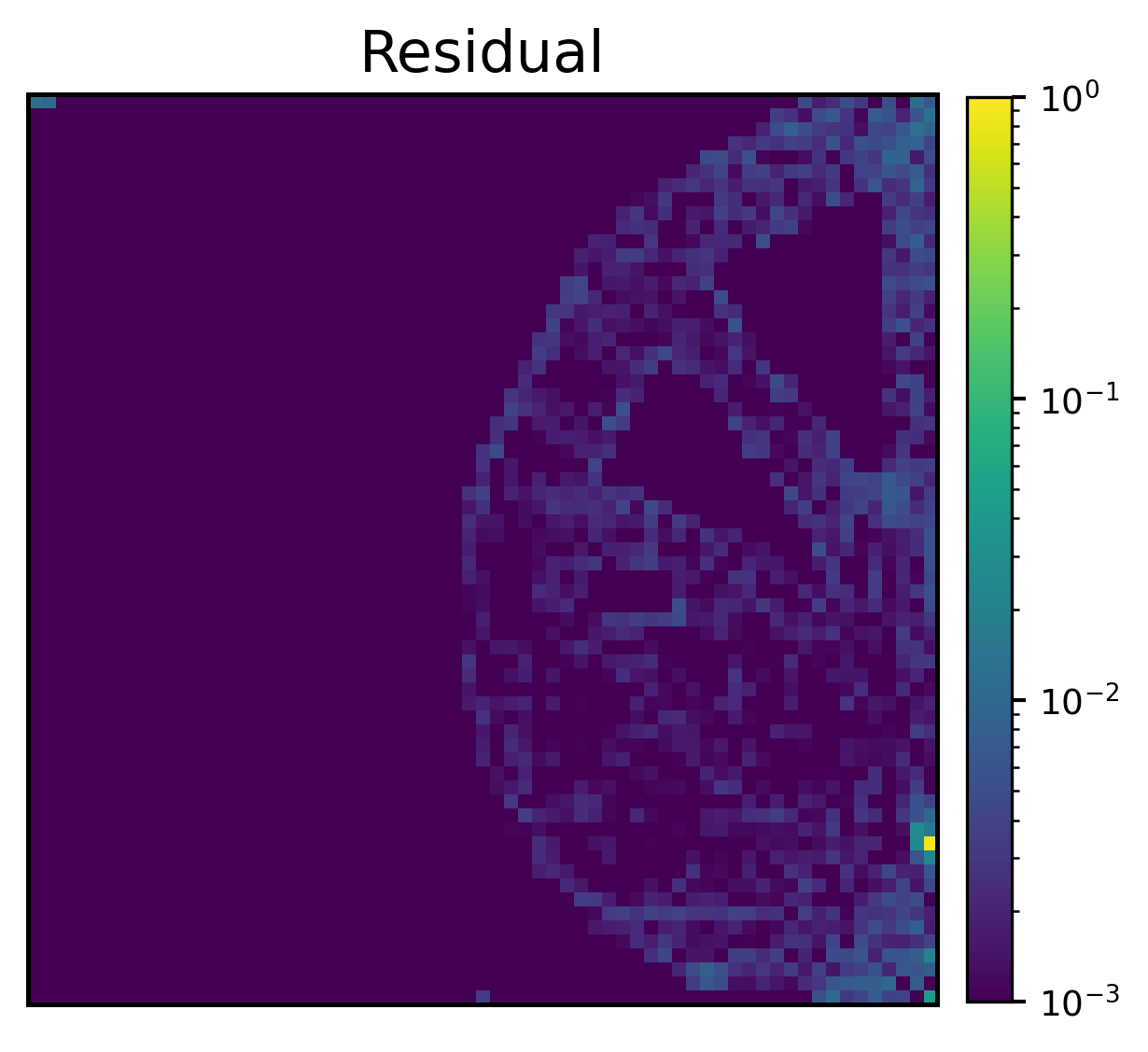}
\toporow{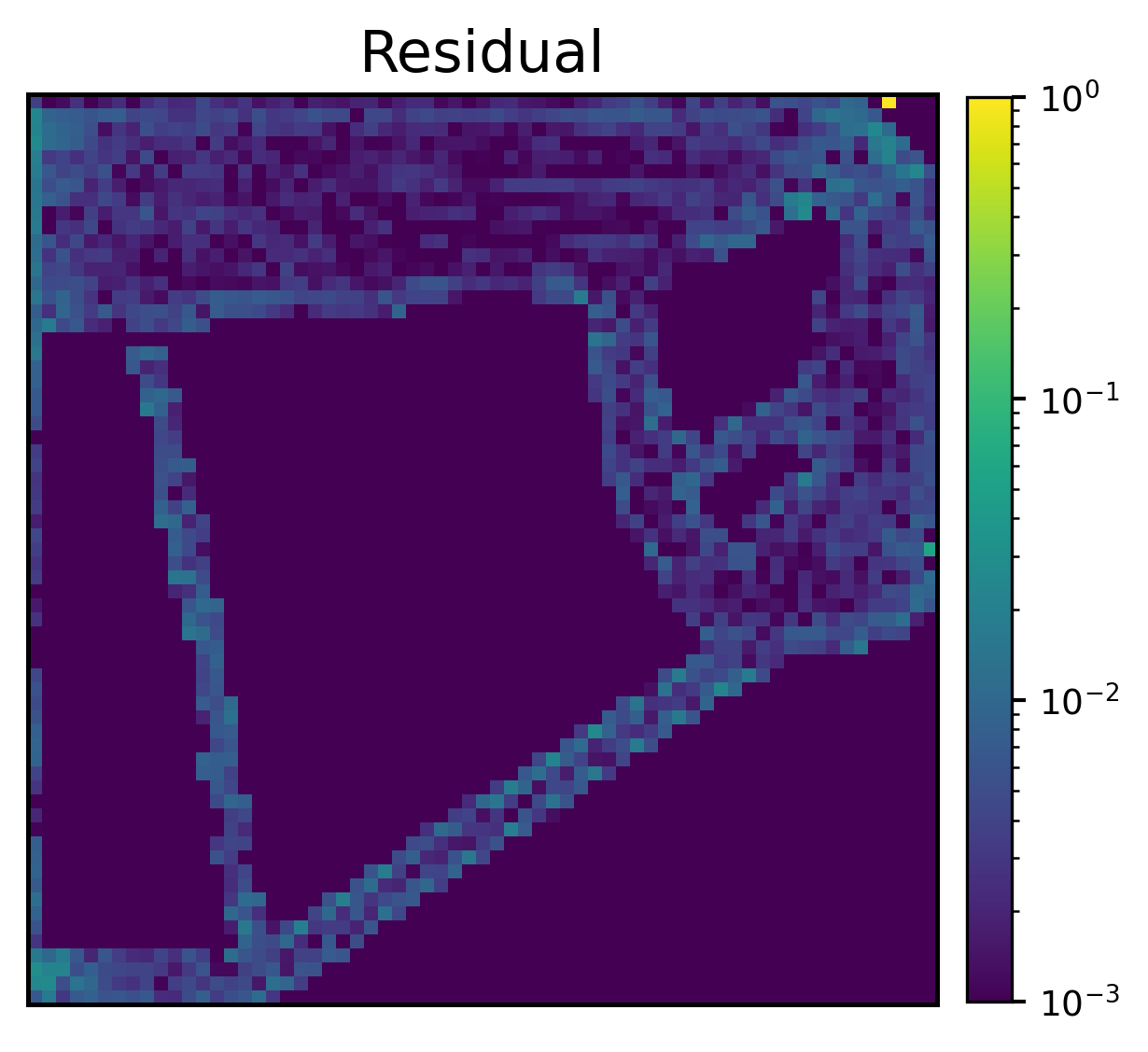}{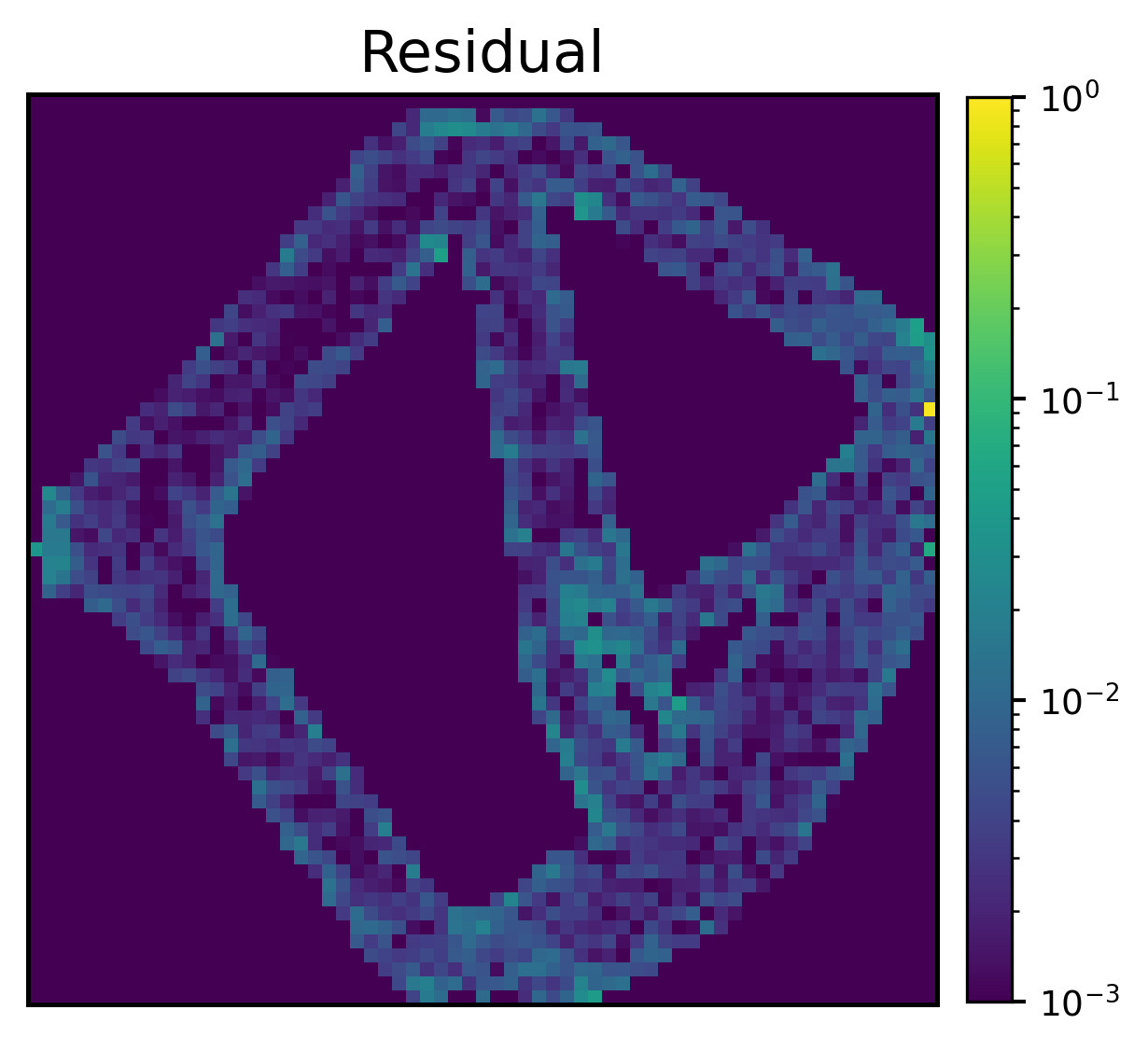}
\toporow{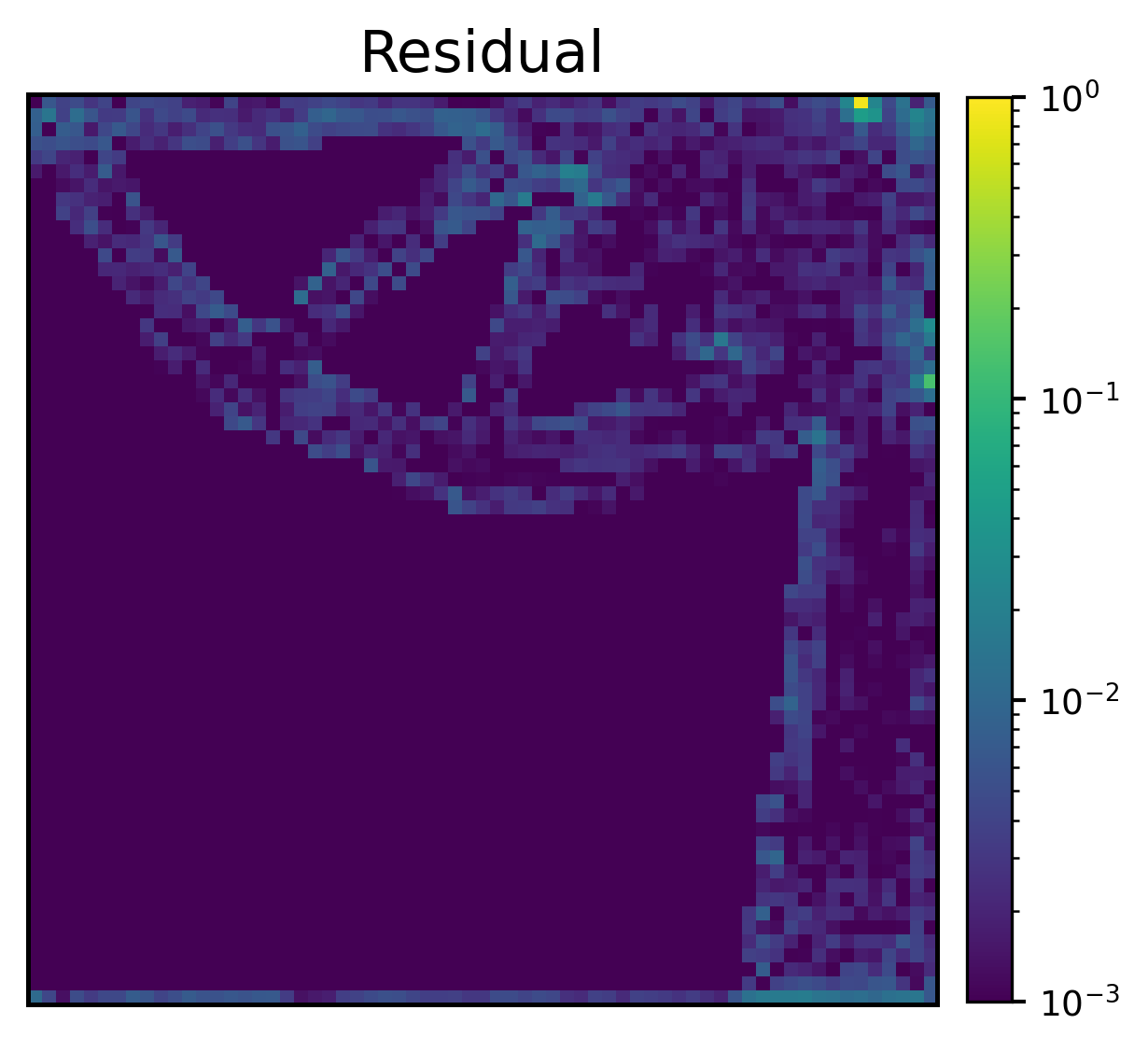}{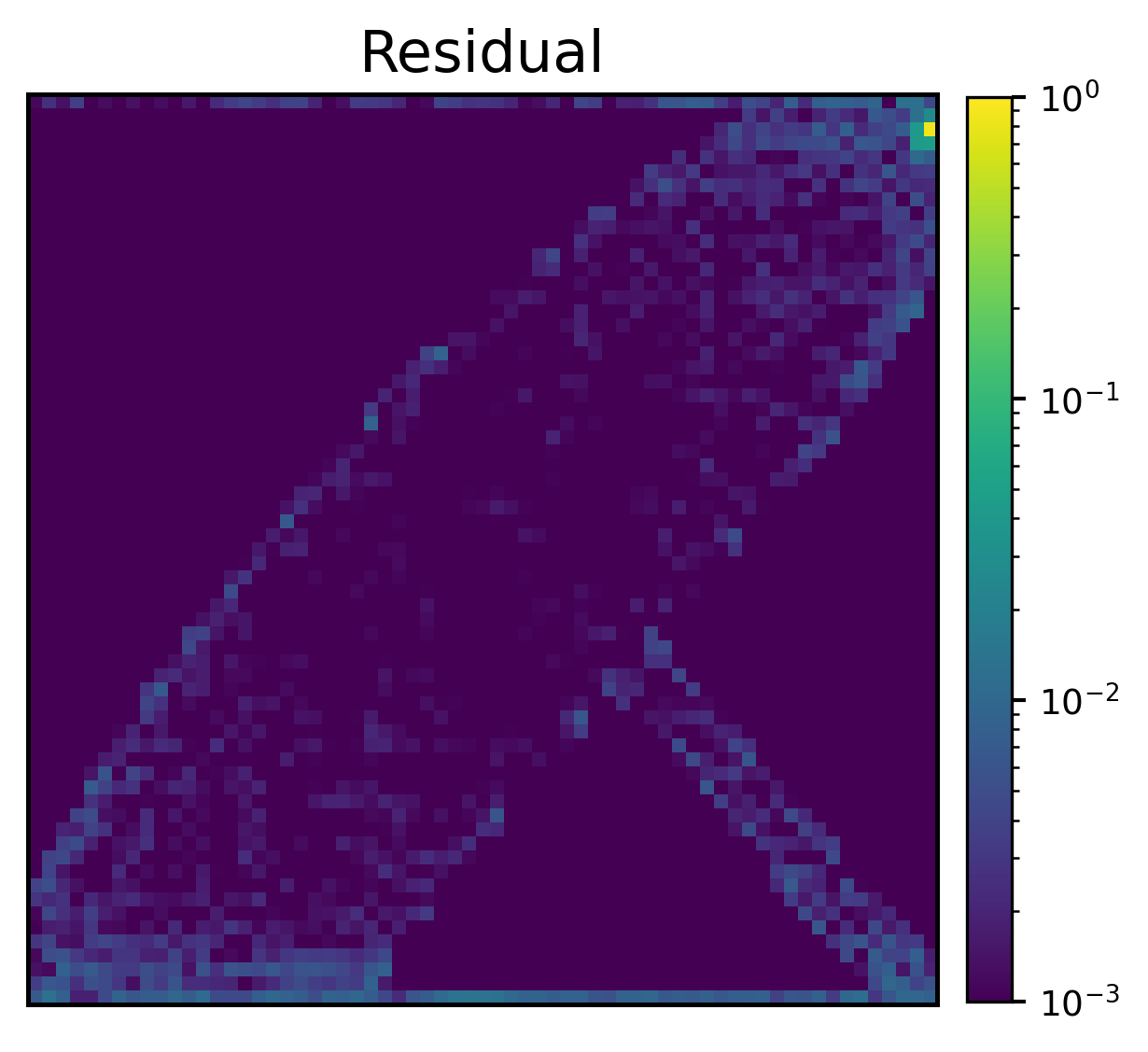}
\toporow{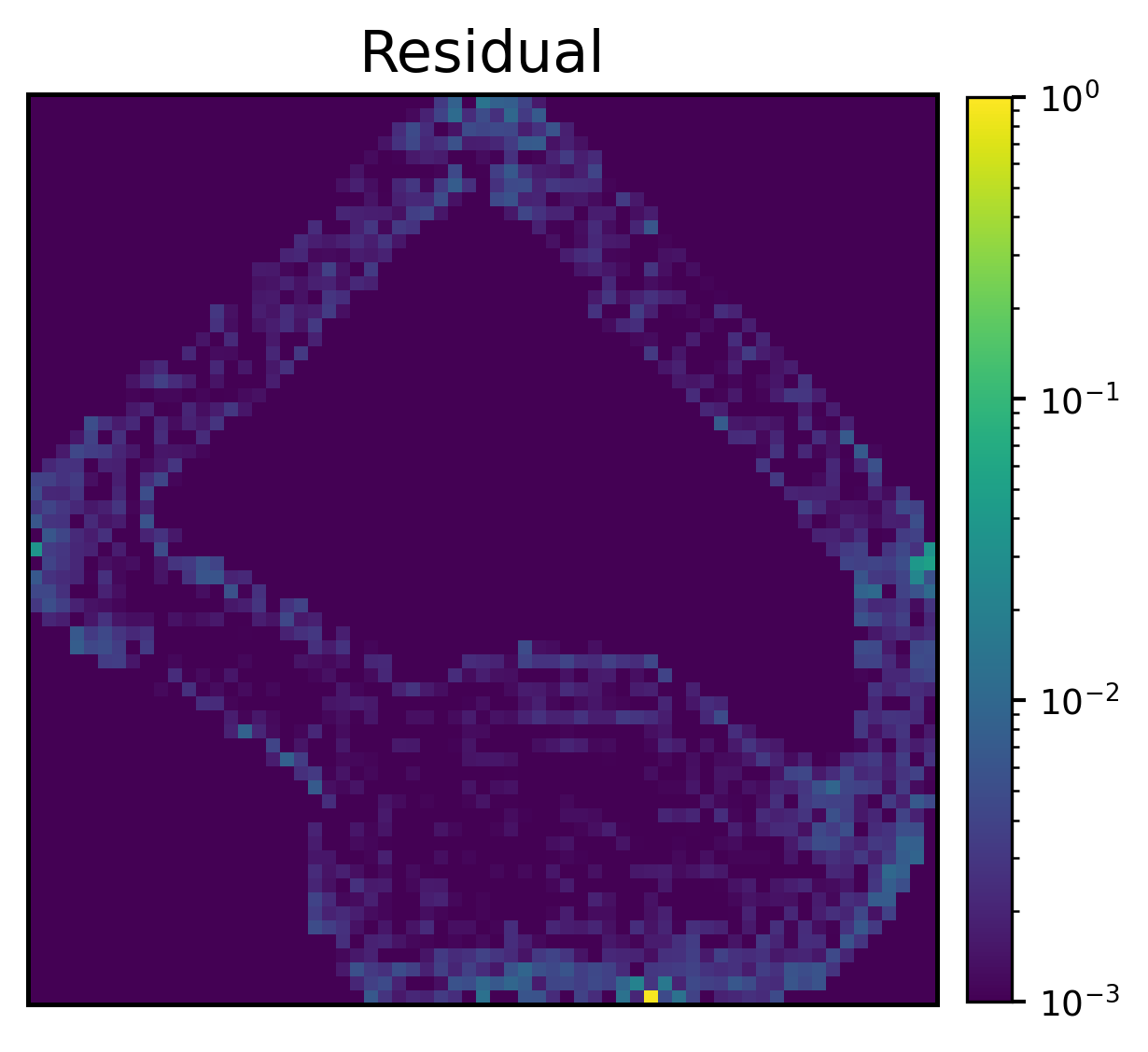}{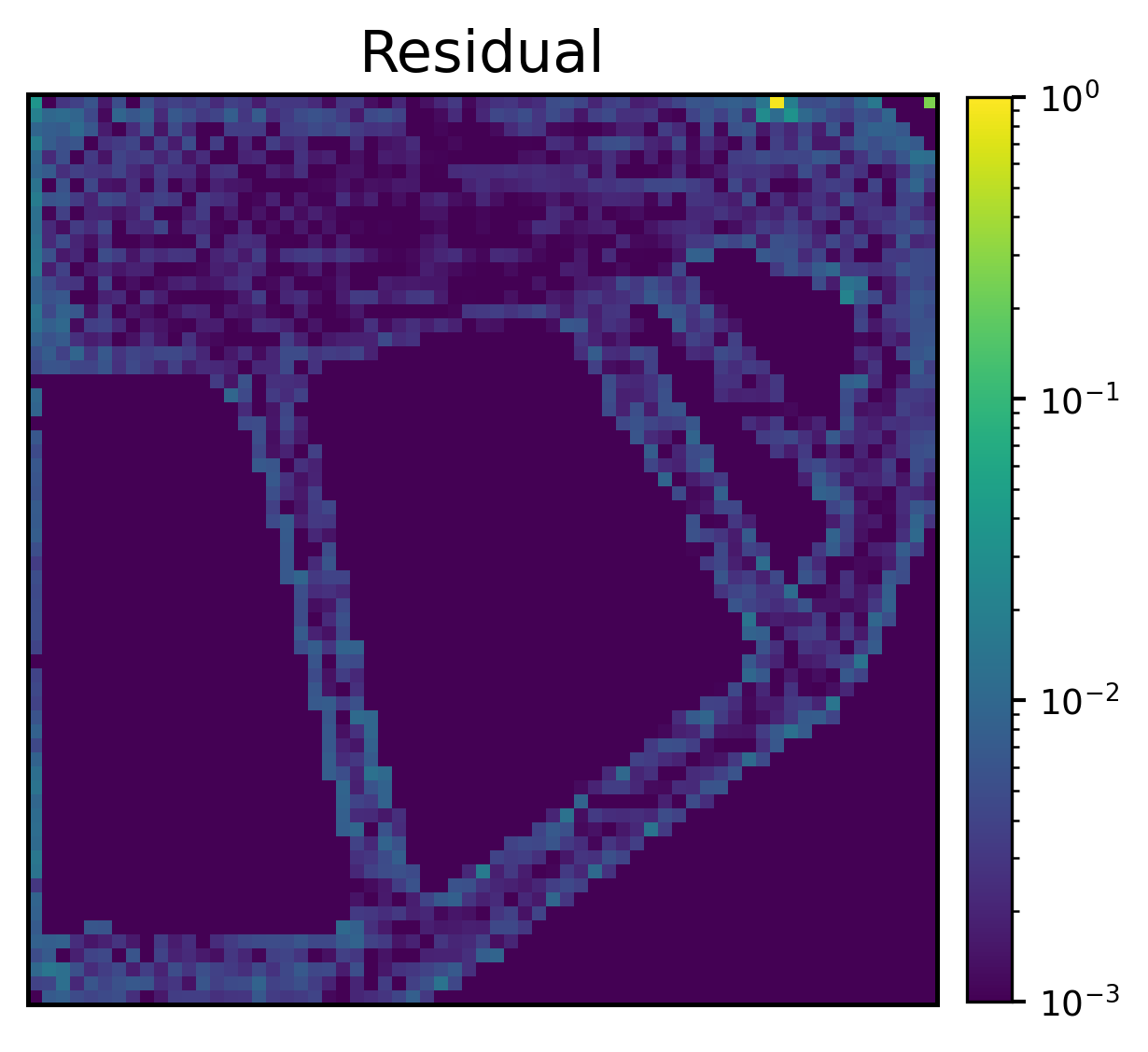}

\caption{Generated topology-optimization samples. Columns 1 and 3 show the generated density fields $\rho$, while columns 2 and 4 show the corresponding equilibrium residuals.}
\label{fig:topology_generated_4x4}
\end{figure}

\paragraph{Visualization Results.} Qualitative results are shown in \autoref{fig:topology_generated_4x4}. Our method generates feasible topology designs while maintaining low equilibrium residuals.

\section{Additional Theoretical Background}

\subsection{Flow Map Learning}\label{sec:flow_map_learning}

As discussed in \autoref{sec:one_step_generation}, flow-map learning aims to directly learn the flow map $\psi_{t\to r}$, so that one-step generation can be performed via $x_1=\psi^\theta_{0\to 1}(x_0)$, where $x_0 \sim p_0$.  Previously, similar to multi-step generative models, there is no easily computable closed-form target flow map $\psi_{t\to r}(x_0)$ available for direct supervision. Therefore, one cannot simply train the model with a direct objective of the form $\|\psi^\theta_{t\to r}(x_0)-\psi_{t\to r}(x_0)\|^2$.(In this work, we address this difficulty through a long-short flow-map formulation.)

Instead, flow-map learning must rely on regression-based surrogate objectives, much like multi-step methods such as Flow Matching. In Flow Matching, this is achieved by constructing a conditional path $p_t(x_t|x_1)$ and using the corresponding conditional velocity $u_t(x_t|x_1)$ to regress the marginal velocity $\mathcal{L}(\theta) = \mathbb{E}_{t, x_1\sim p_1, x_t\sim p_{t|1}(x_t|x_1)}[\|u^\theta(x_t,t)-u_t(x_t| x_1)\|^2]$. The difficulty for flow maps, however, is that unlike instantaneous velocity fields $u_t(x_t)$, they do not admit a self-consistent conditional counterpart \cite{li2026trajectory}.
\begin{theorem}[(Non-existence of conditional flow maps; Theorem 4.1 of \cite{li2026trajectory}).]\label{thm:non_exist}
There exists no conditional flow map $\psi_{t\to r}(x | x_1)$ that simultaneously (i) is consistent with the conditional velocity $u(x | x_1)$ under \autoref{eq:evolv_flow_map}, and (ii) satisfies the consistency relation $\psi_{t\to r}(x)=\mathbb{E}_{x_{1}\sim p_t(x_1| x)}[\psi_{t\to r}(x| x_1)]$ with the marginal flow map. As a result, a self-consistent conditional cumulative field does not exist.    
\end{theorem}
\begin{proof}
Let the mappings $\psi_{t\to r}(x)$ defined by (1) and (2) be denoted by $\psi^{(1)}_{t\to r}(x)$ and $\psi^{(2)}_{t\to r}(x)$, respectively. Specifically, $\psi^{(1)}_{t\to r}(x)=\psi_{0\to r}(\psi_{0\to t}^{-1}(x))$ and $\psi^{(2)}_{t\to r}(x)=\mathbb{E}_{x_1\sim p_t(x_1| x)}[\psi_{t\to r}(x| x_1)]$. 

It suffices to show that $\psi^{(1)}_{t\to r}(x)\neq \psi^{(2)}_{t\to r}(x)$. To do so, it is enough to prove that
$\frac{d}{dr}\psi^{(1)}_{t\to r}(x)\neq \frac{d}{dr}\psi^{(2)}_{t\to r}(x)$.  For $\psi^{(2)}_{t\to r}(x)$, we have
\begin{equation}
    \begin{aligned}
        \frac{d}{dr}\psi^{(2)}_{t\to r}(x)
        &= \frac{d}{dr}\int_{x_1}\psi_{t\to r}(x| x_1)\,p_t(x_1| x)\,dx_1 \\
        &= \int_{x_1}\frac{d}{dr}\psi_{t\to r}(x| x_1)\,p_t(x_1| x)\,dx_1 \\
        &= \int_{x_1}u_r(\psi_{t\to r}(x| x_1)| x_1)\,
        \frac{p_{\mathrm{data}}(x_1)p_t(x| x_1)}{p_t(x)}\,dx_1 \\
        &= \int_{x_1}u_r(\psi_{t\to r}(x| x_1)| x_1)\,p_{\mathrm{data}}(x_1)\,dx_1 \\
        &= \int_{x_1}(x_1-x)\,p_{\mathrm{data}}(x_1)\,dx_1 \\
        &= \mathbb{E}_{x_1\sim p_{\mathrm{data}}}[x_1]-x.
    \end{aligned}
\end{equation}
Consequently, if $\psi^{(1)}_{t\to r}(x)=\psi^{(2)}_{t\to r}(x)$, then we must have 
\begin{equation}
    \begin{aligned}
 \frac{d}{dr}\psi_{0\to r}(x)=\mathbb{E}_{x_1\sim p_{\mathrm{data}}}[x_1]-x       
    \end{aligned}
\end{equation}
which implies $\psi_{0\to r}(x)=(\mathbb{E}_{x_1\sim p_{\mathrm{data}}}[x_1]-x)r+x$. In particular, at $r=1$, this yields
\begin{equation}
    \begin{aligned}
        p_{\mathrm{data}}=(\psi_{0\to 1})_\sharp p_0
        =\delta_{\mathbb{E}_{x_1\sim p_{\mathrm{data}}}[x_1]},   
    \end{aligned}
\end{equation}
where $\delta$ denotes the Dirac distribution concentrated at a single point. This is a contradiction unless the data distribution is degenerate. Therefore, such a conditional flow map cannot exist.
\end{proof}

Because a self-consistent conditional flow map does not exist, flow-map learning is typically carried out through regression objectives built around trajectory consistency (Proposition \autoref{prop:trajectory_consistency}). In particular, one may define the trajectory-consistency loss based on Proposition \autoref{prop:trajectory_consistency} as $\mathcal{L}^{\mathrm{TC}}(\theta) = \mathbb{E}_{t,s,r,x_t\sim p_t}
[
\|
\psi^\theta_{t\to r}(x_t)
-
\psi^\theta_{s\to r}(\psi^\theta_{t\to s}(x_t))
\|_2^2
$.  However, $\mathcal{L}^{\mathrm{TC}}$ does not contain dataset-level supervision, since every term in the loss is produced by the model itself. Therefore, $\mathcal{L}^{\mathrm{TC}}$ can only enforce internal consistency of the learned flow map, but cannot guarantee that the learned map generates the desired target distribution $p_1$.

To address this issue, existing flow-map learning methods mainly fall into two categories: \emph{progressive extension} methods and \emph{derivative-based} methods.
\begin{itemize}
    \item \textbf{Progressive extension}
    (ShortCut \cite{frans2025shortcut}, PSD of FMM \cite{boffi2025flow}, SplitMeanFlow \cite{guo2025splitmeanflow}):
    these methods extend instantaneous dynamics to longer flow maps through consistency constraints.
    For example, ShortCut first learns the instantaneous velocity by regression loss with supervision from the dataset
    \begin{equation}
        \begin{aligned}
    \mathcal{L}(\theta) = \mathbb{E}_{x_1\sim p_1, x\sim p_{t|1}(\cdot|x_1)}[\|u^\theta_{t\to t+d}(x)-u_t(x| x_1)\|^2],
    \qquad
    u_t(x| x_1)=\frac{x_1-x}{1-t},
    \quad x_1\in \mathcal{D},            
        \end{aligned}
    \end{equation}
    and then extends to longer intervals through consistency relations such as
    \begin{equation}
        \begin{aligned}
            \mathcal{L}(\theta) = \mathbb{E}_{x_t\sim p_t}[\| u^\theta_{t\to t+2d}(x_t)- \frac{1}{2}
    (
    u^\theta_{t\to t+d}(x_t)
    +
    u^\theta_{t+d\to t+2d}(x_{t+d})
    )
    \|^2]
        \end{aligned}
    \end{equation}
    where $x_{t+d} = d\,u_{t\to t+d}(x_t) + x_t$, and $d$ is sampled uniformly at random from $\{\frac{1}{128}, \frac{1}{64}, \ldots, \frac{1}{2}\}$.  As a result, the long-range map receives no direct supervision from the dataset $\mathcal{D}$; instead, it is learned recursively from shorter model outputs.
    \item \textbf{Derivative-based methods}
    (MeanFlow \cite{geng2025mean}, Consistency Models \cite{song2023consistency}, ESD and LSD of FMM \cite{boffi2025flow}):  these methods relate the flow map $\psi_{t\to r}$ to the instantaneous velocity field $u_t = \frac{d}{dr}\psi_{t\to r}|_{r=t}$ through the continuous equations $\frac{d}{dt}\psi_{t\to r}(x) + u_t(x) \nabla  \psi_{t\to r}(x) = 0 $ or $\frac{d}{dr}\psi_{t\to r}(x) = u_r(\psi_{t\to r}(x)) $ implied by trajectory consistency (Proposition \autoref{prop:trajectory_consistency}), obtained by appropriate reformulations and by taking the limits $s\to t$ and $s\to r$, respectively. This connection allows them to supervise long-range flow maps $\psi_{t\to r}$ more directly using data-dependent conditional velocities $u_t(x|x_1)$.  For example, MeanFlow uses an objective of the form
    \begin{equation}
        \begin{aligned}
            \mathcal{L}(\theta) = \mathbb{E}_{x\sim p_t}[|u_{t\to r}^\theta(x) - \mathrm{sg}\![u_t(x| x_1) +
    (r-t)(u_t(x| x_1)\partial_x u_{t\to r}^\theta(x)+ \partial_tu_{t\to r}^\theta(x))]|^2]
        \end{aligned}
    \end{equation}
    where $u_t(x| x_1)=\frac{x_1-x}{1-t}, x_1\in \mathcal{D}$.  This provides more direct supervision than progressive extension methods. However, it also leads to a more complicated regression target and typically requires Jacobian-vector products (JVPs), which can significantly increase memory usage, computational cost, and optimization instability.
\end{itemize}

Other methods, such as AlphaFlow \cite{zhang2025alphaflow}, can be viewed as hybrids of these two categories. Overall, compared with multi-step generative models that only need to learn instantaneous fields, the regression objectives for flow-map learning are structurally more complicated. This further increases the difficulty of incorporating physical constraints into one-step or few-step flow-map learning.

\subsection{Closed-Form Solution}\label{appendix:closed_form}
\begin{wrapfigure}{r}{0.45\linewidth}
    \vspace{-8mm}
    \centering
    \includegraphics[width=\linewidth]{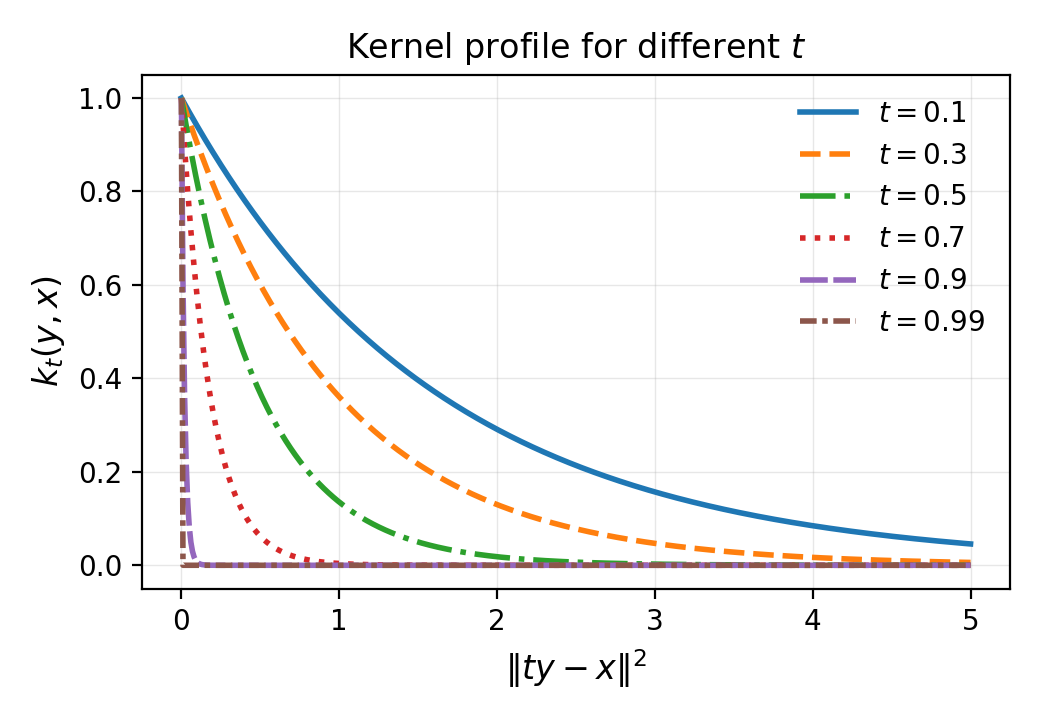}
    \vspace{-7mm}
    \caption{Kernel profiles for different values of $t$.}
    \label{fig:kernel_profile}
    \vspace{-11mm}
\end{wrapfigure}
In addition to optimizing indirect regression losses, prior work~\cite{bertrand2025closed} has also derived a closed-form expression for the Flow Matching velocity field $u_t(x) = \frac{\mathbb{E}_{x_1\sim p_1}[\frac{x_1-x}{1-t}k_{t}(x_1,x)]}{\mathbb{E}_{x_1\sim p_1}[k_{t}(x_1,x)]}$ . 
However, this closed form is mainly used for theoretical analysis rather than practical training. The main reason is that it involves the computation of a kernel $k_{t}(y,x) = e^{-\frac{\|ty-x\|^2}{2(1-t)^2}}$, whose value depends on the time-dependent coefficient $\frac{1}{(1-t)^2}$. When $t$ is far from the endpoint $1$, this coefficient makes the kernel relatively flat, as illustrated in \autoref{fig:kernel_profile}. As a result, accurately estimating the expectation $\mathbb{E}_{x_1\sim p_1}[\cdot k_{t}(x_1,x)]$ inside the kernel-weighted velocity field $u_t$ requires a large number of samples, making direct training with this closed-form velocity ineffective.

In our method, the long-short flow-map decomposition allows us to use the closed-form Flow Matching dynamics only in the near-endpoint regime $t\to 1$. In this regime, the kernel becomes sufficiently sharp, so estimating the expectation $\mathbb{E}_{x_1\sim p_1}[\cdot\, k_t(x_1,x)]$ does not require a large number of samples, as discussed in \autoref{sec:algorithm}. Moreover, to unify the data-driven effect with the physics-driven effect, we do not use the closed-form velocity directly. Instead, we adopt the potential-based formulation in \autoref{thm:potential_close_form}, which also avoids the singularity of the closed-form velocity field at $t=1$.


\begingroup
\color{black}
\captionsetup{font={color=black},labelfont={color=black}}
\section{Additional Darcy Comparisons and Experimental Settings}
\label{app:additional_results}

\subsection{Conflict-Free One-Step Comparisons}
\label{app:add_baselines}
We compare PI-Consistency (implementing sCM), PI-MeanFlow, and PI-Shortcut on Darcy. Each retains its original training objective and procedure, with the same terminal physics loss added under matched architecture, data, batch size, residual, and training budget. Table~\ref{tab:add_conflict} reports results with and without the conflict-free (CF) optimization strategy from PBFM~\cite{baldan2026physics}. This control tests whether resolving data--physics gradient conflicts alone is sufficient to close the performance gap. CF updates improve the physical residuals of these baselines but do not recover the overall data--physics performance of PI-LSFM, which uses no CF update. The remaining gap supports the benefit of our direct data-and-physics objective beyond the gains obtained by conflict-free optimization of the tested indirect objectives.

\begin{table}[H]\color{black}
\centering
\caption{One-step Darcy baselines with and without conflict-free (CF) updates.}
\label{tab:add_conflict}
\begin{tabular}{llrrr}
\toprule
Method & CF & RE $\downarrow$ & WD$\cdot10^2$ $\downarrow$ & JS$\cdot10^1$ $\downarrow$\\
\midrule
PI-Consistency & No & 0.240 & 3.889 & 5.023\\
PI-Consistency & Yes & 0.152 & 3.017 & 3.933\\
PI-MeanFlow & No & 4603.573 & 4.243 & 5.457\\
PI-MeanFlow & Yes & 4.973 & 4.253 & 5.212\\
PI-Shortcut & No & 0.435 & 2.336 & 2.518\\
PI-Shortcut & Yes & 0.276 & 2.184 & 2.363\\
PI-LSFM & No & 0.055 & 0.108 & 0.171\\
\bottomrule
\end{tabular}
\end{table}

\subsection{Three-Seed Darcy Check}
\label{app:add_seeds}
Table~\ref{tab:add_seeds} reports three additional independent Darcy runs trained for 100K steps each. These runs provide a stability check at 100K steps, rather than three fully converged 400K-step runs. Our main benchmark results use the fixed seed 42 without seed selection.

\begin{table}[H]\color{black}
\centering
\caption{Three-seed Darcy stability check at 100K steps.}
\label{tab:add_seeds}
\begin{tabular}{lrrr}
\toprule
Seed & RE $\downarrow$ & WD$\cdot10^2$ $\downarrow$ & JS$\cdot10^1$ $\downarrow$\\
\midrule
43 & 0.132 & 0.269 & 0.241\\
123 & 0.116 & 0.215 & 0.210\\
2026 & 0.127 & 0.308 & 0.259\\
Mean $\pm$ std & $0.125\pm0.008$ & $0.264\pm0.047$ & $0.237\pm0.025$\\
\bottomrule
\end{tabular}
\end{table}

\subsection{Batch-Size Sensitivity}
\label{app:add_batch}
Table~\ref{tab:add_batch} reports the Darcy batch-size sweep, with each setting trained for 100K steps. Batch sizes 32 and 64 retain useful data--physics performance, while the results deteriorate at 16 and 8. The sweep evaluates unconditional Darcy generation.

\begin{table}[H]\color{black}
\centering
\caption{Darcy batch-size sensitivity after 100K training steps per setting.}
\label{tab:add_batch}
\begin{tabular}{lrrr}
\toprule
Batch size & RE $\downarrow$ & WD$\cdot10^2$ $\downarrow$ & JS$\cdot10^1$ $\downarrow$\\
\midrule
64 (default) & 0.117 & 0.213 & 0.216\\
32 & 0.193 & 0.503 & 0.441\\
16 & 13.715 & 7.572 & 2.771\\
8 & 5.971 & 7.395 & 2.570\\
\bottomrule
\end{tabular}
\end{table}

\subsection{Default Settings and Selection Protocol}
\label{app:add_settings}
The data and physics discretizations are selected through the Darcy design studies and retained across all four benchmarks. The bandwidth $\epsilon=0.1$ is shared across tasks. The Darcy physics weight comes from its design-study sweep; the other benchmarks reuse prior-work settings~\cite{baldan2026physics} without a new per-task search. Table~\ref{tab:add_settings} summarizes the relative physics weights $\bar\lambda$, encoders, and discretizations. Darcy and Kolmogorov use an MAE encoder; dynamic stall and topology optimization train without an encoder.

\begin{table}[H]\color{black}
\centering
\caption{Per-benchmark default settings used in the reported experiments.}
\label{tab:add_settings}
\begin{tabular}{lrrlll}
\toprule
Benchmark & $\bar\lambda$ & $\epsilon$ & Encoder & Data & Physics\\
\midrule
Darcy & 3.0 & 0.1 & MAE & TR & FE\\
Kolmogorov & 1.0 & 0.1 & MAE & TR & FE\\
Dynamic stall & 1.0 & 0.1 & None & TR & FE\\
Topology optimization & 0.01 & 0.1 & None & TR & FE\\
\bottomrule
\end{tabular}
\end{table}
\endgroup

\end{document}